\documentclass{article}

\usepackage{arxiv}

\usepackage[utf8]{inputenc} 
\usepackage[T1]{fontenc}    
\usepackage{hyperref}       
\usepackage{url}            
\usepackage{booktabs}       
\usepackage{amsfonts}       
\usepackage{nicefrac}       
\usepackage{microtype}      
\usepackage{lipsum}
\usepackage{graphicx}
\usepackage{amsmath}
\mathchardef\mhyphen="2D
\usepackage{adjustbox}
\usepackage[center]{caption}
\usepackage{subcaption}

\usepackage{multirow}     
\graphicspath{ {./images/} }

\title{Neurochaos Feature Transformation and Classification for Imbalanced Learning}

\author{
 Deeksha Sethi \\
  Department of Electronics and Communication Engineering\\
  BMS Institute of Technology and Management\\
  Bengaluru - 560064, Karnataka, India \\
  \texttt{deeksha.sethi03@gmail.com} \\
   \And
 Nithin Nagaraj \\
  Consciousness Studies Programme\\
  National Institute of Advanced Studies\\
  Indian Institute of Science Campus \\
  Bengaluru - 560012, Karnataka, India \\
  \texttt{nithin@nias.res.in} \\
  \And
 Harikrishnan N B \\
  Consciousness Studies Programme\\
  National Institute of Advanced Studies\\
  Indian Institute of Science Campus \\
  Bengaluru - 560012, Karnataka, India \\
  \\
  The University of Trans-Disciplinary Health Sciences And Technology \\
  Bengaluru - 560064, Karnataka, India \\
  \texttt{harikrishnannb07@gmail.com} \\
}

\begin{document}
\maketitle
\begin{abstract}
Learning from limited and imbalanced data is a challenging problem in the Artificial Intelligence community. Real-time scenarios demand decision-making from rare events wherein the data are typically imbalanced. These situations commonly arise in medical applications, cybersecurity, catastrophic predictions etc. This motivates development of learning algorithms capable of learning from imbalanced data. Human brain effortlessly learns from imbalanced data. Inspired by the chaotic neuronal firing in the human brain, a novel learning algorithm namely \emph{Neurochaos Learning} (NL) was recently proposed. NL is categorized in three blocks: Feature Transformation, Neurochaos Feature Extraction (CFX), and Classification. In this work, the efficacy of neurochaos feature transformation and extraction for classification in imbalanced learning is studied. We propose a unique combination of neurochaos based feature transformation and extraction with traditional ML algorithms. The explored datasets in this study revolve around medical diagnosis, banknote fraud detection, environmental applications and spoken-digit classification. In this study, experiments are performed in both high and low training sample regime.
In the former, five out of nine datasets have shown a performance boost in terms of macro F1-score after using CFX features. The highest performance boost obtained is $\textbf{25.97\%}$ for {\it Statlog (Heart)} dataset using CFX+Decision Tree. In the low training sample regime (from just one to nine training samples per class), the highest performance boost of $\textbf{144.38\%}$ is obtained for {\it Haberman's Survival} dataset using CFX+Random Forest. NL offers enormous flexibility of combining CFX with any ML classifier to boost its performance, especially for learning tasks with limited and imbalanced data.
\end{abstract}


\section{Introduction}
Technological advancements have made a paradigm shift in the evolution of science. A driving force behind this shift is the high storage and computational capacity available in this era. This has given rise to computational techniques for analysis and pattern discovery from data, popularly known as \emph{data-driven science}. Data can be structured or unstructured. In today's world the data analytics techniques have to make sense from big data. This demands intelligent approaches towards meaningful feature extraction from data and thereby contribute to decision making. The bedrock for this approach of data-driven science comes under the purview of Artificial Intelligence (AI). Artificial Intelligence (AI) can be defined as an anarchy of methods~\cite{lehman2014anarchy}. This anarchy involves (a) Symbolic AI (logic-based AI), (b) Statistical Learning (Machine Learning), and (c) Sub-Symbolic AI (brain-inspired learning)~\cite{mitchell2019artificial}. The aforementioned verticals of AI have seen a remarkable progress in recent past. On the other hand, these approaches are limited when it comes to decision making under the presence of rare events~\cite{HAIXIANG2017220}. 

Rare events by definition are events whose frequency of occurrence are significantly less compared to more frequently occurring events~\cite{maalouf2011robust}. Some examples of rare events are: outbreak of a pandemic~\cite{ciotti2020covid, spanishflu}, cyber attack~\cite{BEAMAN2021102490}, fraudulent transactions~\cite{panigrahi2009credit} and natural disasters~\cite{maalouf2011robust}. In this scenario, it is critical to provide an early warning response to take appropriate decisions. Hence, the accurate prediction or classification of rare events is of immense importance. Failing to do so can lead to a catastrophe. For example, in the case of COVID-19 pandemic, wrongly classifying a positive person can increase the spread rate of the viral infection. To curb the spread, it is important to correctly classify the infected people and take appropriate measures. Thus, the major challenge in the classification of all rare events is the lack of sufficient data. This boils down to the problem of imbalanced learning. Imbalanced learning deals with an unequal distribution of data instances amongst distinct classes of a dataset. This can mean either one or more classes in a dataset have relatively greater number of data instances as compared to the remaining classes~\cite{HAIXIANG2017220}. 
Frequently implemented approaches to imbalanced learning can be categorized into (a) pre-processing strategies~\cite{HAIXIANG2017220} and (b) cost sensitive learning~\cite{elkan2001foundations}. The main idea behind pre-processing methods is to optimize the feature space and thereby perform classification by ensemble or algorithmic techniques~\cite{HAIXIANG2017220}. On the other hand, cost-sensitive learning assigns a cost (penalty) for every misclassification, with the end goal of minimizing the overall cost~\cite{elkan2001foundations}.

One widely used pre-processing technique is resampling. Resampling restores balance in the sample space by modifying the samples as follows:
\begin{enumerate}
    \item Over-sampling: This approach addresses resampling by generating new samples for the minority classes. Ex: SMOTE~\cite{chawla2002smote}.
    \item Under-sampling: This type of resampling is done by eliminating the innate samples in the majority classes. Ex: Random Under-Sampling~\cite{tahir2009multiple}.
\end{enumerate}
Combinations of over-sampling and under-sampling methods are known as hybrid methods.

Another common pre-processing technique involves feature selection and feature extraction. Feature selection provides a subset of original input features. Whereas, feature extraction does a functional mapping of input features to generate new features~\cite{HAIXIANG2017220}. Principal Component Analysis~\cite{kuncheva2013pca}, Dynamic Mode Decomposition~\cite{ogundile2021dynamic}, Mel-scale Frequency Cepstral Coefficient~\cite{hossan2010novel}, Non Negative Matrix Factorization~\cite{wang2012nonnegative}, Empirical Mode Decomposition~\cite{nagarajan2019feature} etc. are examples of feature extraction algorithms. A detailed study on pre-processing, feature selection and feature extraction are provided in~\cite{HAIXIANG2017220}. It should be noted that most of these methods make use of {\it linear} transformations.

In this paper, we explore a recent study on a nonlinear chaos based learning algorithm namely \emph{Neurochaos Learning} (NL), proposed in ~\cite{balakrishnan2019chaosnet, harikrishnan2021noise}. NL draws its inspiration from the {\it nonlinear chaotic firing} that is intrinsic to neurons in the brain. The human brain is a highly complex system with an estimated $86$ billion interconnected neurons~\cite{azevedo2009equal}. Individual neurons in the human brain are known to exhibit a wide variety of behaviors -- ranging from periodic, quasi-periodic to fully chaotic. Chaos is central to the brain since chaotic behaviour is found at different spatiotemporal scales starting from individual neurons to networks of neurons~\cite{korn2003there}.

NL also addresses an essential practice in Machine Learning: Feature transformation. 
NL proposes a novel approach by transforming pre-existing features into chaos-based features~\cite{balakrishnan2019chaosnet}. These features are passed to a classifier for decision making. For the first time, in~\cite{balakrishnan2019chaosnet}, the authors employ the rich properties of chaos in learning algorithms for classification. Also, a performance boost in classification in the few-shot learning regime has been shown using chaos-based-features in~\cite{balakrishnan2019chaosnet,  nb2020neurochaos}.

The effectiveness of NL is shown in the classification of various datasets such as {\it MNIST, Iris, Exoplanet}~\cite{balakrishnan2019chaosnet}, {\it Coronavirus Genome classification}~\cite{nb2020neurochaos}, especially in the low training sample regime. Recently,~\cite{kathpalia2022cause} has demonstrated the efficacy of NL in the classification and preservation of cause-effect for $1$D coupled chaotic maps and coupled AR processes. 

There is a need to rigorously test NL, and in particular the nonlinear chaotic transformation of features, to more datasets in the context of imbalanced learning. This study addresses this gap. We combine NL features with classical machine learning algorithms such as Decision Tree (DT), Random Forest (RF), AdaBoost (AB), Support Vector Machine (SVM), $k$-Nearest Neighbors ($k$-NN) and Gaussian Naive Bayes (GNB). A comparison of NL: chaos-based-hybrid ML architectures with stand-alone ML algorithms is brought out for the following balanced and imbalanced datasets: {\it Iris, Ionosphere, Wine, Bank Note Authentication, Haberman's Survival, Breast Cancer Wisconsin, Statlog (Heart), Seeds, Free Spoken Digit Dataset}. A detailed analysis of this comparative study (NL: chaos-based-hybrid ML vs. stand-alone ML) is carried out for both the high and low training sample regime. Learning from limited data/ imbalanced data is a challenging problem in the ML community. This research highlights the performance comparison of NL and classical ML algorithms in learning from limited training instances.

The organization of this paper is as follows: section~\ref{sec:Proposed Method} explains the proposed architecture and methods being investigated in this paper. The description of all datasets used for the experiments is provided in section~\ref{sec:Dataset Description}. The experiments and their corresponding results are available in section~\ref{sec:Experiments & Results}. A detailed discussion on the inferences is provided in section~\ref{sec:Discussion}. The concluding remarks of this study are in section~\ref{sec:Conclusion}.

\section{Proposed Method}
\label{sec:Proposed Method}
\begin{figure}[!ht]
\includegraphics[width=\textwidth]{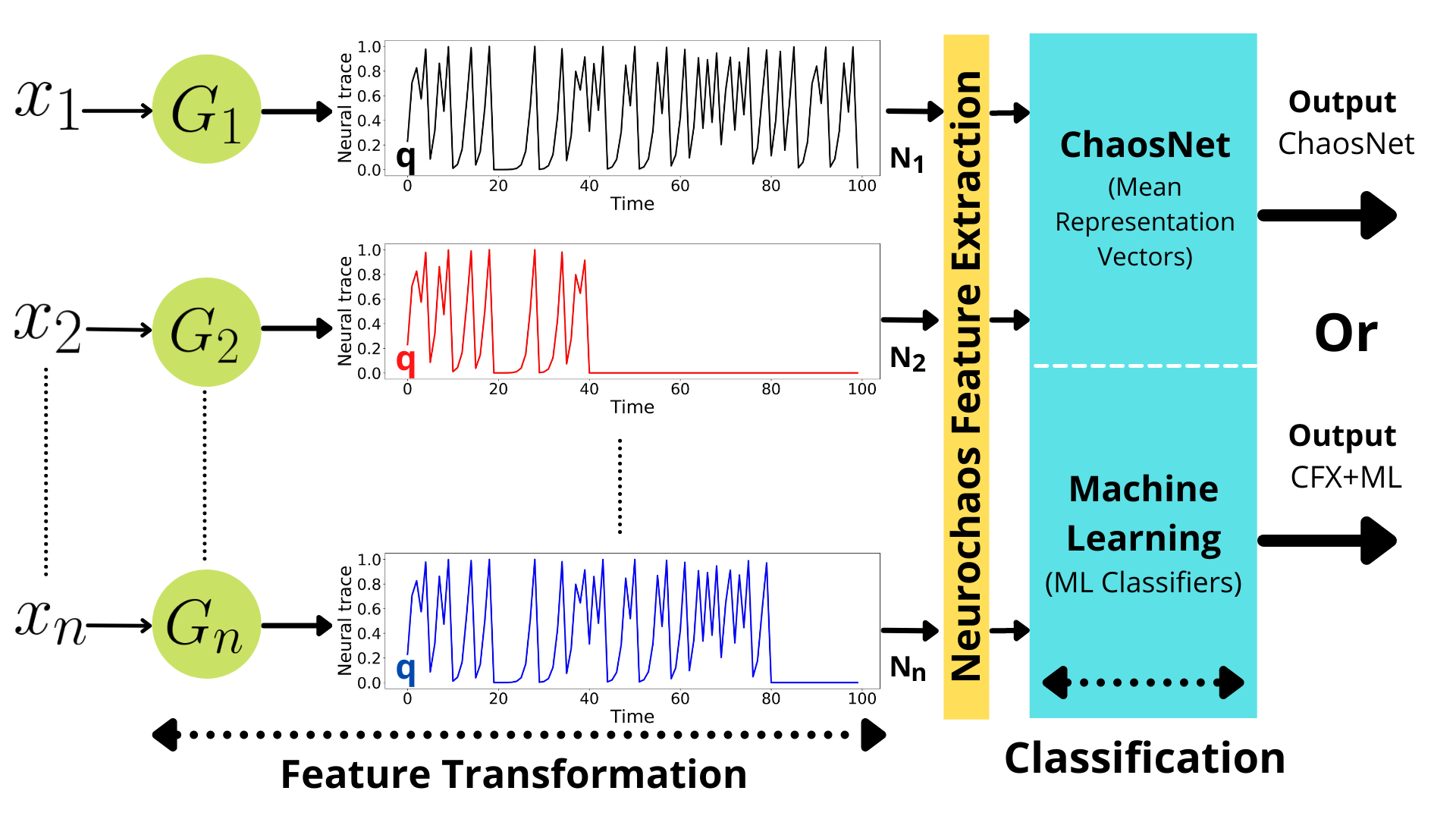}

\caption{The three steps proposed in this study. (a) feature transformation, (b) neurochaos feature extraction, and (c) classification. The classifier chosen could either be ChaosNet or any other ML classifier. }
\label{Fig_nl_archi}
\end{figure}
In this study, a modified form of the recently proposed Neurochaos Learning architecture~\cite{harikrishnan2021noise} is put forward. This modified NL architecture is depicted in Figure~\ref{Fig_nl_archi}. NL has mainly three blocks - (a) feature transformation, (b) neurochaos feature extraction and (b) classification. A detailed description of each block is given below.
\begin{itemize}
    \item \textbf{Input}: Input is the first and most significant step of any learning process. It contains input attributes obtained from the dataset ($x_1, x_2, \ldots, x_n$). These input attributes (after suitable normalization) are further passed to the feature transformation block. 
    \item \textbf{Feature Transformation}: The feature transformation block consists of an input layer of $1$D Generalized L\"uroth Series (GLS) neurons. The $1$D GLS neurons are piece-wise linear chaotic maps. The number of GLS neurons ($G_1, G_2, \ldots, G_n$) in the input layer is equal to the number of input attributes ($n$) in the dataset. Each neuron $G_1, G_2, \ldots, G_n$ has an initial neural activity of $q$ units. Upon arrival of the input attributes (also known as {\it stimuli})  $x_1, x_2, \ldots, x_n$, each of the $1$D GLS neurons ($G_1, G_2, \ldots, G_n$) starts independently firing with an initial neural activity of $q$ units. The neural trace of these chaotic neurons halts when it reaches the $\epsilon$ neighbourhood of the stimulus (at which point we say that it has successfully recognized  the stimulus). The halting of the neural trace is mathematically guaranteed by the topological transitivity property of chaos~\cite{balakrishnan2019chaosnet}. The transformed input attributes are further passed to the neurochaos feature extraction block.
    \item \textbf{Neurochaos Feature Extraction}:
    The following features are extracted from the chaotic neural trace:

        \begin{enumerate}
        
            \item \emph{Firing time ($N$)}: The time taken by the chaotic neural trace to recognise the stimulus~\cite{harikrishnan2020neurochaos}.
            \item \emph{Firing rate ($R$)}: Fraction of time for which the chaotic neural trace exceeds the discrimination threshold $b$ so as to recognize the stimulus~\cite{harikrishnan2020neurochaos}.
            \item \emph{Energy ($E$)}: 
            A chaotic neural trace $z(t)$ with a firing time $N$ has an energy ($E$) defined as:
                \begin{equation}
                    E = \sum_{t = 1}^N z(t)^{2}.
                \end{equation}
            \item \emph{Entropy ($H$)}: The entropy of a chaotic neural trace $z(t)$ is computed using the symbolic sequence $SS(t)$ of $z(t)$. $SS(t)$ is defined as follows:
\begin{eqnarray}                %
SS(t_i) = \left\{\begin{matrix}
 0,& z(t_i) < b, \\
 1, & b \leq z(t_i) < 1, 
\end{matrix}\right.
\end{eqnarray}
                
            where $i = 1$ to $N$ (firing time). From $SS(t)$, Shannon Entropy $H(SS)$ is computed as follows:
                    \begin{equation}
                        H(SS) = -\sum_{i = 1}^2 p_i \log_2(p_i) ~~\text{bits},
                    \end{equation}
            where, $p_1$ and $p_2$ are  the probabilities of occurrence of symbols $0$ and $1$ in $SS(t)$ respectively.

        \end{enumerate}
        
        An input stimulus $x_k$ of a data instance visiting the $k$-th GLS neuron ($G_k$) is transformed to a $4$D vector $[N_{x_{k}}, R_{x_{k}}, E_{x_{k}}, H_{x_{k}}] $. The CFX feature space contains a collection of all the $4$D vectors after feature transformation. These chaos based features are passed to the third block of the NL architecture i.e, classification.
    \item \textbf{Classification}: There are mainly two kinds of NL architecture: (a) \verb|ChaosNet|, (b) ChaosFEX (CFX) + ML. \verb|ChaosNet| architecture computes the mean representation vector for each class. The mean representation vector of class-$k$ contains the mean firing time, mean firing rate, mean energy and mean entropy of the $k$-th class. \verb|ChaosNet| uses a simplistic decision rule, namely, the cosine similarity of testdata instances with the mean representation vectors. The testdata instance is assigned a label = $l$, if the cosine similarity of that testdata instance with $l$-th mean representation vector is the highest. 
    
    Alternatively, we have the flexibility of choosing any other ML classifier instead of ChaosNet. In this kind of NL architecture, the ChaosFEX features are fed directly to the ML classifier (CFX+ML). In this study, we have tested with the following ML algorithms --  Decision Tree (DT), Random Forest (RF), AdaBoost (AB), Support Vector Machine (SVM), $k$-Nearest Neighbors ($k$-NN) and Gaussian Naive Bayes (GNB). CFX+ML is a hybrid NL architecture that combines the best of chaos and machine learning.
    \item \textbf{Output}: The output obtained from the classification block serves as output for that respective NL architecture. On using \verb|ChaosNet| in the classification block, the output is an outcome of classification based on the mean representation vectors. The Machine Learning implementation in the classification block produces an output dependent on the choice of the ML classifier.
    
\end{itemize}

In~\cite{nb2020neurochaos}, it has been shown that \verb|ChaosNet| satisfies the \emph{Universal Approximation Theorem} with a bound on the number of chaotic neurons to approximate the finite support discrete time function. This is due to the uncountably infinite number of dense orbits and the \emph{topological transitivity} property of chaos. Another important feature of \verb|ChaosNet| and CFX+ML is the  natural presence of Stochastic Resonance (noise enhanced signal processing)~\cite{harikrishnan2021noise} found in the architecture. An optimal performance of \verb|ChaosNet| and CFX+ML is obtained for an intermediate value of noise intensity ($\epsilon$). This has been thoroughly shown in~\cite{harikrishnan2021noise}.

\section{Dataset Description}
\label{sec:Dataset Description}
The ChaosFEX (CFX) feature extraction algorithm requires normalization of the dataset and numeric codes for labels. Therefore, to maintain uniformity, all datasets are normalized\footnote{$X\_norm = \frac{X-\min(X)}{\max(X) -\min(X)}.$} for both stand-alone algorithms and their integration with CFX. The labels are renamed to begin from zero in each dataset to ensure compatibility with CFX feature extraction. The rules followed for the numeric coding for the labels of all the datasets are provided in section~\ref{sec:Supplementary Information} (Table~\ref{table:Iris rule} -~\ref{table:FSDD rule}).

\subsection{Iris} \label{subsec:Iris Dataset}

{\it Iris}~\cite{scikit-learn, iris_original} aids classification of three iris plant variants: Iris Setosa, Iris Versicolour, and Iris Virginica. There are $150$ data instances in this dataset with four attributes in each data instance: sepal length, sepal width, petal length, and petal width. All attributes are in \textit{cms}. The specified class distribution provided in Table~\ref{table:test-train split} is in the following order: (Setosa, Versicolour, Virginica).

\subsection{Ionosphere} \label{subsec:Ionosphere Dataset}

The {\it Ionosphere}~\cite{UCI, ionosphere_original} dataset enables a binary classification problem. The classes represent the status of returning a radar signal from the Ionosphere. Label `g' (Good) denotes the return of the radar signal, and label `b' (Bad) indicates no trace of return of the radar signal. The goal of this experiment is to identify the structure of the Ionosphere using radar signals. This dataset has $351$ data instances and $34$ attributes. The specified class distribution provided in Table~\ref{table:test-train split} is as follows: (Bad, Good). 

\subsection{Wine} \label{subsec:Wine Dataset}
      
{\it Wine}~\cite{UCI, wine_original} dataset aims to identify the origin of different wines using chemical analysis. The classes are labeled `$1$', `$2$', and `$3$'. It has $178$ data instances and $13$ attributes ranging from alcohol, malic acid to hue and proline for the collected samples. The specified class distribution provided in Table~\ref{table:test-train split} is as follows: $(1, 2, 3)$.

\subsection{Bank Note Authentication} \label{subsec:BNA Dataset}

{\it Bank-note Authentication}~\cite{UCI, banknote} is a binary classification dataset. The classes involve Genuine and Forgery. A Genuine class refers to an authentic banknote denoted by `$0$', while a Forgery class refers to a forged banknote denoted by `$1$'. The obtained dataset is from images of banknotes belonging to both classes, taken from an industrial camera. It contains $1372$ total data instances. The dataset has four attributes retrieved from the images using wavelet transformation. The specified class distribution provided in Table~\ref{table:test-train split} is as follows: (Genuine, Forgery).

\subsection{Haberman's Survival} \label{subsec:Haberman Dataset}

{\it Haberman's Survival}~\cite{UCI, haberman_original} is a compilation of sections of a study investigating the lifespan of a patient after undergoing a breast cancer surgery. This is a binary classification problem and the dataset provides information for the prediction of the survival of patients beyond five years. Class `$1$' denotes survival of the patient for five years or longer after the surgery. Class `$2$' denotes the death of a patient within five years of the surgery. It contains $306$ total data instances and three attributes. The specified class distribution provided in Table~\ref{table:test-train split} is as follows: $(1, 2)$.

\subsection{Breast Cancer Wisconsin} \label{subsec:BCWD Dataset}

{\it Breast Cancer Wisconsin}~\cite{UCI, bcwd_original} dataset deals with the classification of the intensity of the breast cancer. Class `M' refers to a malignant level of infection and class `B' refers to a benign level of infection. It contains a total of $569$ data instances and $31$ attributes such as radius, perimeter, texture, smoothness, etc. for each cell nucleus. The specified class distribution provided in Table~\ref{table:test-train split} is as follows: (Malignant - M, Benign - B).

\subsection{Statlog (Heart)} \label{subsec:Statlog Dataset}

{\it Statlog (Heart)}~\cite{UCI} enables differentiation between presence and absence of a heart disease in a patient. Class `$1$' denotes the absence while class `$2$' denotes the presence of a heart disease. It contains $270$ total data instances and $13$ attributes including resting blood pressure, chest pain type, exercise induced angina and so on. The specified class distribution provided in Table~\ref{table:test-train split} is as follows: (Absence , Presence).

\subsection{Seeds} \label{subsec:Seeds Dataset}

{\it Seeds}~\cite{UCI} dataset examines three classes of wheat: Kama, Rosa and Canadian using soft X-ray on the wheat kernels to retrieve relevant properties. It contains $210$ total data instances and seven attributes namely compactness, length, width etc. of each wheat kernel. The specified class distribution provided in Table~\ref{table:test-train split} is as follows: (Kama, Rosa, Canadian).

\subsection{Free Spoken Digit Dataset} \label{subsec:FSDD Dataset}
            
The {\it Free Spoken Digit Dataset}~\cite{FSDD} is a time-series dataset comprising recordings of six speakers. Each speaker recites numbers from one to nine. For each number, every speaker makes $50$ recordings. The speaker chosen for all experiments is Jackson. The dataset undergoes preprocessing using a Fast Fourier Transform (FFT) technique. The dataset for speaker Jackson has $500$ data instances, and only instances above a threshold of $3000$ samples are considered to tackle the varying data length through the dataset. In these data instances, only the first $3005$ data samples are examined. Finally, $480$ data instances are filtered to feed into the algorithm. The specified class distribution provided in Table~\ref{table:test-train split} is as follows: $(0, 1, 2, 3, 4, 5, 6, 7, 8, 9)$.

\begin{table}[!ht]
\centering
\caption{Train-Test split in experiments. High training sample regime corresponds to an $80\%$ and $20\%$ split in training and testing respectively. In the Imbalanced data column, `Y' implies yes and `N' implies no. }
\begin{tabular}{|l|l|l|l|l|l|}
\hline
\textbf{Dataset} &
  \textbf{Classes} &
  \textbf{Features} &
  \textbf{\begin{tabular}[c]{@{}l@{}}Training \\ samples\\/class\end{tabular}} &
  \textbf{\begin{tabular}[c]{@{}l@{}}Testing \\ samples\\/class\end{tabular}} & \textbf{\begin{tabular}[c]{@{}l@{}}Imbalanced \\ data (Y/N)\end{tabular}}\\ \hline
Iris                                                                & 3 & 4  & (40, 41, 39) & (10, 9, 11) & N \\ \hline
Ionosphere                                                          & 2 & 34 & (98, 182)    & (28, 43)& Y  \\ \hline
Wine                                                               & 3 & 13 & (45, 57, 40) & (14, 14, 8) & Y \\ \hline
\begin{tabular}[c]{@{}l@{}}Bank Note \\ Authentication\end{tabular} & 2 & 4  & (614, 483)   & (148, 127)  & Y \\ \hline
\begin{tabular}[c]{@{}l@{}}Haberman's\\ Survival\end{tabular}                                                & 2 & 3  & (181, 63)    & (44, 18)  & Y   \\ \hline
\begin{tabular}[c]{@{}l@{}}Breast Cancer \\ Wisconsin\end{tabular}  & 2  & 31   & (169, 286) & (43, 71)    & Y         \\ \hline
Statlog (Heart)                                                     & 2 & 13 & (117, 99)    & (33, 21)  & Y   \\ \hline
Seeds                                                              & 3 & 7  & (59, 56, 53) & (11, 14, 17) & N \\ \hline
FSDD &
  10 &
  3005 &
  \begin{tabular}[c]{@{}l@{}}(40, 35, 44, 42,\\ 38, 34, 37, 44,\\ 33, 37)\end{tabular} &
  \begin{tabular}[c]{@{}l@{}}(10, 15, 6, 8,\\ 8, 7, 13, 6,\\ 10, 13)\end{tabular} & Y\\ \hline
\end{tabular}
\label{table:test-train split}
\end{table}

\section{Experiments \& Results}
\label{sec:Experiments & Results}

In this section, the authors evaluate the efficacy of ChaosFEX (CFX) feature engineering on various classification tasks. For this, hybrid models are developed by combining CFX features extracted from the input data with the classical ML algorithms such as Decision Tree, Random Forest, AdaBoost, Support Vector Machine, $k$-Nearest Neighbors and Gaussian Naive Bayes. The experiments are performed for both low and high training sample regime. The train-test distribution $(80\%-20\%)$ for each dataset in the high training sample regime is available in Table~\ref{table:test-train split}. In low training sample regime, $150$ random trials for training with $1,2, \ldots, 9$ data instances per class are considered. The trends of all algorithms in the stand-alone form and their implementation using CFX features are conveyed in this section.
For all experiments in the paper, the following software:
Python $3.8$ and scikit-learn~\cite{scikit-learn}, CFX~\cite{balakrishnan2019chaosnet} are used.

\subsection{Hyperparameter Tuning}

Every ML algorithm has a set of optimal hyperparameters to be found by hyperparameter tuning. The hyperparameters for all the algorithms with respect to each dataset were tuned using five-fold cross-validation. Table~\ref{Table_Hyperparameter-Tuning} provides the set of hyperparameters tuned for all algorithms in this research.

\begin{table}[!ht]
\centering
\caption{References for the tuned hyperparameters for all algorithms. Owing to space constraints,  most of the tables are moved to the Supplementary Information section (section~\ref{sec:Supplementary Information}). }
\begin{tabular}{|l|l|l|l|}
\hline
\textbf{Algorithm} & \textbf{Acronyms} &
  \textbf{Hyperparameters Tuned} &
  \textbf{Reference} \\ \hline
\verb|ChaosNet| & - &
  $q$, $b$, $\epsilon$ & \begin{tabular}[c]{@{}l@{}}Tables~\ref{table:Iris-HPT},~\ref{table:Ionosphere-HPT},~\ref{table:Wine-HPT},~\ref{table:BNA-HPT},~\ref{table:HS-HPT},~\ref{table:BCWD-HPT},~\ref{table:SH-HPT},~\ref{table:Seeds-HPT}and~\ref{table:FSDD-HPT}\end{tabular}
   \\ \hline
Decision Tree & DT &
  \begin{tabular}[c]{@{}l@{}}$min\_samples\_leaf$, \\ $max\_depth$,\\ $ccp\_alpha$\end{tabular} &
  \begin{tabular}[c]{@{}l@{}}Tables~\ref{table:Hyperparameter Tuning 1 - Decision Tree} and~\ref{table:Hyperparameter Tuning 2 - Decision Tree} in\\ Supplementary Information section\\ \end{tabular} \\ \hline
Random Forest & RF &
  \begin{tabular}[c]{@{}l@{}}$n\_estimators$,\\ $max\_depth$\end{tabular} &
  \begin{tabular}[c]{@{}l@{}}Tables~\ref{table:Hyperparameter Tuning 1 - Random Forest} and~\ref{table:Hyperparameter Tuning 2 - Random Forest} in\\ Supplementary Information section\end{tabular} \\ \hline
AdaBoost & AB &
  $n\_estimators$ &
  \begin{tabular}[c]{@{}l@{}}Table~\ref{table:Hyperparameter Tuning - AdaBoost} in\\ Supplementary Information section\end{tabular} \\ \hline
\begin{tabular}[c]{@{}l@{}}Support Vector \\ Machine\end{tabular} & SVM & 
  $C$, $kernel$ &
  \begin{tabular}[c]{@{}l@{}}Table~\ref{table:Hyperparameter Tuning - SVM} in\\ Supplementary Information section\end{tabular} \\ \hline
\begin{tabular}[c]{@{}l@{}}$k$-Nearest\\ Neighbors\end{tabular} & $k$-NN, KNN &
  $k$ &
  \begin{tabular}[c]{@{}l@{}}Table~\ref{table:Hyperparameter Tuning - $k$-NN} in \\ Supplementary Information section\end{tabular} \\ \hline
\begin{tabular}[c]{@{}l@{}}Gaussian Naive\\ Bayes\end{tabular} & GNB &
  - &  -\\
  \hline
\end{tabular}

\label{Table_Hyperparameter-Tuning}
\end{table}

In the case of CFX+ML for {\it Iris, Ionosphere} and {\it Wine}, both CFX $(q, b, \epsilon)$ and ML hyperparameters were tuned. For the remaining datasets, the hyperparameters tuned for \verb|ChaosNet| $(q, b, \epsilon)$ were retained and only the ML hyperparameters were tuned. 

\subsection{Performance Metric}

\begin{figure}[!ht]
\centering
\includegraphics[width=0.35\textwidth]{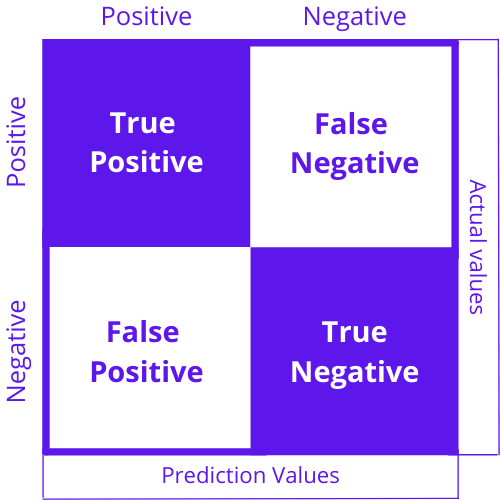}
\caption{Pictorial representation of a confusion matrix corresponding to a binary classification problem.}
\label{Fig_Confusion Matrix}
\end{figure}

The most common performance metric used in ML is \emph{Accuracy}~\cite{accuracy}. In the case of imbalanced datasets, this metric will lead to misinterpretation of results. Thus, the performance metric used for all experiments in this study is \textit{Macro F1-Score}. This metric is computed from the confusion matrix. Figure \ref{Fig_Confusion Matrix} shows a confusion matrix for a binary classification problem. Prediction values represent the predictions made by the classifier. Actual values stand for the true/ target values for the predictions being made by the classifier. They both have two categories: Positive and Negative. A true positive stands for a target value equal to positive, also correctly classified by the classifier as positive. True negatives are target values equal to negative, also correctly classified by the classifier as negative. False negative implies a target value equal to positive which was incorrectly classified as negative by the classifier. Similarly, false positive means a target value equal to negative that was incorrectly classified as positive by the classifier. 

F1-Score depends on two parameters: 
\begin{enumerate}
    \item Precision - The ratio of the number positives correctly classified as positive by the algorithm to the total number of instances classified as positive,  given by - 
    \begin{equation}
        Precision = \frac{True\:Positive}{True\:Positive + False\:Positive}. 
    \end{equation}
    \item Recall - The ratio of the number positives correctly classified as positive by the algorithm  to the total number of actual positives, given by -
    \begin{equation}
        Recall = \frac{True\:Positive}{True\:Positive + False\:Negative}.
    \end{equation}
\end{enumerate}

Thus, F1-Score which is the harmonic mean of Precision and Recall is given by,
\begin{equation}
    F1\mhyphen Score (F1) = \frac{2 \cdot Precision \cdot Recall}{Precision + Recall}.
\end{equation}

Macro F1-Score is obtained by averaging the F1 scores for all classes in the dataset, given by
\begin{equation}
    Macro\:F1\mhyphen Score = \frac{F1_{Class\mhyphen 1} + F1_{Class\mhyphen 2} + \ldots + F1_{Class\mhyphen n}}{n}, \label{eq:macro_f1}
\end{equation} 
where, $n$ stands for the number of distinct classes in the dataset. F1-score of $Class_k$ considers the instances of $Class_k$ as positive and all remaining instances of classes as negative. Thus, the task is transformed to a binary classification problem. All instances of $Class_k$ correctly classified as positive or negative by the classifier are termed as true positives or true negatives respectively. All instances that are incorrectly classified by the classifier are placed under the false positive and false negative category accordingly. Hence the associated equations of precision and recall for this example are given by,
\begin{equation}
    Precision_{Class_k}\: (P_k) = \frac{True\:Positive_{Class_k}}{True\:Positive_{Class_k} + False\:Positive_{Class_k}},
\end{equation} 
\begin{equation}
    Recall_{Class_k}\: (R_k) = \frac{True\:Positive_{Class_k}}{True\:Positive_{Class_k} + False\:Negative_{Class_k}}.
\end{equation}

The F1-score for $Class_k$ is computed by:
\begin{equation}
    F1\mhyphen Score_{Class_k} = \frac{2 \cdot P_k \cdot R_k}{P_k+R_k}.
\end{equation}

Similarly, the F1-scores for all classes in the classification problem are computed and applied in Eq~\ref{eq:macro_f1}.

\subsection{Performance of ChaosNet}\label{Performance of ChaosNet}
\verb|ChaosNet| has three hyperparameters -- initial neural activity $(q)$, discrimination threshold ($b$), and noise intensity ($\epsilon$)~\cite{harikrishnan2021noise}. Specifics of the same are provided in Table~\ref{Table_Hyperparameter-Tuning}.

\begin{table}[!ht]
\centering
\caption{ChaosNet results: High training sample regime results for the nine datasets used in the experiments.}
\begin{tabular}{|l|l|}
\hline
\textbf{Dataset}                                                    & \textbf{Macro-F1 Score (Test Data)} \\ \hline
Iris                                                                & 1.000                               \\ \hline
Ionosphere                                                          & 0.860                               \\ \hline
Wine                                                                & 0.976                               \\ \hline
\begin{tabular}[c]{@{}l@{}}Bank Note \\ Authentication\end{tabular} & 0.845                               \\ \hline
Haberman's Survival                                                 & 0.560                               \\ \hline
\begin{tabular}[c]{@{}l@{}}Breast Cancer \\ Wisconsin\end{tabular}  & 0.927                                    \\ \hline
Statlog (Heart)                                                     & 0.738                               \\ \hline
Seeds                                                               & 0.845                               \\ \hline
FSDD                                                                & 0.897                                    \\ \hline
\end{tabular}
\label{table:Experiment Results - ChaosNet}
\end{table}

\subsection{Comparative Performance Evaluation}
In this section, the authors represent the results in two formats (a) bar graph and (b) line graph. The comparative results for \verb|ChaosNet|, CFX+ML and stand-alone ML in the high training sample regime are depicted using bar graph. All values plotted in the bar graphs for each dataset are provided in the Supplementary Information section (section~\ref{sec:Supplementary Information}) from Table~\ref{table:Experiment Results - Decision Tree} -~\ref{table:Experiment Results - Gaussian Naive Bayes}. On the other hand, the line graph depicts the comparative performance of \verb|ChaosNet|, CFX+ML and stand-alone ML in the low training sample regime.

\subsubsection{Results for Iris} \label{Results for Iris}
\begin{figure*}[!ht]
\centering
\begin{subfigure}{0.99\textwidth}
\centering
\includegraphics[width=\textwidth]{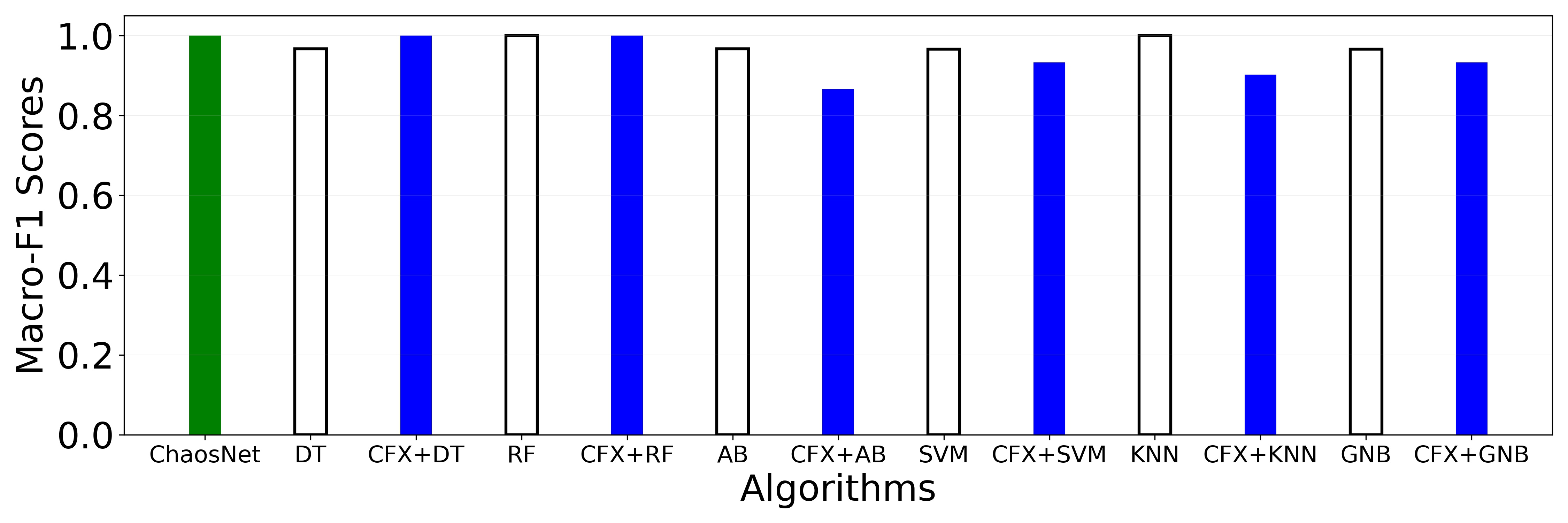}
\caption{}\label{Iris-HTS}
\end{subfigure}
\begin{subfigure}{0.49\textwidth}
\centering
\includegraphics[width=\textwidth]{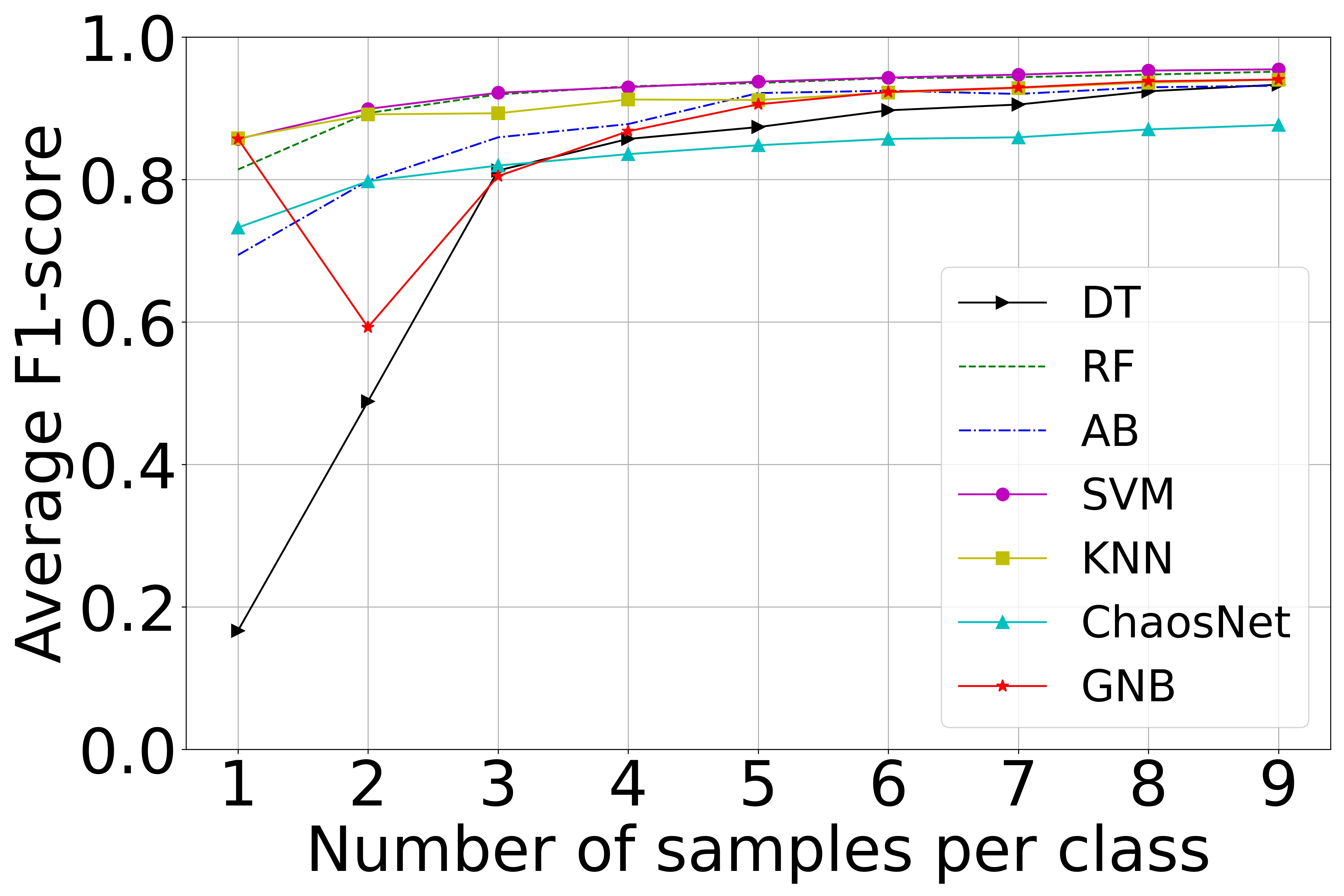}
\caption{}\label{Iris-LTS-SA}
\end{subfigure}
\hfill
\begin{subfigure}{0.49\textwidth}
\centering
\includegraphics[width=\textwidth]{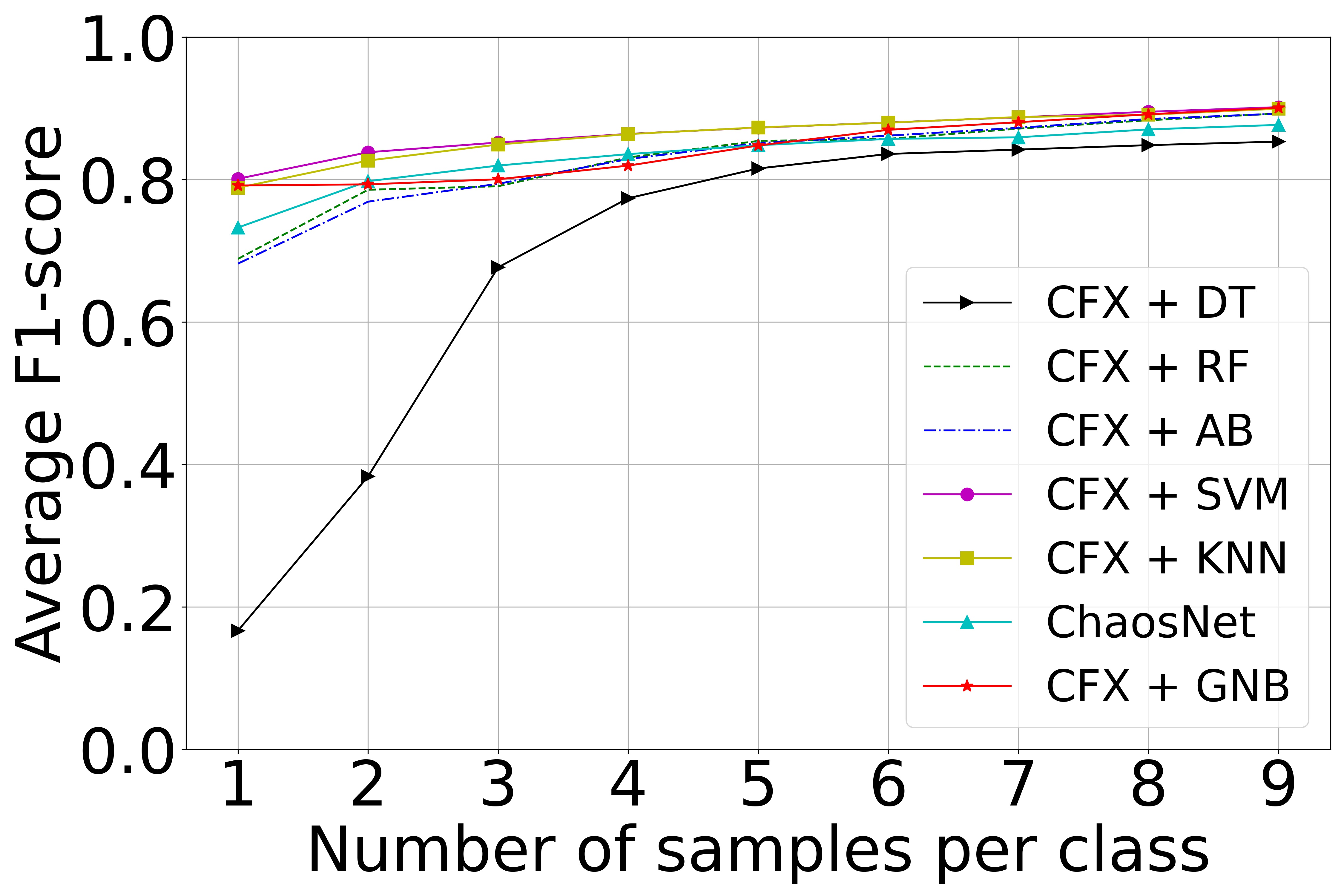}
\caption{}\label{Iris-LTS-CFX}
\end{subfigure}
\caption{{\it Iris}. (\subref{Iris-HTS}) High training sample regime. (\subref{Iris-LTS-SA}) Comparative performance of stand-alone algorithms in the low training sample regime. (\subref{Iris-LTS-CFX}) Comparative performance of CFX+ML algorithms in the low training sample regime.}
\label{Iris-RES}
\end{figure*}

The tuned hyperparameters used and all experiment results for the {\it Iris} dataset are available in Table~\ref{table:Iris-HPT} and Figure~\ref{Iris-RES} respectively.
\begin{table}[!ht]
\centering
\caption{Hyperparameters used for {\it Iris} dataset for high and low training sample regime experiments.}
\begin{tabular}{|l|l|}
\hline
\textbf{Hyperparameter} & \textbf{Tuned Value} \\ \hline
q                       & 0.141                \\ \hline
b                       & 0.499                \\ \hline
$\epsilon$              & 0.147                \\ \hline
\end{tabular}

\label{table:Iris-HPT}
\end{table}

\newpage

\subsubsection{Results for Ionosphere}
\begin{figure*}[!ht]
\centering
\begin{subfigure}{0.99\textwidth}
\centering
\includegraphics[width=\textwidth]{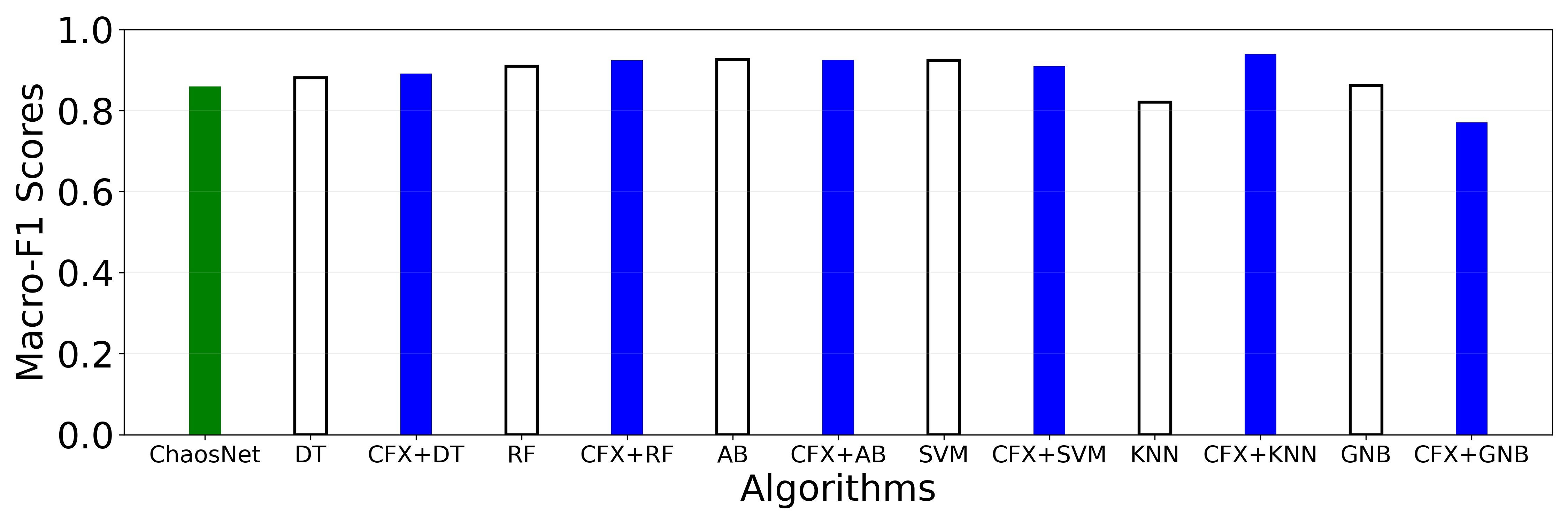}
\caption{}\label{Ionosphere-HTS}
\end{subfigure}
\begin{subfigure}{0.49\textwidth}
\centering
\includegraphics[width=\textwidth]{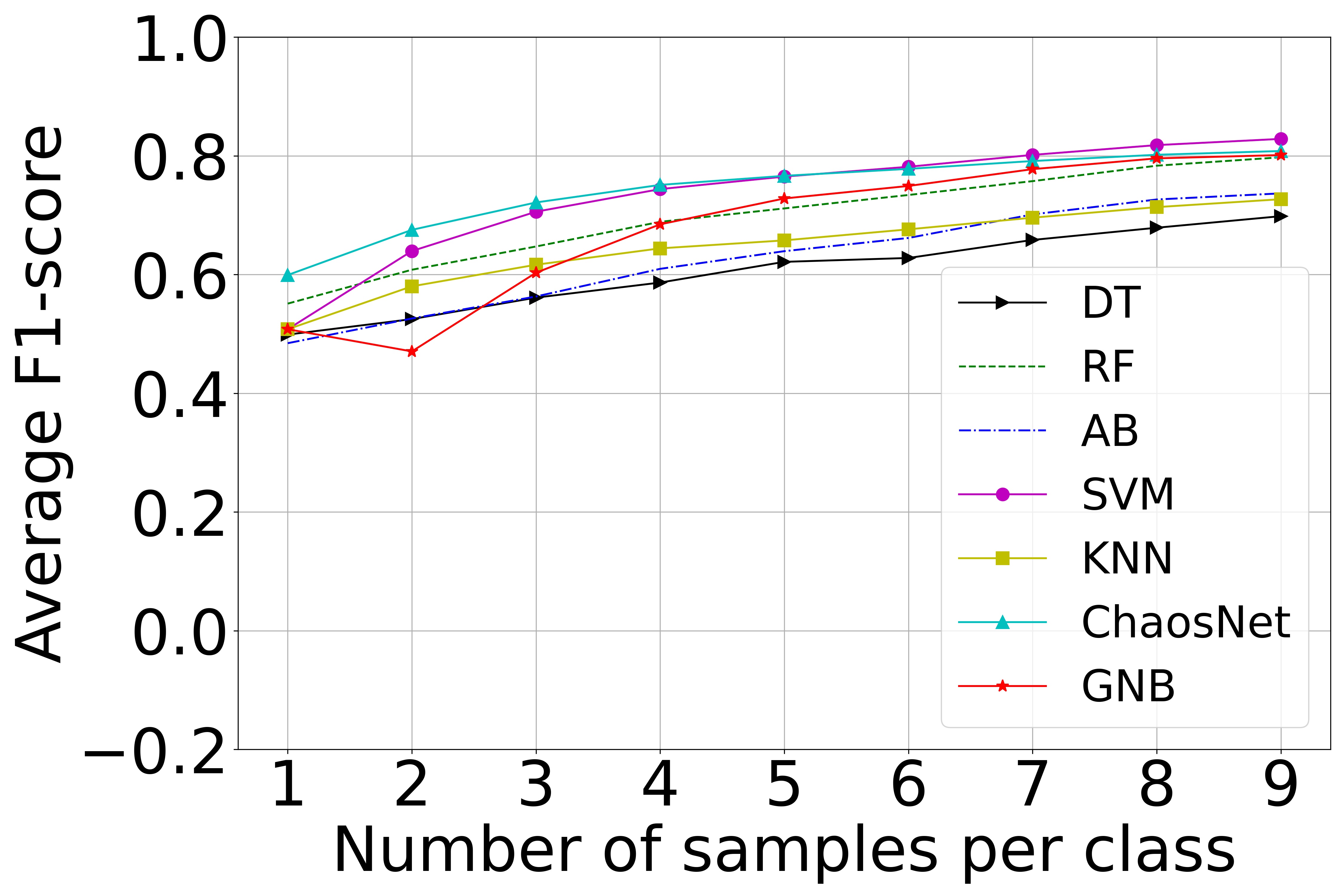}
\caption{}\label{Ionosphere-LTS-SA}
\end{subfigure}
\hfill
\begin{subfigure}{0.49\textwidth}
\centering
\includegraphics[width=\textwidth]{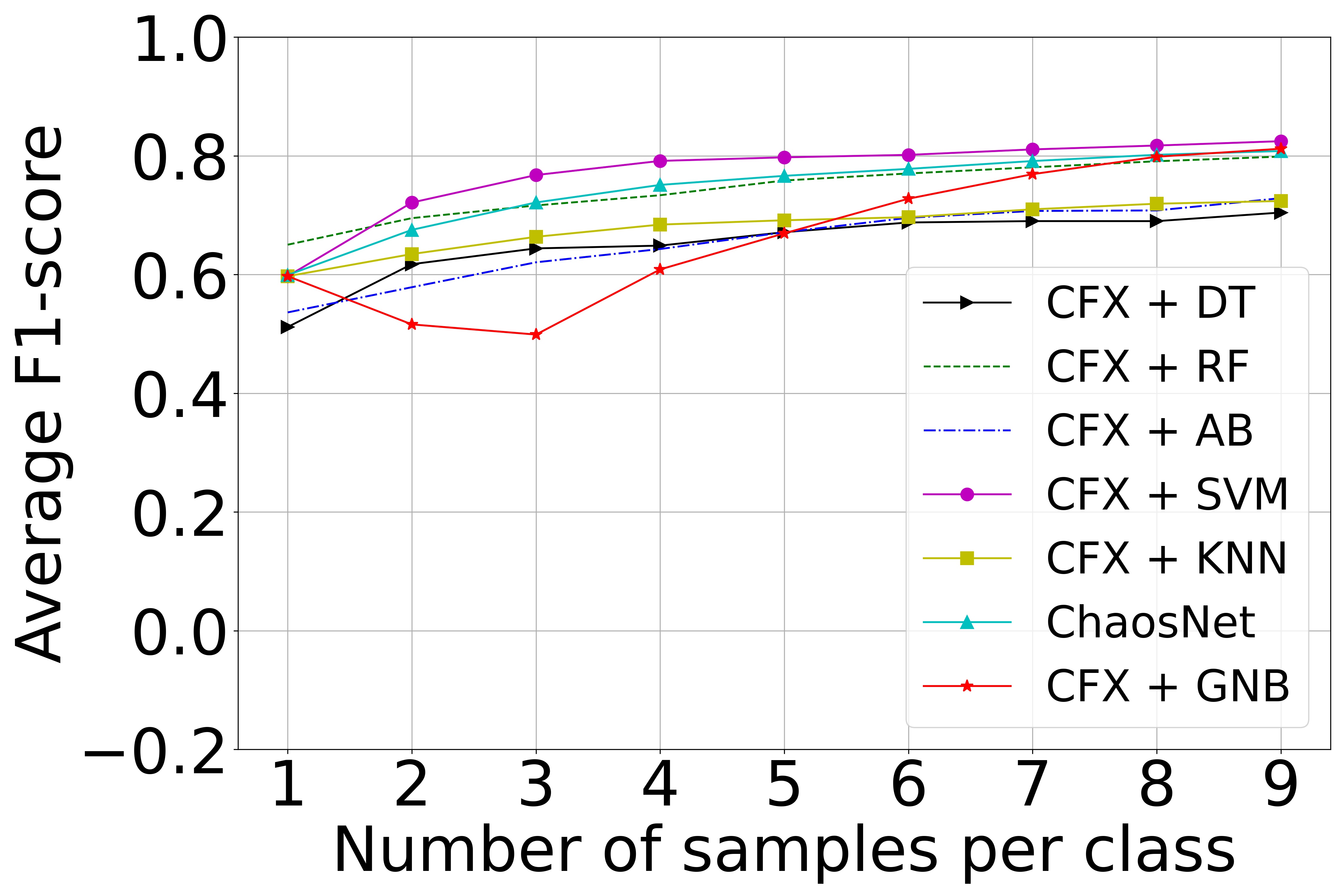}
\caption{}\label{Ionosphere-LTS-CFX}
\end{subfigure}
\caption{{\it Ionosphere.} (\subref{Ionosphere-HTS}) High training sample regime. (\subref{Ionosphere-LTS-SA}) Comparative performance of stand-alone algorithms in the low training sample regime. (\subref{Ionosphere-LTS-CFX}) Comparative performance of CFX+ML algorithms in the low training sample regime.}
\label{Ionosphere-RES}
\end{figure*}

The tuned hyperparameters used and all experiment results for the {\it Ionosphere} dataset are available in Table~\ref{table:Ionosphere-HPT} and Figure~\ref{Ionosphere-RES} respectively.
\begin{table}[!ht]
\centering
\caption{Hyperparameters used for {\it Ionosphere} dataset for high and low training sample regime experiments. }
\begin{tabular}{|l|l|}
\hline
\textbf{Hyperparameter} & \textbf{Tuned Value} \\ \hline
q                       & 0.680                \\ \hline
b                       & 0.969                \\ \hline
$\epsilon$              & 0.164                \\ \hline
\end{tabular}

\label{table:Ionosphere-HPT}
\end{table}
\newpage

\subsubsection {Results for Wine}
\begin{figure*}[!ht]
\centering
\begin{subfigure}{0.99\textwidth}
\centering
\includegraphics[width=\textwidth]{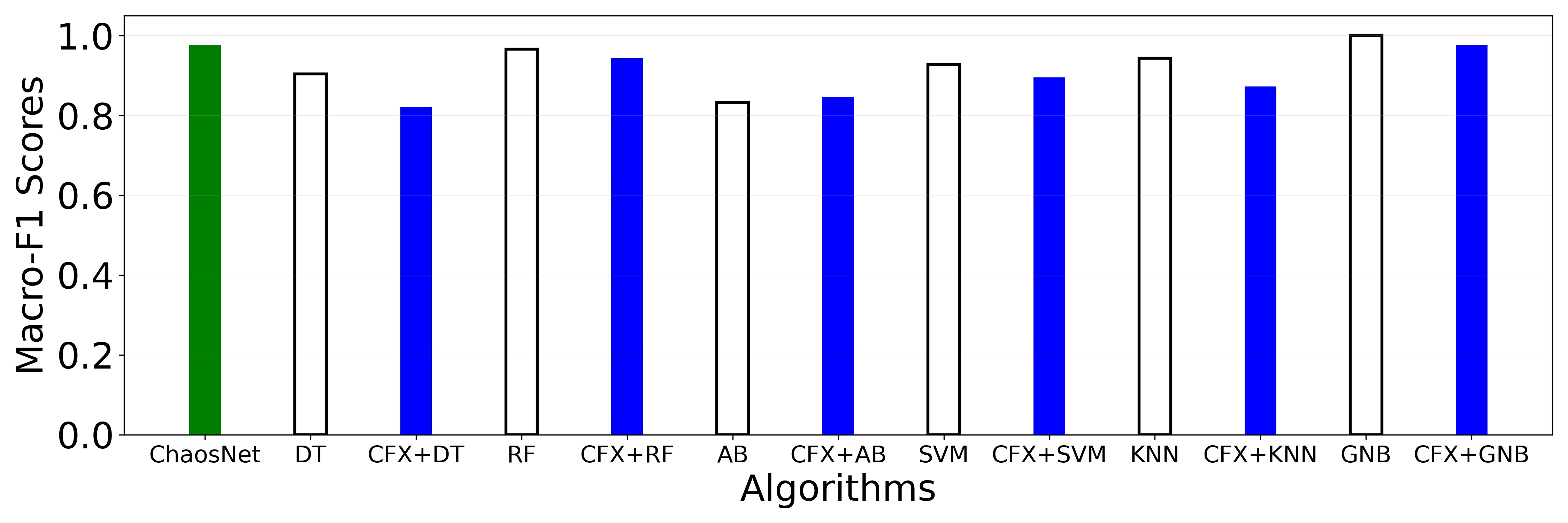}
\caption{}\label{Wine-HTS}
\end{subfigure}
\begin{subfigure}{0.49\textwidth}
\centering
\includegraphics[width=\textwidth]{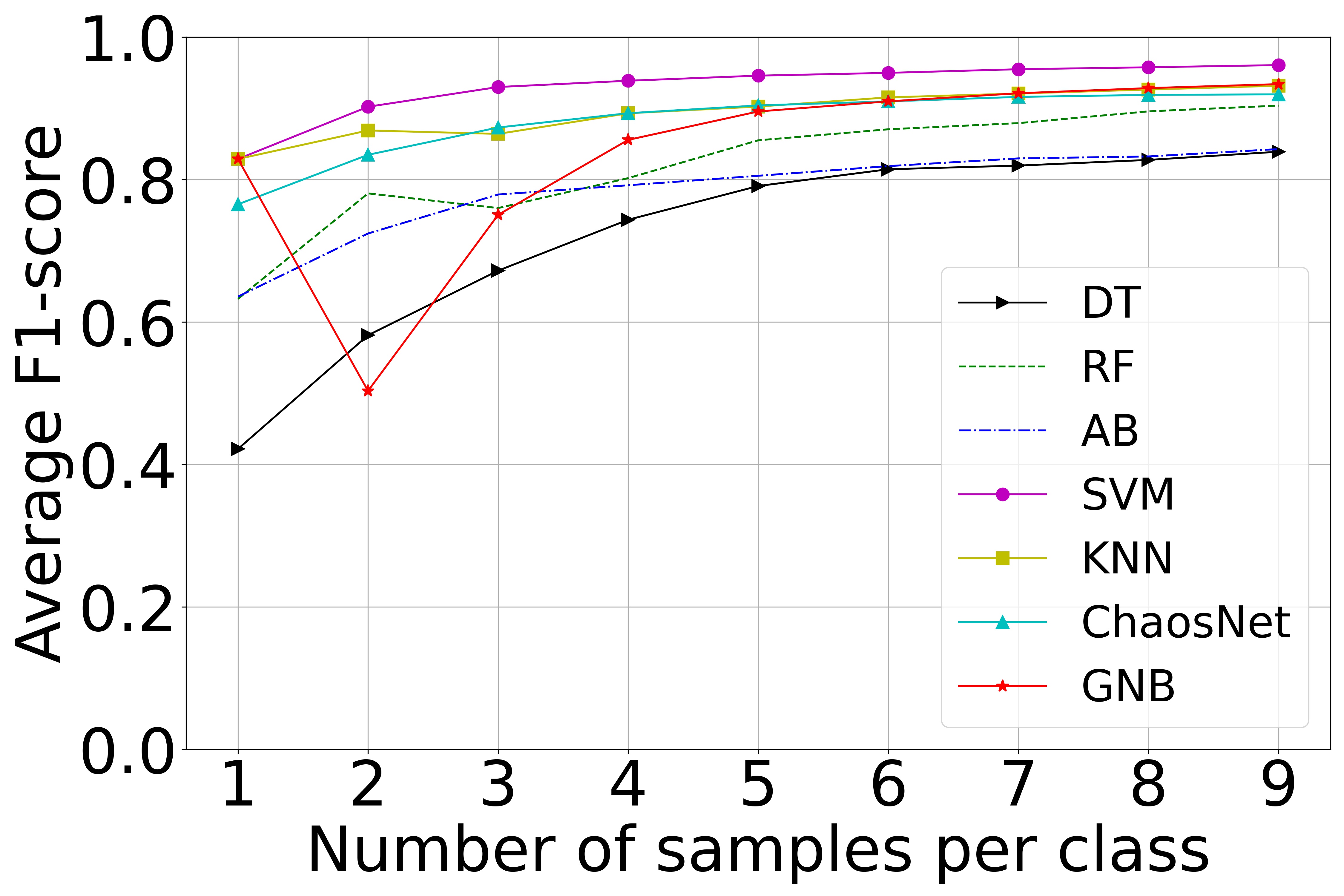}
\caption{}\label{Wine-LTS-SA}
\end{subfigure}
\hfill
\begin{subfigure}{0.49\textwidth}
\centering
\includegraphics[width=\textwidth]{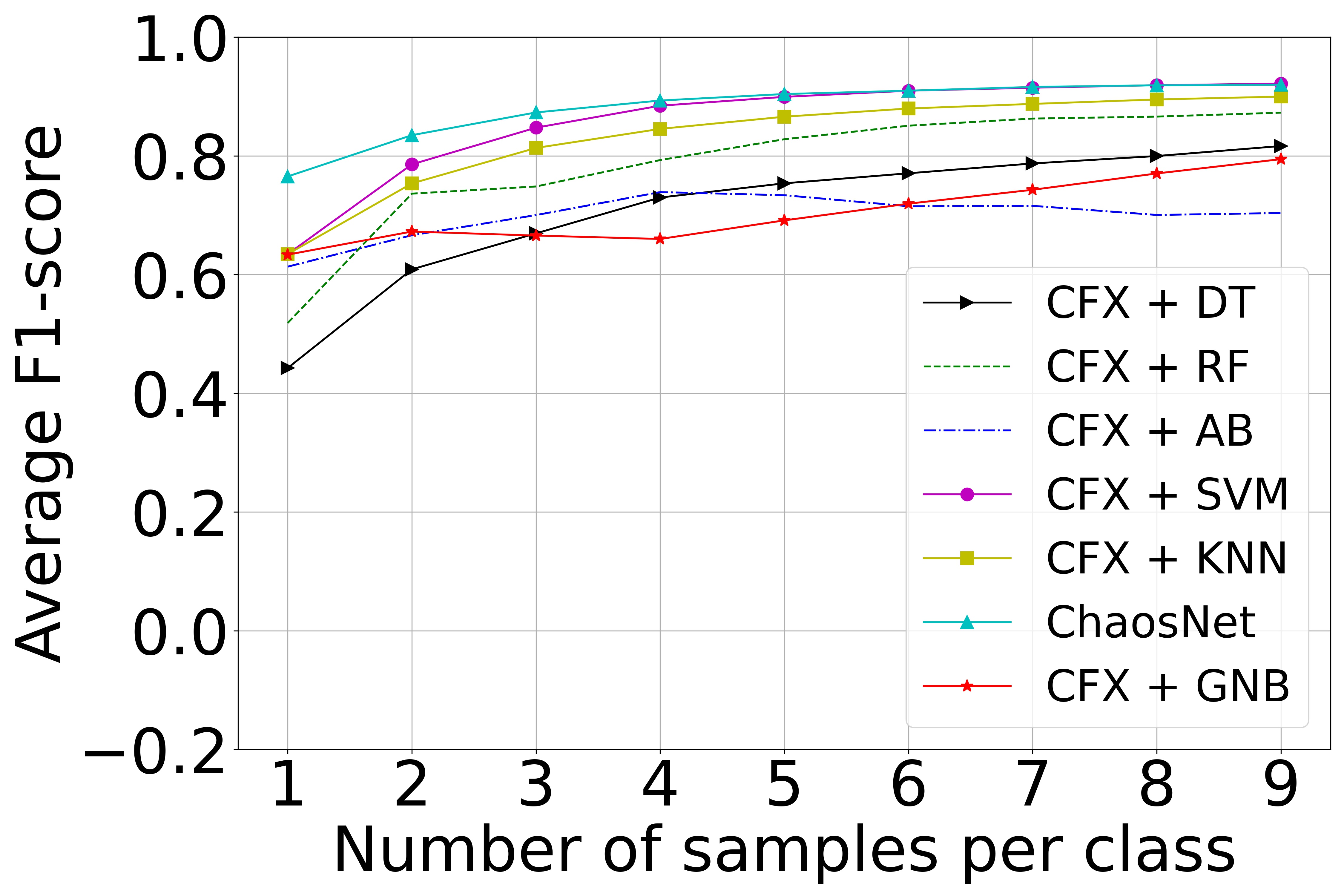}
\caption{}\label{Wine-LTS-CFX}
\end{subfigure}
\caption{{\it Wine. }(\subref{Wine-HTS}) High training sample regime. (\subref{Wine-LTS-SA}) Comparative performance of stand-alone algorithms in the low training sample regime.  (\subref{Wine-LTS-CFX}) Comparative performance of CFX+ML algorithms in the low training sample regime.}
\label{Wine-RES}
\end{figure*}

The tuned hyperparameters used and all experiment results for the {\it Wine} dataset are available in Table~\ref{table:Wine-HPT} and Figure~\ref{Wine-RES} respectively.
\begin{table}[!ht]
\centering
\caption{Hyperparameters used for {\it Wine} dataset for high and low training sample regime experiments. }
\begin{tabular}{|l|l|}
\hline
\textbf{Hyperparameter} & \textbf{Tuned Value} \\ \hline
q                       & 0.790                \\ \hline
b                       & 0.499                \\ \hline
$\epsilon$              & 0.262                \\ \hline
\end{tabular}

\label{table:Wine-HPT}
\end{table}

\newpage

\subsubsection{Results for Bank Note Authentication}
\begin{figure*}[!ht]
\centering
\begin{subfigure}{0.99\textwidth}
\centering
\includegraphics[width=\textwidth]{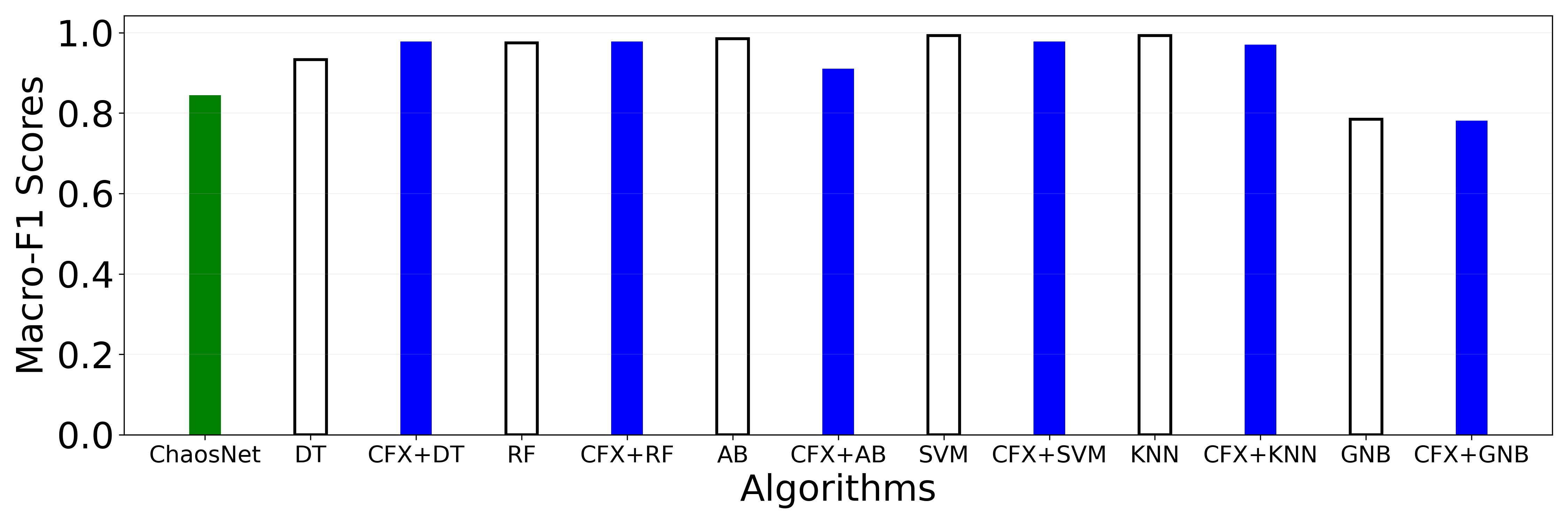}
\caption{}\label{BNA-HTS}
\end{subfigure}
\begin{subfigure}{0.49\textwidth}
\centering
\includegraphics[width=\textwidth]{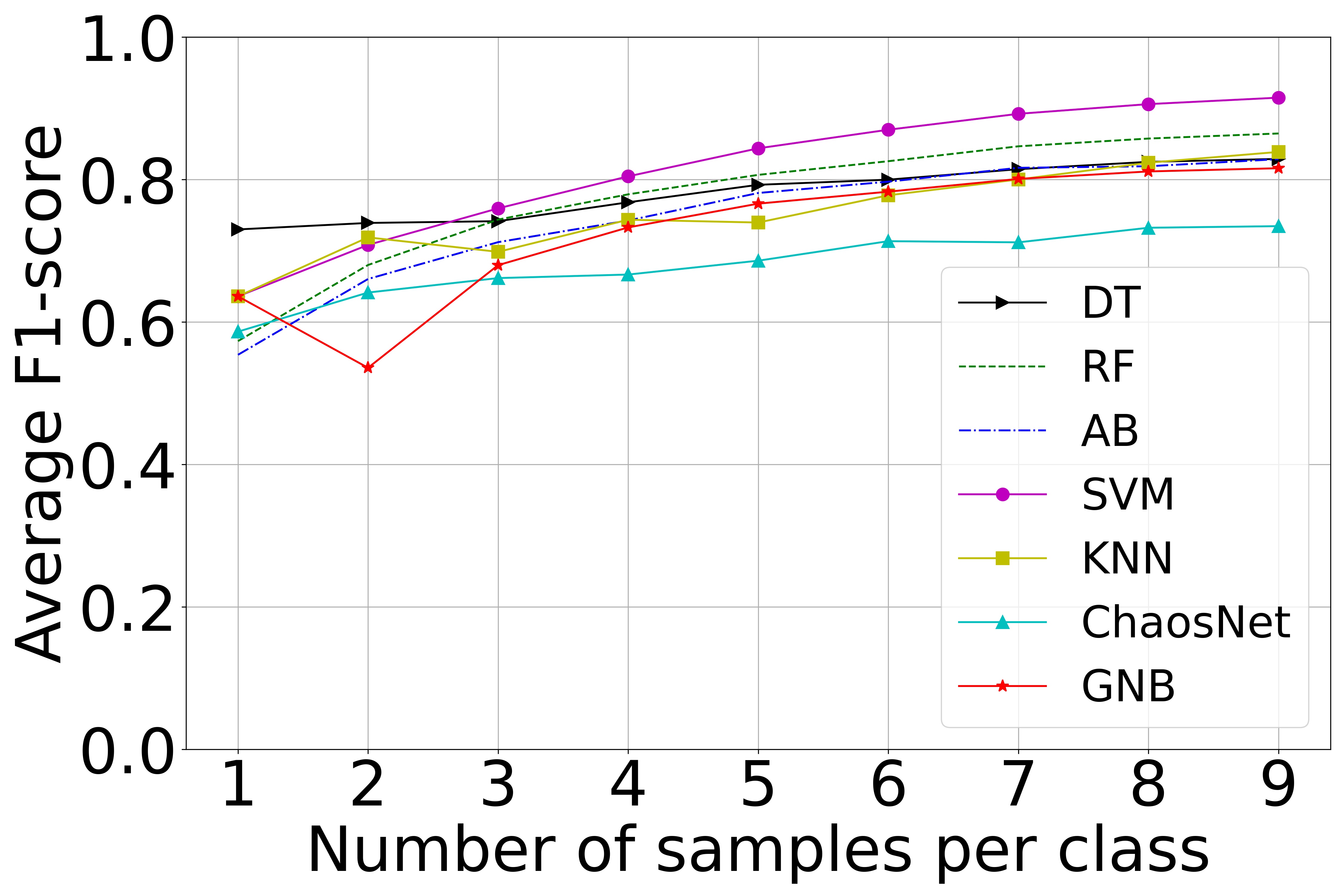}
\caption{}\label{BNA-LTS-SA}
\end{subfigure}
\hfill
\begin{subfigure}{0.49\textwidth}
\centering
\includegraphics[width=\textwidth]{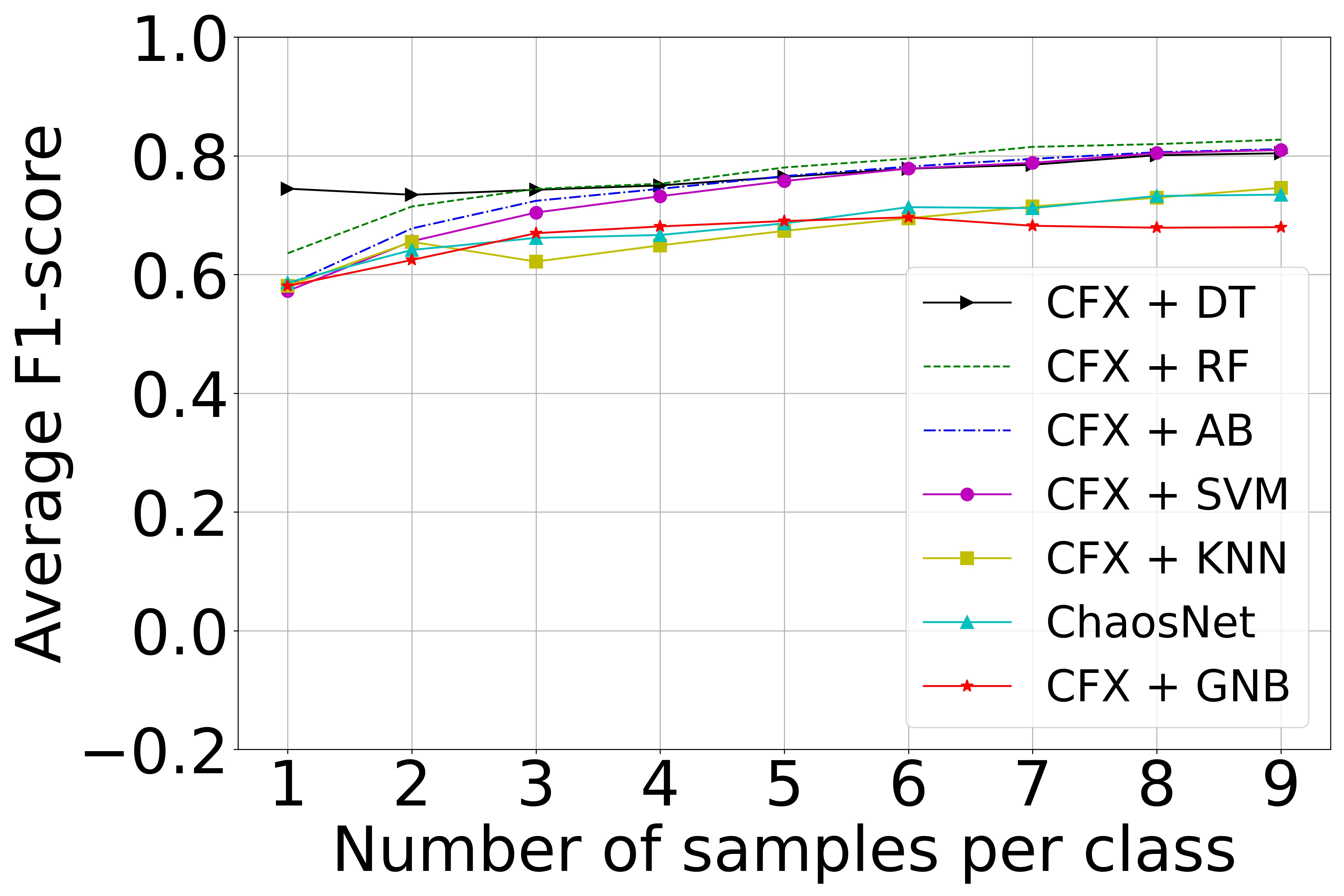}
\caption{}\label{BNA-LTS-CFX}
\end{subfigure}
\caption{{\it Bank Note Authentication. }(\subref{BNA-HTS}) High training sample regime. (\subref{BNA-LTS-SA}) Comparative performance of stand-alone algorithms in the low training sample regime.  (\subref{BNA-LTS-CFX}) Comparative performance of CFX+ML algorithms in the low training sample regime.}
\label{BNA-RES}
\end{figure*}

The tuned hyperparameters used and all experiment results for the {\it Bank Note Authentication} dataset are available in Table~\ref{table:BNA-HPT} and Figure~\ref{BNA-RES} respectively.
\begin{table}[!ht]
\centering
\caption{Hyperparameters used for {\it Bank Note Authentication} dataset for high and low training sample regime experiments. }
\begin{tabular}{|l|l|}
\hline
\textbf{Hyperparameter} & \textbf{Tuned Value} \\ \hline
q                       & 0.080                \\ \hline
b                       & 0.250                \\ \hline
$\epsilon$              & 0.233                \\ \hline
\end{tabular}

\label{table:BNA-HPT}
\end{table}
\newpage

\subsubsection{Results for Haberman's Survival}

\begin{figure*}[!ht]
\centering
\begin{subfigure}{0.99\textwidth}
\centering
\includegraphics[width=\textwidth]{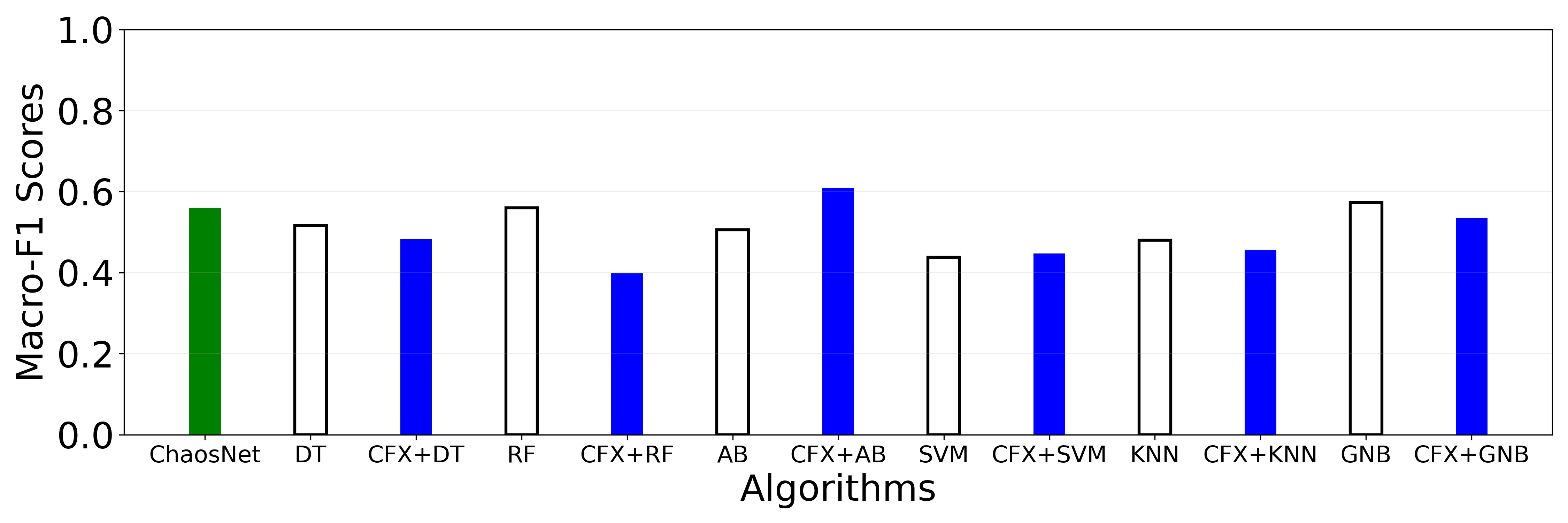}
\caption{}\label{HS-HTS}
\end{subfigure}
\begin{subfigure}{0.49\textwidth}
\centering
\includegraphics[width=\textwidth]{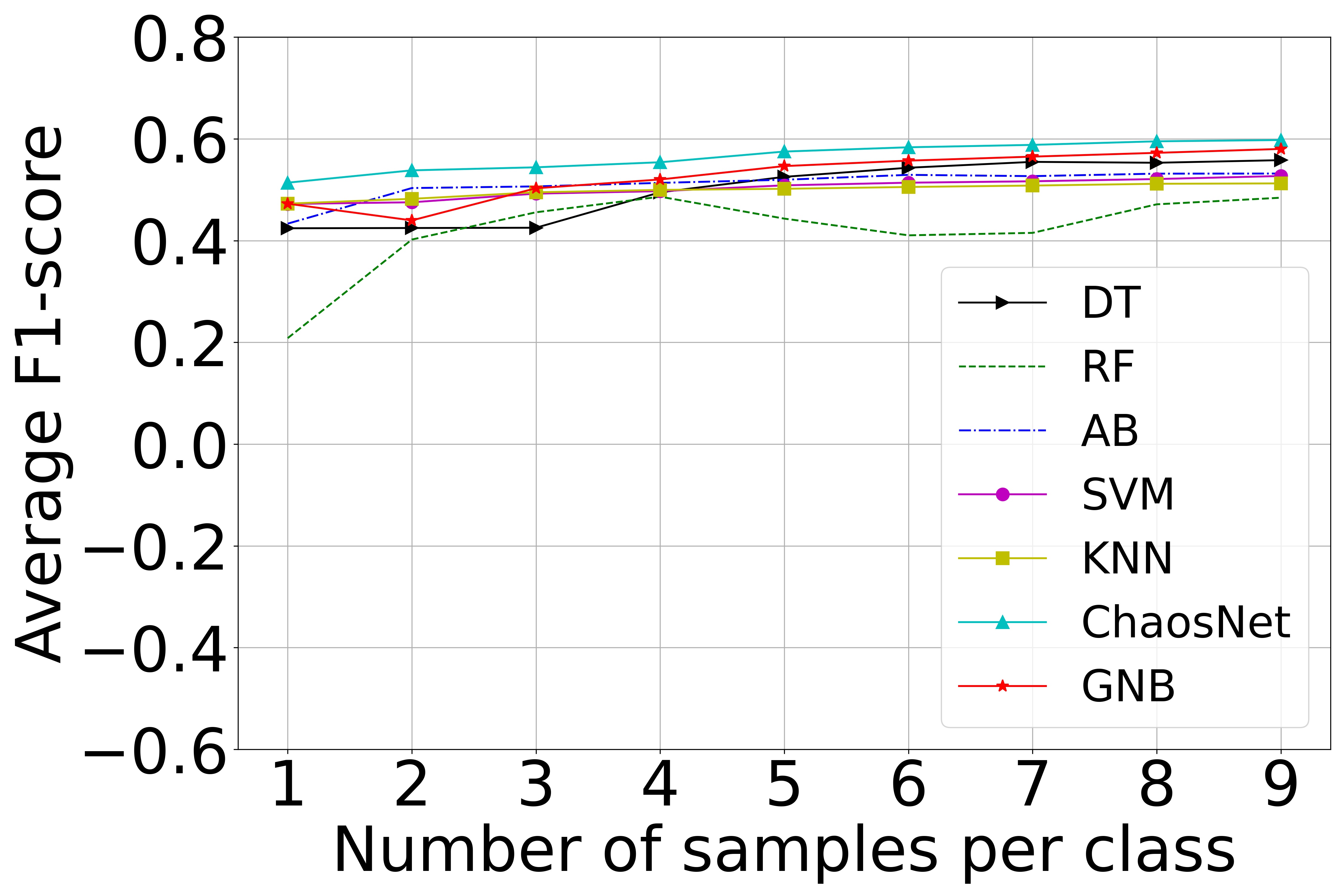}
\caption{}\label{HS-LTS-SA}
\end{subfigure}
\hfill
\begin{subfigure}{0.49\textwidth}
\centering
\includegraphics[width=\textwidth]{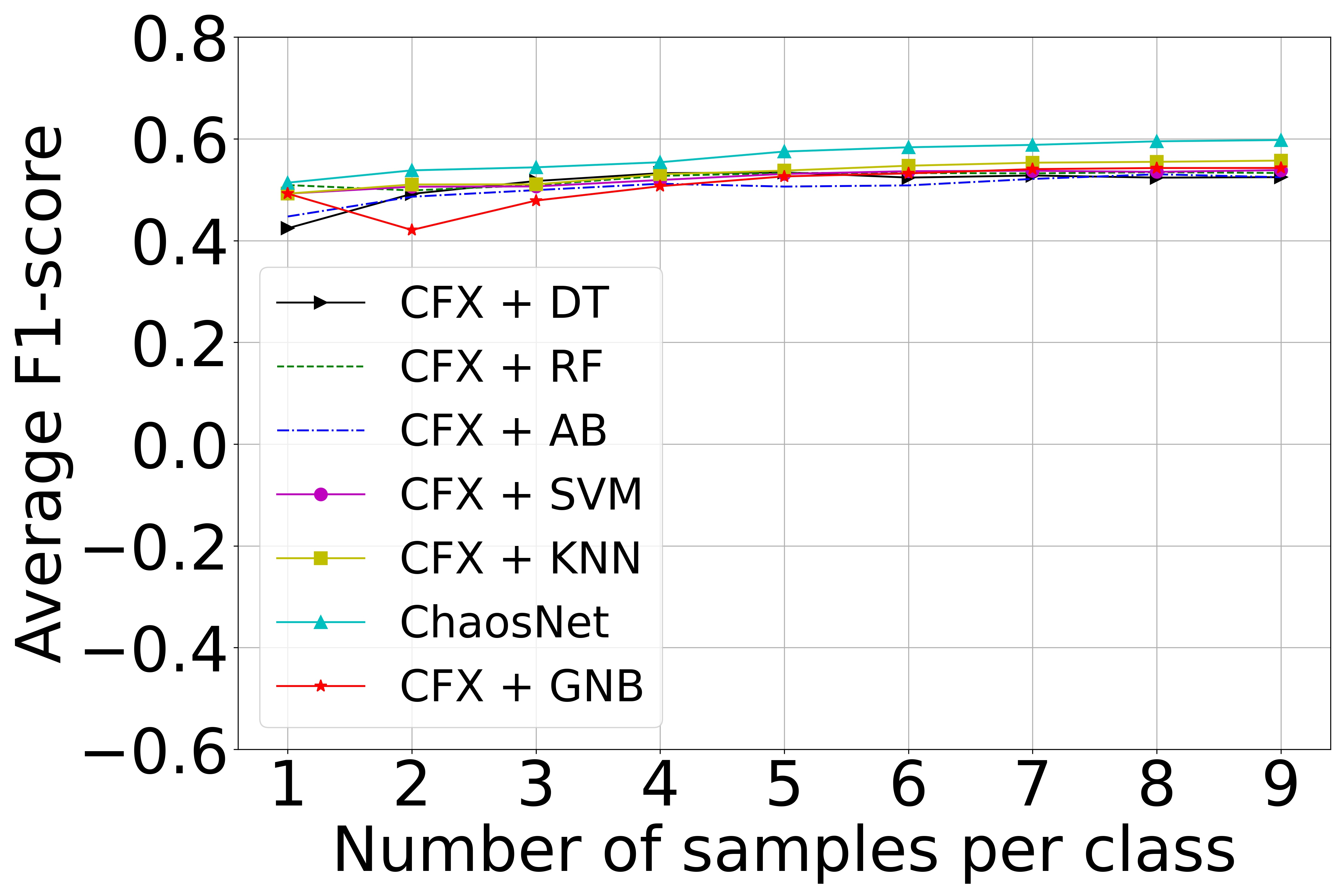}
\caption{}\label{HS-LTS-CFX}
\end{subfigure}
\caption{{\it Haberman's Survival. }(\subref{HS-HTS}) High training sample regime. (\subref{HS-LTS-SA}) Comparative performance of stand-alone algorithms in the low training sample regime. (\subref{HS-LTS-CFX}) Comparative performance of CFX+ML algorithms in the low training sample regime.}
\label{HS-RES}
\end{figure*}

The tuned hyperparameters used and all experiment results for the {\it Haberman's Survival} dataset are available in Table~\ref{table:HS-HPT} and Figure~\ref{HS-RES} respectively.
\begin{table}[!ht]
\centering
\caption{Hyperparameters used for {\it Haberman's Survival} dataset for high and low training sample regime experiments. }
\begin{tabular}{|l|l|}
\hline
\textbf{Hyperparameter} & \textbf{Tuned Value} \\ \hline
q                       & 0.810               \\ \hline
b                       & 0.140                \\ \hline
$\epsilon$              & 0.003                \\ \hline
\end{tabular}
\label{table:HS-HPT}
\end{table}

\newpage

\subsubsection{Results for Breast Cancer Wisconsin}

\begin{figure*}[!ht]
\centering
\begin{subfigure}{0.99\textwidth}
\centering
\includegraphics[width=\textwidth]{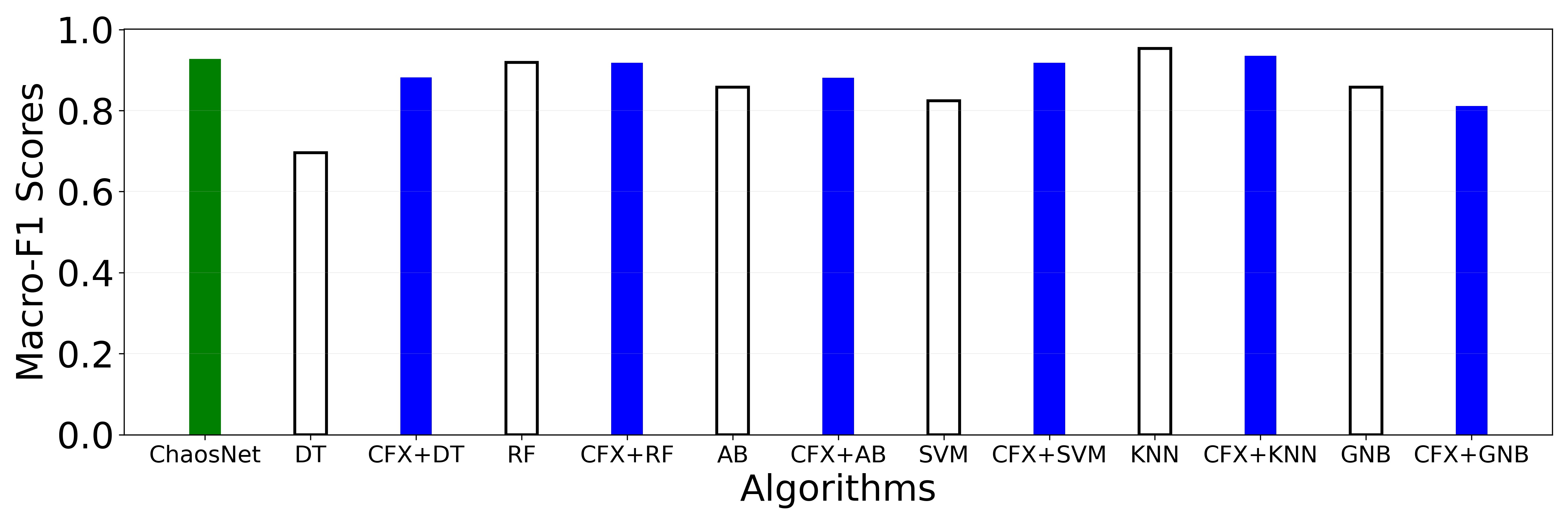}
\caption{}\label{BCWD-HTS}
\end{subfigure}
\begin{subfigure}{0.49\textwidth}
\centering
\includegraphics[width=\textwidth]{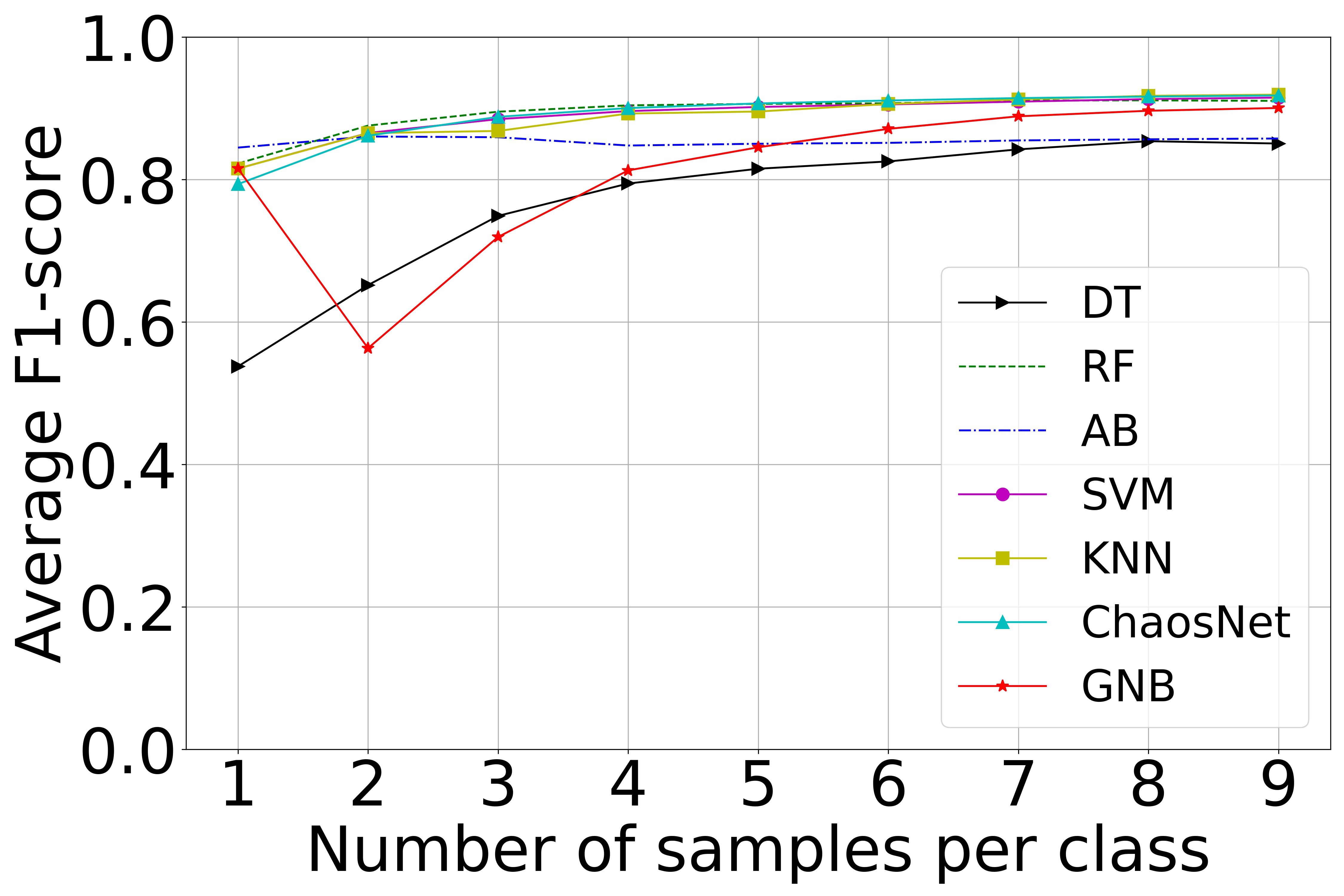}
\caption{}\label{BCWD-LTS-SA}
\end{subfigure}
\hfill
\begin{subfigure}{0.49\textwidth}
\centering
\includegraphics[width=\textwidth]{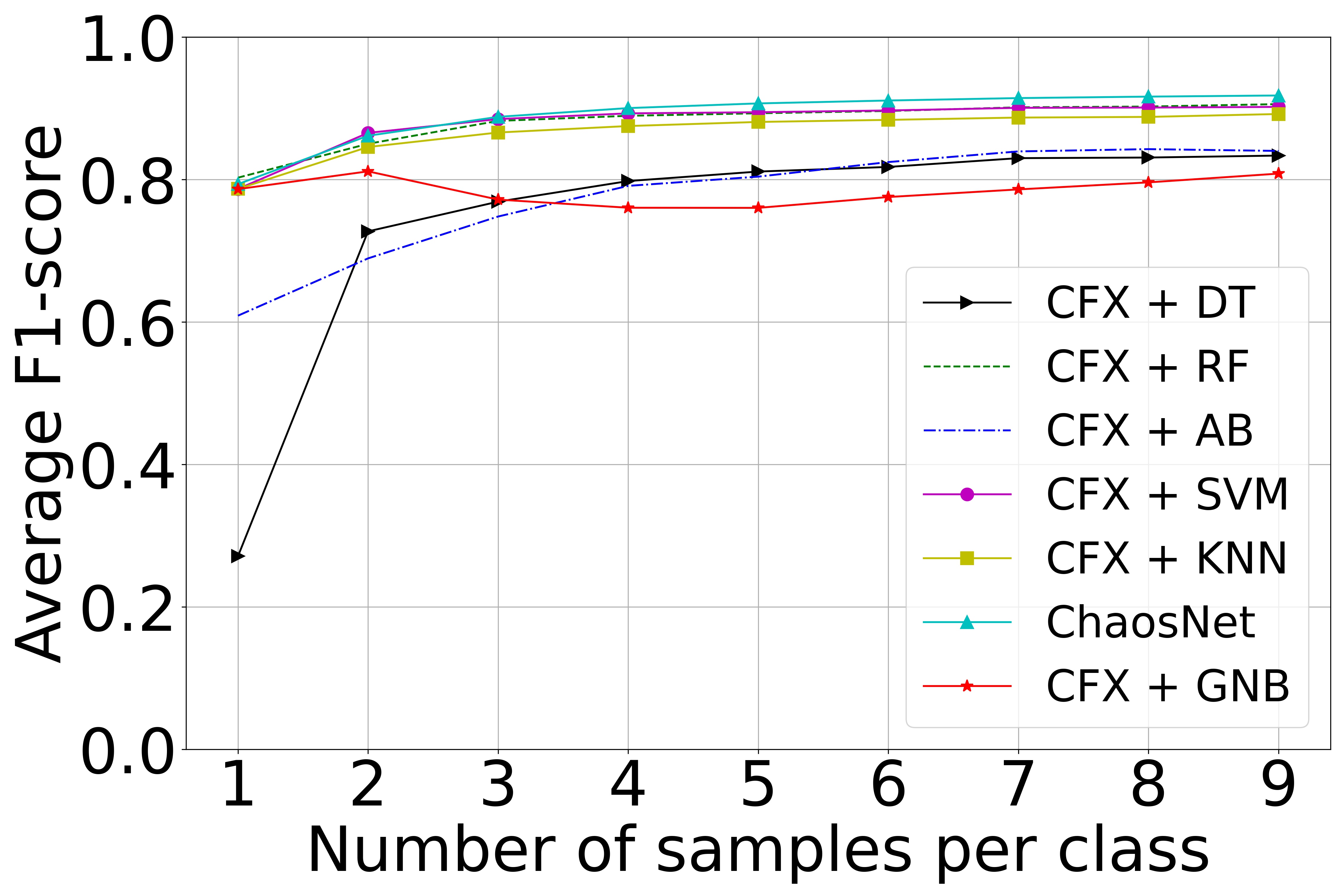}
\caption{}\label{BCWD-LTS-CFX}
\end{subfigure}
\caption{{\it Breast Cancer Wisconsin. }(\subref{BCWD-HTS}) High training sample regime. (\subref{BCWD-LTS-SA}) Comparative performance of stand-alone algorithms in the low training sample regime. (\subref{BCWD-LTS-CFX}) Comparative performance of CFX+ML algorithms in the low training sample regime.}
\label{BCWD-RES}
\end{figure*}

The tuned hyperparameters used and all experiment results for the {\it Breast Cancer Wisconsin} dataset are available in Table~\ref{table:BCWD-HPT} and Figure~\ref{BCWD-RES} respectively.
\begin{table}[!ht]
\centering
\caption{Hyperparameters used for {\it Breast Cancer Wisconsin} dataset for high and low training sample regime experiments. }
\begin{tabular}{|l|l|}
\hline
\textbf{Hyperparameter} & \textbf{Tuned Value} \\ \hline
q                       & 0.930               \\ \hline
b                       & 0.490                \\ \hline
$\epsilon$              & 0.159                \\ \hline
\end{tabular}

\label{table:BCWD-HPT}
\end{table}

\newpage

\subsubsection{Results for Statlog (Heart)}

\begin{figure*}[!ht]
\centering
\begin{subfigure}{0.99\textwidth}
\centering
\includegraphics[width=\textwidth]{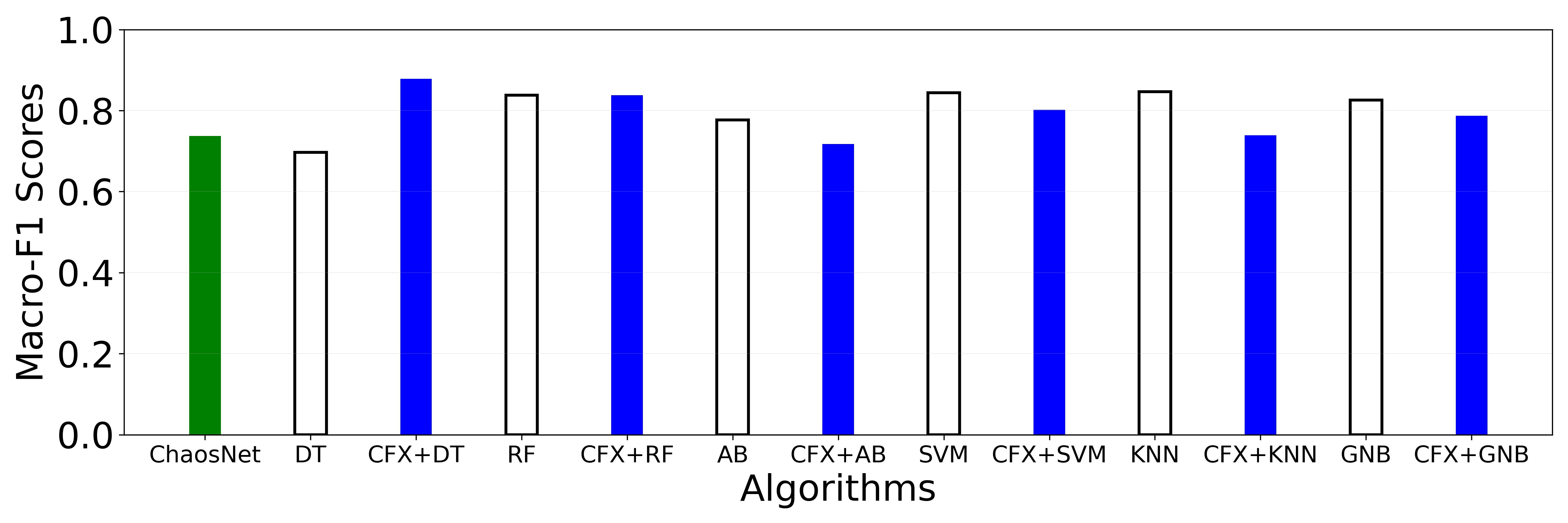}
\caption{}\label{SH-HTS}
\end{subfigure}
\begin{subfigure}{0.49\textwidth}
\centering
\includegraphics[width=\textwidth]{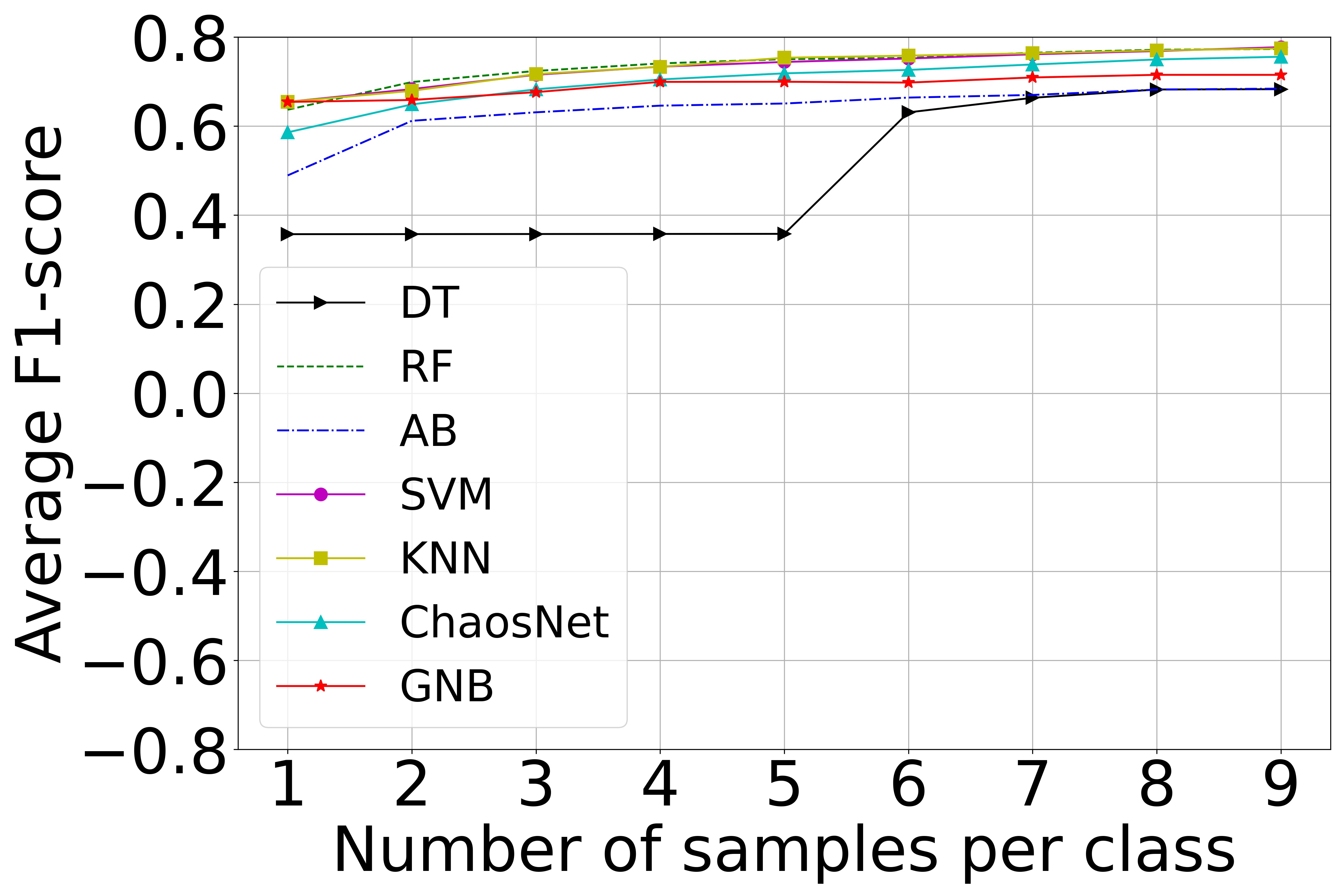}
\caption{}\label{SH-LTS-SA}
\end{subfigure}
\hfill
\begin{subfigure}{0.49\textwidth}
\centering
\includegraphics[width=\textwidth]{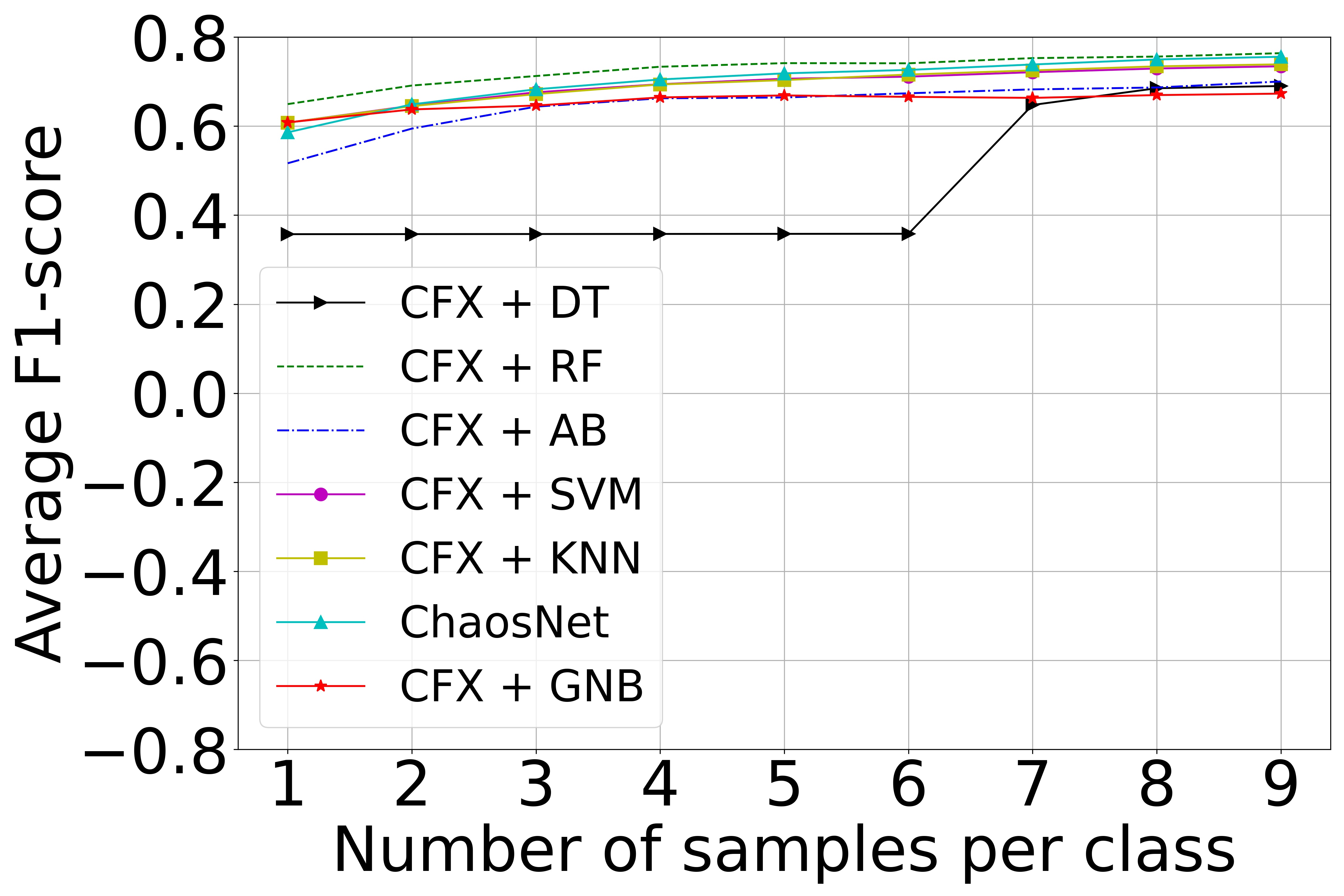}
\caption{}\label{SH-LTS-CFX}
\end{subfigure}
\caption{{\it Statlog (Heart). }(\subref{SH-HTS}) High training sample regime. (\subref{SH-LTS-SA}) Comparative performance of stand-alone algorithms in the low training sample regime. (\subref{SH-LTS-CFX}) Comparative performance of CFX+ML algorithms in the low training sample
regime.}
\label{SH-RES}
\end{figure*}

The tuned hyperparameters used and all experiment results for the {\it Statlog (Heart)} dataset are available in Table~\ref{table:SH-HPT} and Figure~\ref{SH-RES} respectively.
\begin{table}[!ht]
\centering
\caption{Hyperparameters used for {\it Statlog (Heart)} dataset for high and low training sample regime experiments. }
\begin{tabular}{|l|l|}
\hline
\textbf{Hyperparameter} & \textbf{Tuned Value} \\ \hline
q                       & 0.080               \\ \hline
b                       & 0.060                \\ \hline
$\epsilon$              & 0.170                \\ \hline
\end{tabular}

\label{table:SH-HPT}
\end{table}

\newpage

\subsubsection{Results for Seeds}

\begin{figure*}[!ht]
\centering
\begin{subfigure}{0.99\textwidth}
\centering
\includegraphics[width=\textwidth]{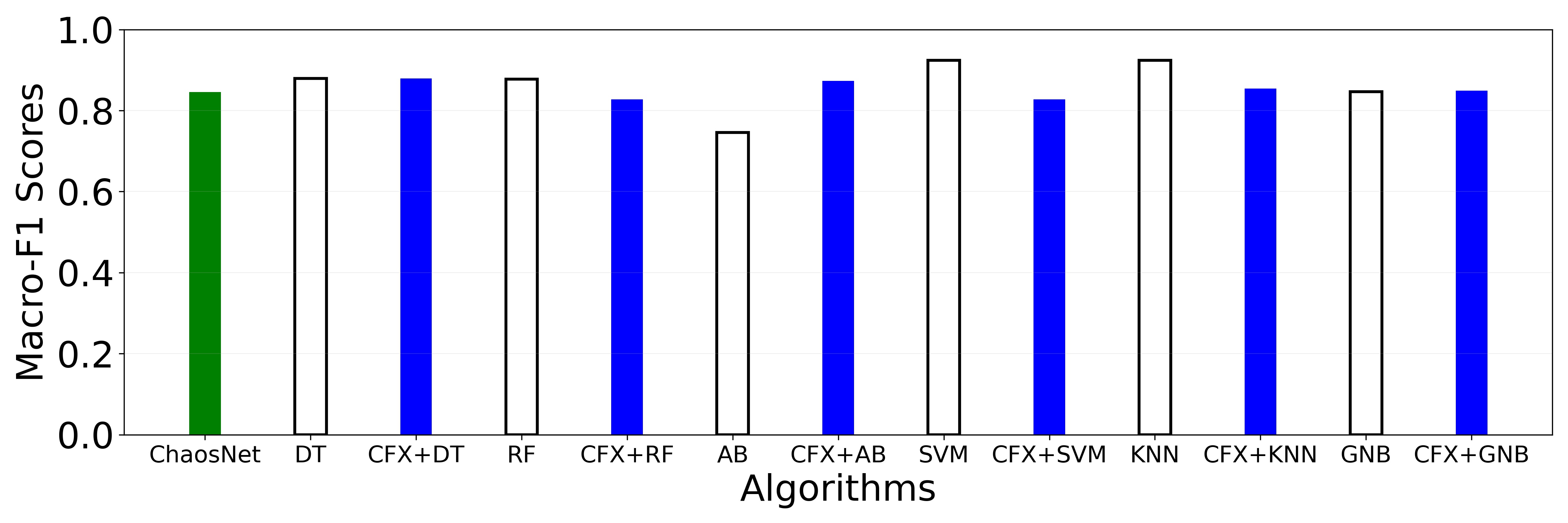}
\caption{}\label{Seeds-HTS}
\end{subfigure}
\begin{subfigure}{0.49\textwidth}
\centering
\includegraphics[width=\textwidth]{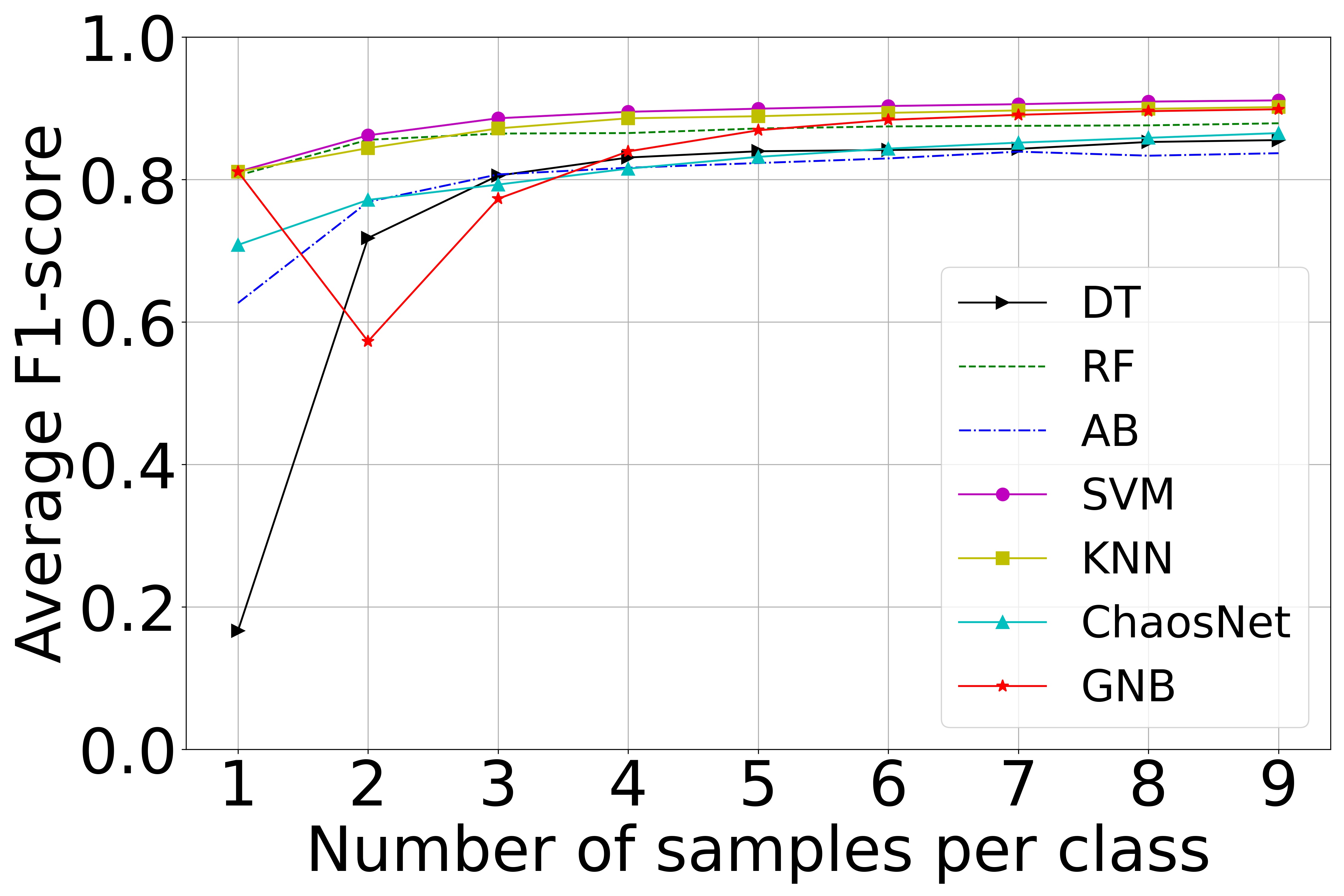}
\caption{}\label{Seeds-LTS-SA}
\end{subfigure}
\hfill
\begin{subfigure}{0.49\textwidth}
\centering
\includegraphics[width=\textwidth]{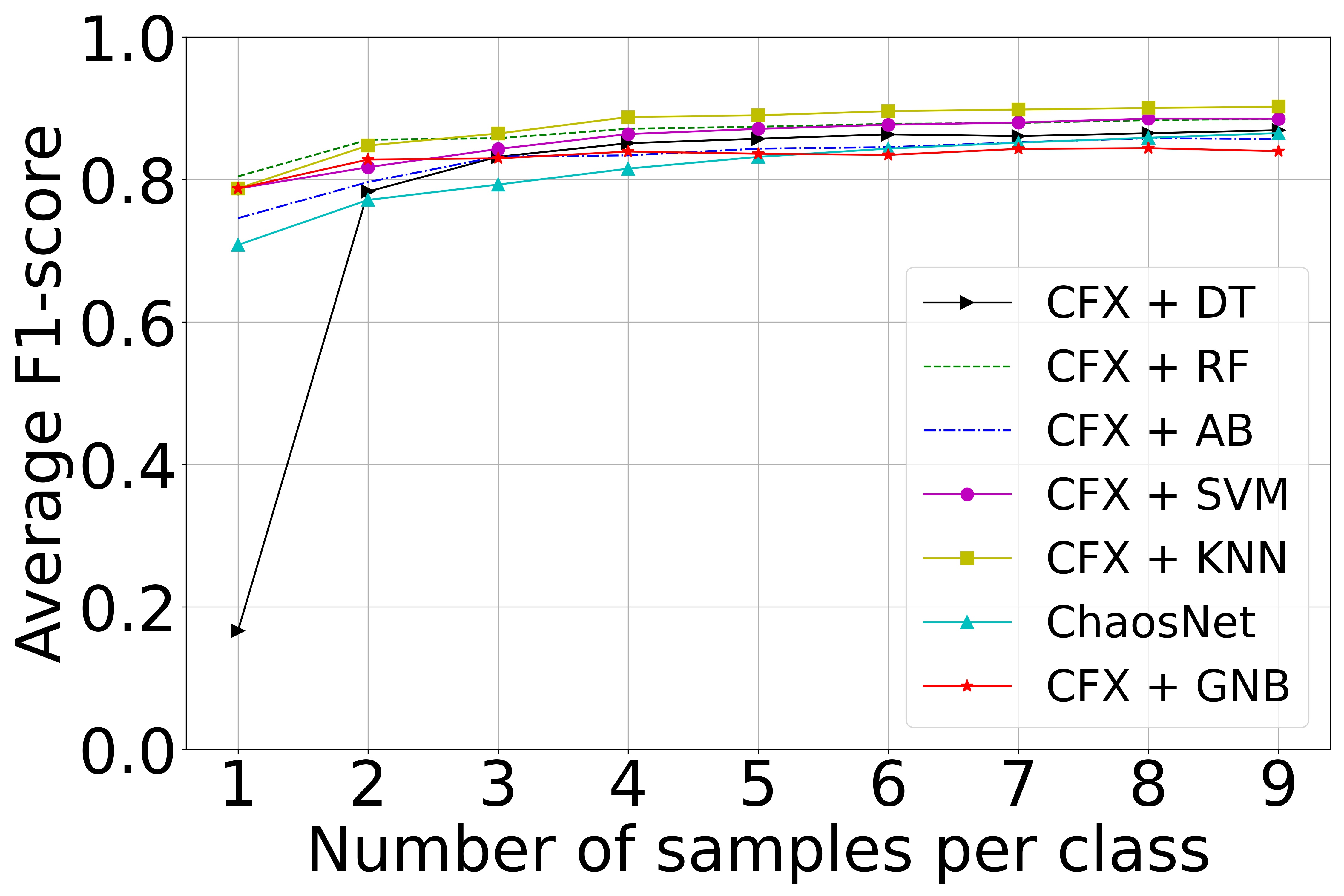}
\caption{}\label{Seeds-LTS-CFX}
\end{subfigure}
\caption{{\it Seeds. }(\subref{Seeds-HTS}) High training sample regime. (\subref{Seeds-LTS-SA}) Comparative performance of stand-alone algorithms in the low training sample regime. (\subref{Seeds-LTS-CFX}) Comparative performance of CFX+ML algorithms in the low training sample regime.}
\label{Seeds-RES}
\end{figure*}

The tuned hyperparameters used and all experiment results for the {\it Seeds} dataset are available in Table~\ref{table:Seeds-HPT} and Figure~\ref{Seeds-RES} respectively.
\begin{table}[!ht]
\centering
\caption{Hyperparameters used for {\it Seeds} dataset for high and low training sample regime experiments. }
\begin{tabular}{|l|l|}
\hline
\textbf{Hyperparameter} & \textbf{Tuned Value} \\ \hline
q                       & 0.020               \\ \hline
b                       & 0.070                \\ \hline
$\epsilon$              & 0.238                \\ \hline
\end{tabular}

\label{table:Seeds-HPT}
\end{table}

\newpage

\subsubsection{Results for Free Spoken Digit Dataset (FSDD)} \label{Results for FSDD}

\begin{figure*}[!ht]
\centering
\begin{subfigure}{0.99\textwidth}
\centering
\includegraphics[width=\textwidth]{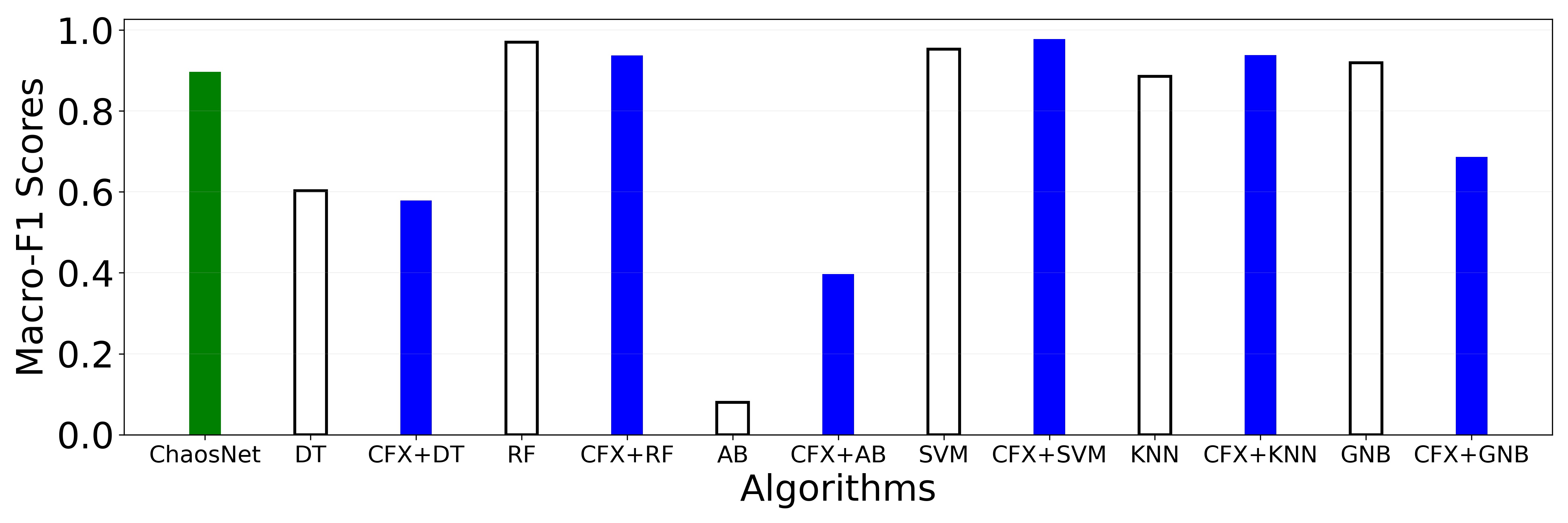}
\caption{}\label{FSDD-HTS}
\end{subfigure}
\begin{subfigure}{0.49\textwidth}
\centering
\includegraphics[width=\textwidth]{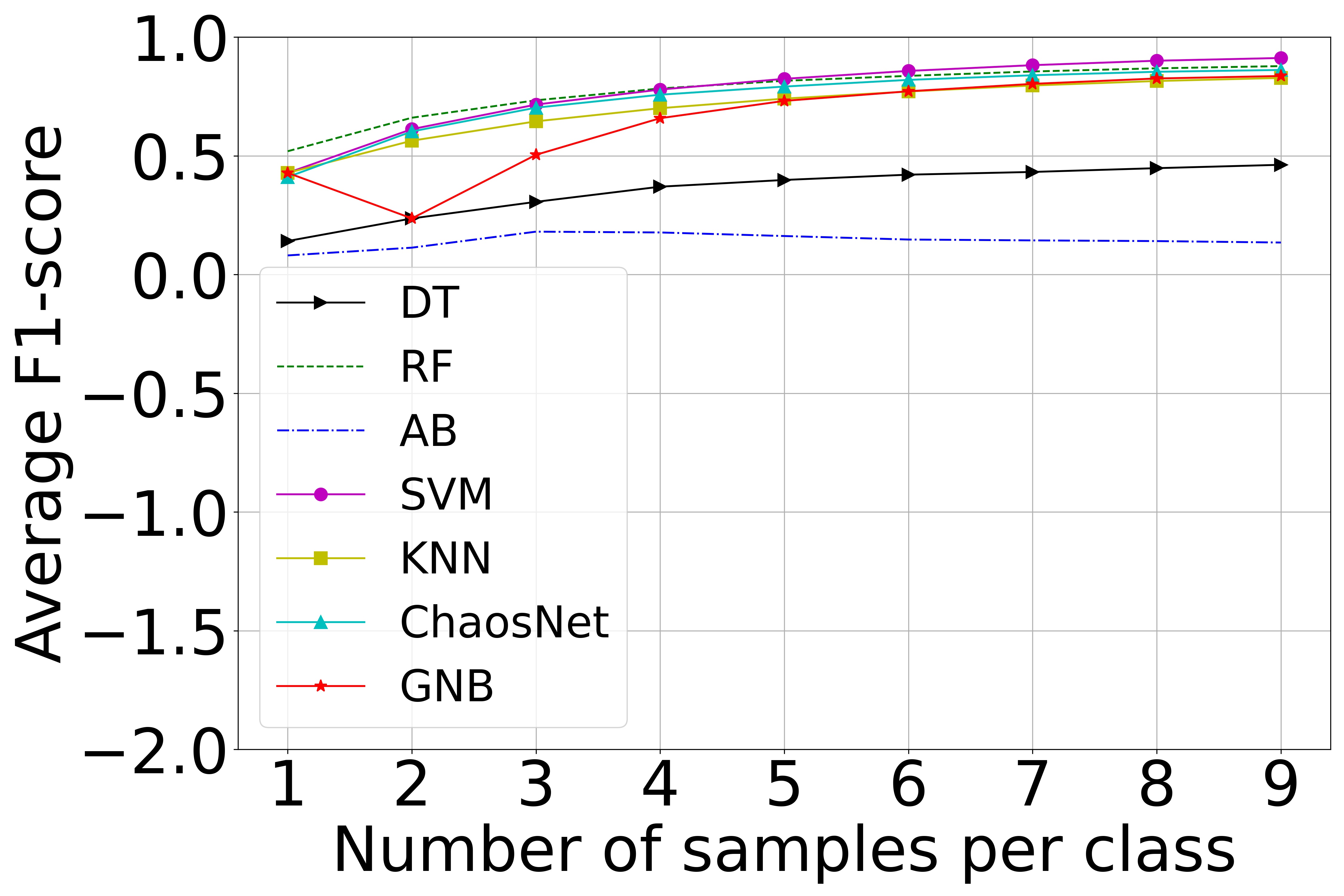}
\caption{}\label{FSDD-LTS-SA}
\end{subfigure}
\hfill
\begin{subfigure}{0.49\textwidth}
\centering
\includegraphics[width=\textwidth]{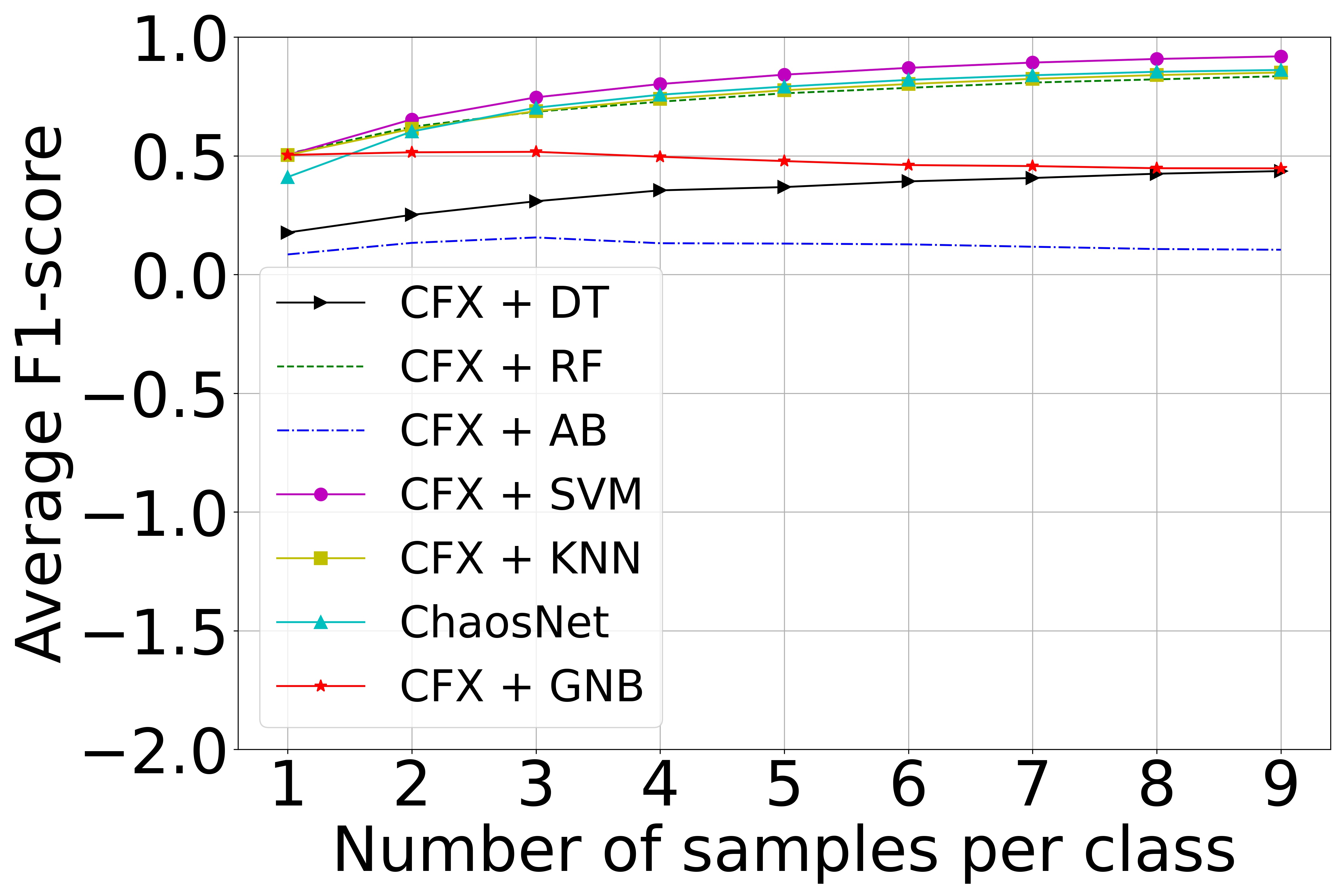}
\caption{}\label{FSDD-LTS-CFX}
\end{subfigure}
\caption{{\it Free Spoken Digit Dataset. }(\subref{FSDD-HTS}) High training sample regime. (\subref{FSDD-LTS-SA}) Comparative performance of stand-alone algorithms in the low training sample regime. (\subref{FSDD-LTS-CFX}) Comparative performance of CFX+ML algorithms in the low training sample regime.}
\label{FSDD-RES}
\end{figure*}

The tuned hyperparameters used and all experiment results for the {\it FSDD} dataset are available in Table~\ref{table:FSDD-HPT} and Figure~\ref{FSDD-RES} respectively.
\begin{table}[!ht]
\centering
\caption{Hyperparameters used for {\it FSDD} dataset for high and low training sample regime experiments~\cite{harikrishnan2021noise}). }
\begin{tabular}{|l|l|}
\hline
\textbf{Hyperparameter} & \textbf{Tuned Value} \\ \hline
q                       & 0.340               \\ \hline
b                       & 0.499                \\ \hline
$\epsilon$              & 0.178                \\ \hline
\end{tabular}
\label{table:FSDD-HPT}
\end{table}

\newpage

\section{Discussion\label{sec:Discussion}} 
\subsection{High Training Sample Regime}

The overall comparative performance of different algorithms is provided in Table~\ref{Table_overall_comparative_perfomance}. In the high training sample regime, the efficacy of using CFX features is evident from Table~\ref{Table_overall_comparative_perfomance} for {\it Iris, Ionosphere, Haberman's Survival, Statlog (Heart)} and {\it FSDD}. While {\it Iris} is a balanced dataset, {\it Ionosphere, Haberman's Survival} and {\it Statlog (Heart)} are imbalanced. Through this, a versatility in the algorithm's ability to perform with both balanced and imbalanced datasets can be established. 

The performance boost after using CFX features is calculated as follows:
\begin{equation}
    Boost = \Big(\frac{F1_{\mbox{\tiny CFX+ML}} - F1_{\mbox{\tiny ML}}}{F1_{\mbox{\tiny ML}}}\Big) \cdot 100 \%,
    \label{eqn_boost}
\end{equation}
where $F1_{\mbox{\tiny ML}}$ and $F1_{\mbox{\tiny CFX+ML}}$ refers to the macro F1-score for the stand-alone ML algorithm and the hybrid NL (CFX$+$ML) algorithm respectively. 

\begin{table}[!ht]
\centering
\caption{Overall comparative performance of different algorithms in the high training sample regime. Percentages in parenthesis indicate the $Boost$ computed using Eq~\ref{eqn_boost}.}
\label{Table_overall_comparative_perfomance}
\begin{tabular}{|l|c|c|c|}
\hline
Dataset & \begin{tabular}[c]{@{}c@{}}\verb|ChaosNet|\\ (Macro F1-Score)\end{tabular} & \begin{tabular}[c]{@{}c@{}}Best\\ Algorithm\end{tabular} & \begin{tabular}[c]{@{}c@{}}Best \\ Macro F1-Score\end{tabular} \\ \hline
Iris & \textbf{1.0} & \begin{tabular}[c]{@{}c@{}}\verb|ChaosNet|, \\ RF, KNN\\ \textbf{CFX+DT (3.41\%)},\\ \textbf{CFX+RF (0\%)}\end{tabular} & \textbf{1.0} \\ \hline
Ionosphere & 0.860 & \textbf{CFX+KNN (14.37\%)} & \textbf{0.939} \\ \hline
Wine & 0.976 & GNB & \textbf{1.0} \\ \hline
\begin{tabular}[c]{@{}l@{}}Bank Note\\ Authentication\end{tabular} & 0.845 & SVM, KNN & \textbf{0.993} \\ \hline
Haberman's Survival & 0.560 & \textbf{CFX+AB (20.59\%)} & \textbf{0.609} \\ \hline
Breast Cancer Wisconsin & 0.927 & $k$-NN & \textbf{0.954} \\ \hline
Statlog (Heart) & 0.738 & \textbf{CFX+DT (25.97\%)} & \textbf{0.878} \\ \hline
Seeds & 0.845 & KNN & \textbf{0.924} \\ \hline
FSDD & 0.897 & \textbf{CFX+SVM (2.73\%)} & \textbf{0.978} \\ \hline
\end{tabular}
\end{table}

\newpage

\subsection{Low Training Sample Regime}

Table~\ref{Table_overall_comparative_perfomance_LTS} demonstrates the overall performance of all datasets in the low training sample regime after employing the CFX features. A \checkmark refers to a performance boost of ML algorithms after using CFX features. CFX features have shown an evident improvement in performance for {\it Ionosphere, Haberman's Survival, Statlog (Heart), Seeds} and {\it FSDD}. $150$ random trials of training in the low training sample regime with $1, 2, \ldots, 9$ samples per class are performed. The $150$ random trials of training ensures the model does not overfit. 

\begin{table}[!ht]
\centering

\caption{Overall performance boost using CFX features in the low training sample regime. A \checkmark refers to a performance boost of ML algorithms after using CFX features. We report $(Minimum, Maximum)$ of $Boost$ values only in these instances.}
\label{Table_overall_comparative_perfomance_LTS}
\scalebox{0.82}{
\begin{tabular}{|l|c|c|c|c|c|c|}
\hline
Dataset & DT & RF & AB & SVM & KNN & GNB \\ \hline
Iris & \textbf{} &  & \textbf{} &  &  &  \\ \hline
Ionosphere & \begin{tabular}[c]{@{}c@{}}\checkmark\\ ($0.88\%$, $17.63\%$)\end{tabular} & \begin{tabular}[c]{@{}c@{}}\checkmark\\ $(0.15\%, 18.01\%)$\end{tabular} & \begin{tabular}[c]{@{}c@{}}\checkmark\\ $(0.83\%, 10.71\%)$\end{tabular} & \begin{tabular}[c]{@{}c@{}}\checkmark\\ $(1.15\%, 17.60\%)$\end{tabular} & \begin{tabular}[c]{@{}c@{}}\checkmark\\ $(0.80\%, 17.60\%)$\end{tabular} & \\ \hline
Wine &  &  & \textbf{} &  &  &  \\ \hline
\begin{tabular}[c]{@{}l@{}}Bank Note\\ Authentication\end{tabular} &  &  & \textbf{} &  &  &  \\ \hline
\begin{tabular}[c]{@{}l@{}}Haberman's\\ Survival\end{tabular}  & \begin{tabular}[c]{@{}c@{}}\checkmark\\ $(0.0\%, 21.59\%)$\end{tabular} & \begin{tabular}[c]{@{}c@{}}\checkmark\\ $(8.43\%, 144.38\%)$\end{tabular} &  & \begin{tabular}[c]{@{}c@{}}\checkmark\\ $(2.19\%, 6.43\%)$\end{tabular} & \begin{tabular}[c]{@{}c@{}}\checkmark\\ $(3.16\%, 8.85\%)$\end{tabular} &  \\ \hline
\begin{tabular}[c]{@{}l@{}}Breast\\ Cancer\\ Wisconsin\end{tabular} &  &  &  &  &  &  \\ \hline
\begin{tabular}[c]{@{}l@{}}Statlog\\ (Heart)\end{tabular} & \begin{tabular}[c]{@{}c@{}}\checkmark\\ $(0.0\%, 1.05\%)$\end{tabular} &  & \begin{tabular}[c]{@{}c@{}}\checkmark\\ $(0.69\%, 5.58\%)$\end{tabular} &  &  &  \\ \hline
Seeds & \begin{tabular}[c]{@{}c@{}}\checkmark\\ $(0.0\%, 9.08\%)$\end{tabular} & \begin{tabular}[c]{@{}c@{}}\checkmark\\ $(0.01\%, 0.84\%)$\end{tabular} & \begin{tabular}[c]{@{}c@{}}\checkmark\\ $(1.56\%, 19.02\%)$\end{tabular} &  & \begin{tabular}[c]{@{}c@{}}\checkmark\\ $(0.04\%, 0.43\%)$\end{tabular} &  \\ \hline
FSDD &  &  &  & \begin{tabular}[c]{@{}c@{}}\checkmark\\ $(0.76\%, 17.63\%)$\end{tabular}  & \begin{tabular}[c]{@{}c@{}}\checkmark\\ $(2.76\%, 17.63\%)$\end{tabular}  &  \\ \hline

\end{tabular}}
\end{table}

\newpage
\subsection{Inferences based on consistency of algorithms across all datasets}

The consistency of an algorithm can be inferred by evaluating the range of minimum and maximum macro F1-scores produced by it in the high training sample regime. Table~\ref{Table_Consistency} shows the ranges of macro F1-scores of different algorithms, measured across all nine datasets used in the research. The provided format for the range of macro F1-scores in Table~\ref{Table_Consistency} is [Minimum, Maximum]. \verb|ChaosNet| ranks second in the least difference between the maximum and minimum macro F1-scores. This observation owes to the consistency and tolerance towards dataset diversity in \verb|ChaosNet|. In algorithms such as AdaBoost (AB) and Support Vector Machine (SVM), a relatively low difference between F1-scores is realised after the usage of CFX features. Gaussian Naive Bayes (GNB) shows the least difference between F1-scores. Along datasets of different domains considered in this study, the performance of \verb|ChaosNet| can be seen as comparable with GNB. 

\begin{table}[!ht]
\centering
\caption{Depiction of consistency of algorithms across all nine datasets used in this study (high training sample regime). The minimum and maximum macro F1-scores for each algorithm corresponding to all nine datasets are provided in the format: $[Minimum, Maximum]$. }
\begin{tabular}{|cc|cc|}
\hline
\multicolumn{2}{|c|}{\textbf{Stand-alone ML}}          & \multicolumn{2}{c|}{\textbf{CFX + ML}}              \\ \hline
\multicolumn{1}{|c|}{\textbf{Algorithm}} &
  \textbf{\begin{tabular}[c]{@{}c@{}}F1-Score\\ Range\end{tabular}} &
  \multicolumn{1}{c|}{\textbf{Algorithm}} &
  \textbf{\begin{tabular}[c]{@{}c@{}}F1-Score\\ Range\end{tabular}} \\ \hline
\multicolumn{1}{|c|}{ChaosNet} & {[}0.56, 1.0{]}    & \multicolumn{1}{c|}{-}         &             -       \\ \hline
\multicolumn{1}{|c|}{DT}       & {[}0.516, 0.967{]} & \multicolumn{1}{c|}{CFX + DT}  & {[}0.482, 1.0{]}   \\ \hline
\multicolumn{1}{|c|}{RF}       & {[}0.56, 1.0{]}    & \multicolumn{1}{c|}{CFX + RF}  & {[}0.398, 1.0{]}   \\ \hline
\multicolumn{1}{|c|}{AB}       & {[}0.08, 0.985{]}  & \multicolumn{1}{c|}{CFX + AB}  & {[}0.397, 0.925{]} \\ \hline
\multicolumn{1}{|c|}{SVM}      & {[}0.437, 0.993{]} & \multicolumn{1}{c|}{CFX + SVM} & {[}0.447, 0.978{]} \\ \hline
\multicolumn{1}{|c|}{KNN}      & {[}0.48, 1.0{]}    & \multicolumn{1}{c|}{CFX + KNN} & {[}0.455, 0.971{]} \\ \hline
\multicolumn{1}{|c|}{GNB}      & {[}0.572, 1.0{]}   & \multicolumn{1}{c|}{CFX + GNB} & {[}0.535, 0.976{]} \\ \hline
\end{tabular}

\label{Table_Consistency}
\end{table}

\subsection{Limitations}

With the current implementation of the ChaosFEX algorithm, computation of image datasets is a costly process. This may limit the use of NL architectures in practical situations involving images. 

Furthermore, NL architectures have been based on certain assumptions. One assumption revolves around the separability of data. NL assumes that applying a nonlinear chaotic transformation will result in separable data suitable for classification, which may not be true in all cases. As it stands, the input layer of NL treats input attributes as independent of each other. It establishes no connection between the neurons for each input attribute. This limitation can be addressed by using coupled chaotic neurons in the input layer. Currently, we have not considered multi-layered NL (Deep-NL) which could significantly enhance performance (by careful choice of coupling between adjacent layers). Another limitation of the NL architectures is the lack of a principled approach to tune the best hyperparameters for a classification task. Currently, cross-validation experiments are used to tune the hyperparameters. The connection between the degree of chaos as measured by lyapunov exponent~\cite{dingwell2006lyapunov})
and learnability is also worth exploring for future research.

\section{Conclusion}
\label{sec:Conclusion}

Decision making under the presence of rare events is a challenging problem in the ML community. This is because rare events have limited data instances, and this problem boils down to imbalanced learning. In this work, we have evaluated the effectiveness of \emph{ChaosFEX (CFX)} feature transformation used in \emph{Neurochaos Learning (NL)} architectures for imbalanced learning. Nine benchmark datasets were used in this study to bring out this evaluation. Seven out of nine datasets used are imbalanced (Refer to Table~\ref{table:test-train split}). This paper accomplishes a comparative study on the performance of NL architecture: \verb|ChaosNet| and CFX+ML with classical Machine Learning (ML) algorithms. The obtained results reflect an evident performance boost in terms of macro F1-score after a {\it nonlinear chaotic transformation} (ChaosFEX or CFX features). Additionally, the efficacy of CFX features can be observed in five out of nine balanced and imbalanced datasets in the high training sample regime, with a boost ranging from $\textbf{2.73\%}$ (\emph{Free Spoken Digit Dataset}) to $\textbf{25.97\%}$ (\emph{Statlog - Heart}). In the low training sample regime, the integration of CFX features has boosted the performance of classical ML algorithms in five datasets, from a total of nine datasets. A maximum boost of $\textbf{144.38\%}$ on the \emph{Haberman's Survival} dataset using CFX+RF is obtained. Refer to  Table~\ref{Table_overall_comparative_perfomance} and~\ref{Table_overall_comparative_perfomance_LTS} for the detailed performance boost using CFX features. This is the first study thoroughly evaluating the performance of NL in the imbalanced learning scenario. 

NL is a unique combination of chaos and noise-enhanced classification. The enormous flexibility of NL offers endless possibilities for development of novel NL: chaos-based-hybrid ML models that suit the application at hand. As new ML algorithms get invented, they can be readily combined in the NL framework. We forsee exciting combinations of CFX with DL and other ML algorithms in the future.

\section{Acknowledgments}
Deeksha Sethi is thankful to Saneesh Cleatus T, Associate Professor, BMS Institute of Technology and Management for enabling her with this research opportunity. Sethi dedicates this work to her family. 
Harikrishnan N. B. thanks ``The University of Trans-Disciplinary Health Sciences and Technology (TDU)'' for permitting this research as part of the PhD programme. The authors gratefully acknowledge the financial support of Tata Trusts. The authors acknowledge the computational facility supported by NIAS Consciousness Studies Programme.

\section*{Declarations}
\begin{itemize}
    \item Funding: The authors gratefully acknowledge the financial support of Tata Trusts.
    \item Conflicts of Interest/Competing interests: There is no conflict of interest or competing interests.
    \item Ethics approval: Not Applicable.
    \item Consent to participate: Not Applicable. 
    \item Consent for publication: Not Applicable.
    \item Availability of data and materials: Not Applicable.
    \item Code availability: The codes used in this research are available in the following link: \url{https://github.com/deeksha-sethi03/nl-imbalanced-learning}.
    \item Authors' contributions: 
    
    \begin{itemize}
        \item Conceptualization: Harikrishnan NB
        \item Methodology: Harikrishnan NB
        \item Formal analysis and investigation: Deeksha Sethi, Harikrishnan NB, Nithin Nagaraj
        \item Code implementation: Deeksha Sethi (Template codes provided by Harikrishnan NB)
        \item Writing - original draft preparation: Deeksha Sethi, Harikrishnan NB
        \item Writing - review and editing: Deeksha Sethi, Harikrishnan NB, Nithin Nagaraj
        \item Funding acquisition: Nithin Nagaraj
        \item Resources: Conciousness Studies Programme, National Institute of Advanced Studies
        \item Supervision: Harikrishnan NB, Nithin Nagaraj
    \end{itemize}

\end{itemize}


\newpage
\section{Supplementary Information} 
\label{sec:Supplementary Information}
This is the supplementary information pertaining to the main manuscript. It contains the following -- (1) description of datasets used in our study including the coding rule for the labels of different classes, (2) hyperparamter tuning details for each dataset and for each  ML algorithm (Decision Tree, Random Forest, AdaBoost, SVM, $k$-NN) and NL algorithm (\verb|ChaosNet|) used in the study, (3) the test data macro F1-scores for each algorithm in the high training sample regime for each dataset.

\subsection{Dataset Description}
\label{subsec:Dataset Description}

 \subsubsection{Iris}
 \begin{table}[!ht]
\begin{center}
\caption{\textbf{\textit{Iris}}: Rule followed for renaming of the class labels.}

\begin{tabular}{|l|l|l|}
\hline
\textbf{Class Label} & \textbf{Numeric Code} & \textbf{\begin{tabular}[c]{@{}l@{}}Number of Total\\ Data Instances\end{tabular}} \\ \hline
Iris-Setosa      & 0 & 50 \\ \hline
Iris-Versicolour & 1 & 50 \\ \hline
Iris-Virginica   & 2 & 50 \\ \hline
\end{tabular}
\label{table:Iris rule}
\end{center}
\end{table}

\subsubsection{Ionosphere}
\begin{table}[!ht]
\centering
\caption{\textbf{\textit{Ionosphere}}: Rule followed for renaming of the class labels.}
\begin{tabular}{|l|l|l|}
\hline
\textbf{Class Label} & \textbf{Numeric Code} & \textbf{\begin{tabular}[c]{@{}l@{}}Number of Total\\ Data Instances\end{tabular}} \\ \hline
b (Bad)              & 0                     & 126                                                                                               \\ \hline
g (Good)             & 1                     & 225                                                                                               \\ \hline
\end{tabular}

\label{table:Ionosphere rule}
\end{table}

\subsubsection{Wine}
\begin{table}[!ht]
\centering
\caption{\textbf{\textit{Wine}}: Rule followed for renaming of the class labels.}
\begin{tabular}{|l|l|l|}
\hline
\textbf{Class Label} & \textbf{Numeric Code} & \textbf{\begin{tabular}[c]{@{}l@{}}Number of Total\\ Data Instances\end{tabular}} \\ \hline
1 & 0 & 59 \\ \hline
2 & 1 & 71 \\ \hline
3 & 2 & 48 \\ \hline
\end{tabular}

\label{table:Wine rule}
\end{table}
\newpage
\subsubsection{Bank Note Authentication}
\begin{table}[!ht]
\centering
\caption{\textbf{\textit{Bank Note Authentication}}: Rule followed for renaming of the class labels.}
\begin{tabular}{|l|l|l|}
\hline
\textbf{Class Label} & \textbf{Numeric Code} & \textbf{\begin{tabular}[c]{@{}l@{}}Number of Total\\ Data Instances\end{tabular}} \\ \hline
0 (Genuine)          & 0                     & 762                                                                                               \\ \hline
1 (Forgery)          & 1                     & 610                                                                                               \\ \hline
\end{tabular}

\label{table:Bank Note Authentication rule}
\end{table}

\subsubsection{Haberman's Survival}
\begin{table}[!ht]
\centering
\caption{\textbf{\textit{Haberman's Survival}}: Rule followed for renaming of the class labels.}
\begin{tabular}{|l|l|l|}
\hline
\textbf{Class Label} & \textbf{Numeric Code} & \textbf{\begin{tabular}[c]{@{}l@{}}Number of Total\\ Data Instances\end{tabular}} \\ \hline
1 ($<$ 5yrs)   & 0                     & 225                                                                                               \\ \hline
2 ($\geq$ 5yrs)      & 1                     & 81                                                                                                \\ \hline
\end{tabular}

\label{table:Haberman's Survival rule}
\end{table}

\subsubsection{Breast Cancer Wisconsin}
\begin{table}[!ht]
\centering
\caption{\textbf{\textit{Breast Cancer Wisconsin}}: Rule followed for renaming of the class labels.}
\begin{tabular}{|l|l|l|}
\hline
\textbf{Class Label} & \textbf{Numeric Code} & \textbf{\begin{tabular}[c]{@{}l@{}}Number of Total\\ Data Instances\end{tabular}} \\ \hline
M (Malignant)          & 0                     & 212                                                                                         \\ \hline
B (Benign)         & 1                     & 357                                                                                          \\ \hline
\end{tabular}

\label{table:Breast Cancer rule}
\end{table}

\subsubsection{Statlog (Heart)}
\begin{table}[!ht]
\centering
\caption{\textbf{\textit{Statlog (Heart)}}: Rule followed for renaming of the class labels.}
\begin{tabular}{|l|l|l|}
\hline
\textbf{Class Label} & \textbf{Numeric Code} & \textbf{\begin{tabular}[c]{@{}l@{}}Number of Total\\ Data Instances\end{tabular}} \\ \hline
1 (Absence)          & 0                     & 150                                                                                          \\ \hline
2 (Presence)         & 1                     & 120                                                                                          \\ \hline
\end{tabular}

\label{table:Statlog (Heart) rule}
\end{table}

\subsubsection{Seeds}
\begin{table}[!ht]
\centering
\caption{\textbf{\textit{Seeds}}: Rule followed for renaming of the class labels.}
\begin{tabular}{|l|l|l|}
\hline
\textbf{Class Label} & \textbf{Numeric Code} & \textbf{\begin{tabular}[c]{@{}l@{}}Number of Total\\ Data Instances\end{tabular}} \\ \hline
1.0 (Kama)     & 0 & 70 \\ \hline
2.0 (Rosa)     & 1 & 70 \\ \hline
3.0 (Canadian) & 2 & 70 \\ \hline
\end{tabular}

\label{table:Seeds rule}
\end{table}

\subsubsection{FSDD}
\begin{table}[!ht]
\centering
\caption{\textbf{\textit{FSDD}}: Rule followed for renaming of the class labels.}
\begin{tabular}{|l|l|l|}
\hline
\textbf{Class Label} & \textbf{Numeric Code} & \textbf{\begin{tabular}[c]{@{}l@{}}Number of Total\\ Data Instances\end{tabular}} \\ \hline
0 & 0 & 50 \\ \hline
1 & 1 & 50 \\ \hline
2 & 2 & 50 \\ \hline
3 & 3 & 50 \\ \hline
4 & 4 & 46 \\ \hline
5 & 5 & 41 \\ \hline
6 & 6 & 50 \\ \hline
7 & 7 & 50 \\ \hline
8 & 8 & 43 \\ \hline
9 & 9 & 50 \\ \hline
\end{tabular}

\label{table:FSDD rule}
\end{table}

\newpage
\subsection{Hyperparameter Tuning}
The hyperparameter tuning for all algorithms for the respective datasets are provided below:

\subsubsection{Decision Tree}

Following are the hyperparameters tuned for Decision Tree:

\begin{enumerate}
\item \textbf{min\_samples\_leaf}: Defines the minimum number of samples required for a leaf node in the decision tree. It is tuned from $1$ to $10$ with a step-size of $1$. 

\item \textbf{max\_depth}: Declares the maximum depth to which a decision tree can be grown. It is tuned from $1$ to $10$ with a step-size of $1$. 

\item \textbf{ccp\_alpha}: A numpy array of alpha values obtained by devising Cost Complexity Pruning on the original decision tree. This array is obtained using the \textit{cost\_complexity\_pruning\_path}. 
\end{enumerate}
All remaining hyperparameters offered by scikit-learn are retained in their default forms. The results of hyperparameter tuning for Decision Tree are available in Table ~\ref{table:Hyperparameter Tuning 1 - Decision Tree} and ~\ref{table:Hyperparameter Tuning 2 - Decision Tree}.\\

                \begin{table}[!ht]
                \centering
                \caption{\textbf{\textit{Decision Tree}}: Tuned hyperparameters for all nine datasets (Part I). The performance metric used for the provided results is macro F1-score.}
                \begin{tabular}{| l | l | l | l |} 
                \hline
                 Dataset & Implementation & Tuned Hyperparameters & F1 Score \\ [0.5ex] 
                  \hline

                 \multirow{9}{*}{Iris} & \multirow{3}{*}{Stand-Alone Decision Tree} & $min\_samples\_leaf = 3$ & \multirow{3}{*}{0.931}\\
                 \cline{3-3}
                  &  & $max\_depth = 3$ & \\
                 \cline{3-3}
                   & & $ccp\_alpha = 0.0$ & \\
                 \cline{2-4}
                  & \multirow{6}{*}{ChaosFEX + Decision Tree} & $min\_samples\_leaf = 3$ & \multirow{3}{*}{0.955}\\
                 \cline{3-3}
                  &  & $max\_depth = 4$ & \\
                 \cline{3-3}
                  &  & $ccp\_alpha = 0.0$ & \\
                 \cline{3-3}
                  &  & $q = 0.21$ & \\
                 \cline{3-3}
                  & & $b = 0.969$ & \\
                 \cline{3-3}
                  & & $\epsilon = 0.13$ & \\
                 \hline

                  \multirow{9}{*}{Ionosphere} & \multirow{3}{*}{Stand-Alone Decision Tree} & $min\_samples\_leaf = 1$ & \multirow{3}{*}{0.882}\\
                 \cline{3-3}
                  &  & $max\_depth = 2$ & \\
                 \cline{3-3}
                   & & $ccp\_alpha = 0.0$ & \\
                 \cline{2-4}
                  & \multirow{6}{*}{ChaosFEX + Decision Tree } & $min\_samples\_leaf = 1$ & \multirow{6}{*}{0.921}\\
                 \cline{3-3}
                  &  & $max\_depth = 7$ & \\
                 \cline{3-3}
                  &  & $ccp\_alpha = 0.0$ & \\
                 \cline{3-3}
                  &  & $q = 0.21$ & \\
                 \cline{3-3}
                  & & $b = 0.969$ & \\
                 \cline{3-3}
                  & & $\epsilon = 0.22$ & \\
                \hline

                \multirow{9}{*}{Wine} & \multirow{3}{*}{Stand-Alone  Decision Tree} & $min\_samples\_leaf = 1$ & \multirow{3}{*}{0.916}\\
                 \cline{3-3}
                  &  & $max\_depth = 3$ & \\
                 \cline{3-3}
                   & & $ccp\_alpha = 0.0$ & \\
                 \cline{2-4}
                  & \multirow{6}{*}{ChaosFEX +  Decision Tree} & $min\_samples\_leaf = 1$ & \multirow{6}{*}{0.949}\\
                 \cline{3-3}
                  &  & $max\_depth = 6$ & \\
                 \cline{3-3}
                  & & $ccp\_alpha = 0.0$ & \\
                 \cline{3-3}
                  &  & $q = 0.21$ & \\
                 \cline{3-3}
                  & & $b = 0.969$ & \\
                 \cline{3-3}
                  & & $\epsilon = 0.07$ & \\
                \hline

                \multirow{9}{*}{Bank Note Authentication} & \multirow{3}{*}{Stand-Alone  Decision Tree} & $min\_samples\_leaf = 1$ & \multirow{3}{*}{0.977}\\
                 \cline{3-3}
                  &  & $max\_depth = 8$ & \\
                 \cline{3-3}
                   & & $ccp\_alpha = 0.0$ & \\
                 \cline{2-4}
                  & \multirow{6}{*}{ChaosFEX +  Decision Tree} & $min\_samples\_leaf = 1$ & \multirow{6}{*}{0.961}\\
                 \cline{3-3}
                  &  & $max\_depth = 6$ & \\
                 \cline{3-3}
                  &  & $ccp\_alpha = 0.000836$ & \\
                 \cline{3-3}
                  &  & $q = 0.080$ & \\
                 \cline{3-3}
                  & & $b = 0.250$ & \\
                 \cline{3-3}
                  & & $\epsilon = 0.233$ & \\
                
                \hline
                
                 \multirow{9}{*}{Haberman's Survival} & \multirow{3}{*}{Stand-Alone  Decision Tree} & $min\_samples\_leaf = 4$ & \multirow{3}{*}{0.614}\\
                 \cline{3-3}
                   &  & $max\_depth = 6$ & \\
                 \cline{3-3}
                   & & $ccp\_alpha = 0.005389$ & \\
                 \cline{2-4}
                  & \multirow{6}{*}{ChaosFEX + Decision Tree} & $min\_samples\_leaf = 2$ & \multirow{6}{*}{0.648}\\
                 \cline{3-3}
                  &  & $max\_depth = 8$ & \\
                 \cline{3-3}
                  & & $ccp\_alpha =  0.005389$ & \\
                 \cline{3-3}
                  &  & $q = 0.81$ & \\
                 \cline{3-3}
                  & & $b = 0.14$& \\
                 \cline{3-3}
                  & & $\epsilon = 0.003$ & \\
                
                  \hline
                 
                 \end{tabular}
                
                \label{table:Hyperparameter Tuning 1 - Decision Tree}
                \end{table}
                \newpage
                
                \begin{table}[!ht]
                \centering
                \caption{\textbf{\textit{Decision Tree}}: Tuned hyperparameters for all nine datasets (Part II). The performance metric used for the provided results is macro F1-score.}
                \begin{tabular}{| l | l | l | l |} 
                \hline
                 Dataset & Implementation & Tuned Hyperparameters & F1 Score \\ [0.5ex] 
                  \hline
                
                  \multirow{9}{*}{Breast Cancer Wisconsin (Diagnostic)} & \multirow{3}{*}{Stand-Alone Decision Tree} & $min\_samples\_leaf = 1$ & \multirow{3}{*}{0.920}\\
                 \cline{3-3}
                   &  & $max\_depth = 4$ & \\
                 \cline{3-3}
                   & & $ccp\_alpha = 0.0$ & \\
                 \cline{2-4}
                  & \multirow{6}{*}{ChaosFEX +  Decision Tree} & $min\_samples\_leaf = 2$ & \multirow{6}{*}{0.945}\\
                 \cline{3-3}
                  &  & $max\_depth = 8$ & \\
                 \cline{3-3}
                  &  & $ccp\_alpha = 0.005595$ & \\
                 \cline{3-3}
                  &  & $q = 0.930$ & \\
                 \cline{3-3}
                  & & $b = 0.490$& \\
                 \cline{3-3}
                  & & $\epsilon = 0.159$ & \\
                  
                 \hline
                
                  \multirow{9}{*}{Statlog (Heart)} & \multirow{3}{*}{Stand-Alone Decision Tree} & $min\_samples\_leaf = 6$ & \multirow{3}{*}{0.779}\\
                 \cline{3-3}
                   &  & $max\_depth = 3$ & \\
                 \cline{3-3}
                   & & $ccp\_alpha = 0.006818$ & \\
                 \cline{2-4}
                  & \multirow{6}{*}{ChaosFEX + Decision Tree} & $min\_samples\_leaf = 7$ & \multirow{6}{*}{0.786}\\
                 \cline{3-3}
                  &  & $max\_depth = 4$ & \\
                 \cline{3-3}
                  &  & $ccp\_alpha = 0.0$ & \\
                 \cline{3-3}
                  &  & $q = 0.08$ & \\
                 \cline{3-3}
                  & & $b = 0.06$& \\
                 \cline{3-3}
                  & & $\epsilon = 0.17$ & \\

                 \hline
                \multirow{9}{*}{Seeds} & \multirow{3}{*}{Stand-Alone Decision Tree} & $min\_samples\_leaf = 2$ & \multirow{3}{*}{0.918}\\
                 \cline{3-3}
                   &  & $max\_depth = 4$ & \\
                 \cline{3-3}
                   & & $ccp\_alpha = 0.005844$ & \\
                 \cline{2-4}
                  & \multirow{6}{*}{ChaosFEX +  Decision Tree} & $min\_samples\_leaf = 2$ & \multirow{6}{*}{0.949}\\
                 \cline{3-3}
                  &  & $max\_depth = 5$ & \\
                 \cline{3-3}
                  &  & $ccp\_alpha = 0.0$ & \\
                 \cline{3-3}
                  &  & $q = 0.02$ & \\
                 \cline{3-3}
                  & & $b = 0.07$& \\
                 \cline{3-3}
                  & & $\epsilon = 0.238$ & \\

                  \hline
                  
                \multirow{9}{*}{FSDD} & \multirow{3}{*}{Stand-Alone  Decision Tree} & $min\_samples\_leaf = 1$ & \multirow{3}{*}{0.536}\\
                 \cline{3-3}
                  &  & $max\_depth = 9$ & \\
                 \cline{3-3}
                   & & $ccp\_alpha = 0.0099206$ & \\
                 \cline{2-4}
                  & \multirow{6}{*}{ChaosFEX +  Decision Tree} & $min\_samples\_leaf = 1$ & \multirow{6}{*}{0.537}\\
                 \cline{3-3}
                  &  & $max\_depth = 10$ & \\
                 \cline{3-3}
                  &  & $ccp\_alpha = 0.0061384$ & \\
                 \cline{3-3}
                  &  & $q = 0.34$ & \\
                 \cline{3-3}
                  & & $b = 0.499$ & \\
                 \cline{3-3}
                  & & $\epsilon = 0.178$ & \\

                \hline 
                 
                \end{tabular}
                
                \label{table:Hyperparameter Tuning 2 - Decision Tree}
                \end{table}

\subsubsection{Random Forest}
Following are the hyperparameters tuned for Random Forest:

\begin{enumerate}
\item \textbf{n\_estimators}: Defines the number of trees in the forest being grown. The values are tuned from the array $[1, 10, 100, 1000, 10000]$.

\item \textbf{max\_depth}: Declares the maximum depth each tree the the forest is allowed to have. It is tuned from $1$ to $10$ with a step-size of $1$.

\end{enumerate}
All  remaining  hyperparameters  offered  by  scikit-learn  are  retained  in  their default forms. The results of hyperparameter tuning for Random Forest are available in Table ~\ref{table:Hyperparameter Tuning 1 - Random Forest} and ~\ref{table:Hyperparameter Tuning 2 - Random Forest}\\
                \begin{table}[!ht]
                \centering
                \caption{\textbf{\textit{Random Forest}}: Tuned hyperparameters for all nine datasets (Part I). The performance metric used for the provided results is macro F1-score.}
                \begin{tabular}{| l | l | l | l |} 
                \hline
                 Dataset & Implementation & Tuned Hyperparameters & F1 Score \\ [0.5ex] 
                \hline
                  
                 \multirow{7}{*}{Iris} & \multirow{2}{*}{Stand-Alone  Random Forest} & $n\_estimators = 100$ & \multirow{2}{*}{0.944}\\
                 \cline{3-3}
                  &  & $max\_depth = 3$ & \\
                 
                 \cline{2-4}
                  & \multirow{5}{*}{ChaosFEX +  Random Forest} & $n\_estimators = 10$ & \multirow{5}{*}{0.944}\\
                 \cline{3-3}
                  &  & $max\_depth = 5$ & \\
                 \cline{3-3}
                  &  & $q = 0.21$ & \\
                 \cline{3-3}
                  & & $b = 0.969$ & \\
                 \cline{3-3}
                  & & $\epsilon = 0.11$ & \\
                \hline

                  \multirow{7}{*}{Ionosphere} & \multirow{2}{*}{Stand-Alone  Random Forest} & $n\_estimators = 10000$ & \multirow{2}{*}{0.923}\\
                 \cline{3-3}
                  & & $max\_depth = 4$ & \\
                 
                 \cline{2-4}
                  & \multirow{5}{*}{ChaosFEX +  Random Forest} & $n\_estimators = 100$ & \multirow{5}{*}{0.928}\\
                 \cline{3-3}
                  &  & $max\_depth = 10$ & \\
                 \cline{3-3}
                  & & $q = 0.21$ & \\
                 \cline{3-3}
                  & & $b = 0.969$ & \\
                 \cline{3-3}
                  & & $\epsilon = 0.23$ & \\

                \hline
                \multirow{7}{*}{Wine} & \multirow{2}{*}{Stand-Alone Random Forest} & $n\_estimators = 10$ & \multirow{2}{*}{0.980}\\
                 \cline{3-3}
                  &  & $max\_depth = 4$ & \\
                 
                 \cline{2-4}
                  & \multirow{5}{*}{ChaosFEX + Random Forest} & $n\_estimators = 10$ & \multirow{5}{*}{0.987}\\
                 \cline{3-3}
                  &  & $max\_depth = 5$ & \\
                 \cline{3-3}
                  & & $q = 0.21$ & \\
                 \cline{3-3}
                  & & $b = 0.969$ & \\
                 \cline{3-3}
                  & & $\epsilon = 0.05$ & \\
                \hline

                  \multirow{7}{*}{Bank Note Authentication} & \multirow{2}{*}{Stand-Alone Random Forest} & $n\_estimators = 100$ & \multirow{2}{*}{0.991}\\
                 \cline{3-3}
                  &  & $max\_depth = 8$ & \\
                 
                 \cline{2-4}
                  & \multirow{5}{*}{ChaosFEX + Random Forest} & $n\_estimators = 100$ & \multirow{5}{*}{0.967}\\
                 \cline{3-3}
                  &  & $max\_depth = 7$ & \\
                 \cline{3-3}
                  &  & $q = 0.080$ & \\
                 \cline{3-3}
                  & & $b = 0.250$ & \\
                 \cline{3-3}
                  & & $\epsilon = 0.233$ & \\
                \hline

                \multirow{7}{*}{Haberman's Survival} & \multirow{2}{*}{Stand-Alone Random Forest} & $n\_estimators = 1$ & \multirow{2}{*}{0.621}\\
                 \cline{3-3}
                  &  & $max\_depth = 3$ & \\
                 
                 \cline{2-4}
                  & \multirow{5}{*}{ChaosFEX + Random Forest} & $n\_estimators = 1000$ & \multirow{5}{*}{0.651}\\
                 \cline{3-3}
                  &  & $max\_depth = 9$ & \\
                 \cline{3-3}
                  & & $q = 0.810$ & \\
                 \cline{3-3}
                  & & $b = 0.140$ & \\
                 \cline{3-3}
                  & & $\epsilon = 0.003$ & \\
                \hline
                
                \multirow{7}{*}{Breast Cancer Wisconsin (Diagnostic)} & \multirow{2}{*}{Stand-Alone Random Forest} & $n\_estimators = 1000$ & \multirow{2}{*}{0.956}\\
                 \cline{3-3}
                  &  & $max\_depth = 9$ & \\
                 
                 \cline{2-4}
                  & \multirow{5}{*}{ChaosFEX + Random Forest} & $n\_estimators = 100$ & \multirow{5}{*}{0.955}\\
                 \cline{3-3}
                  &  & $max\_depth = 7$ & \\
                 \cline{3-3}
                  &  & $q = 0.930$ & \\
                 \cline{3-3}
                  & & $b = 0.490$ & \\
                 \cline{3-3}
                  & & $\epsilon = 0.159$ & \\

                \hline

                 \end{tabular}
                
                \label{table:Hyperparameter Tuning 1 - Random Forest}
                \end{table}
                \newpage
                
                \begin{table}[!ht]
                \centering
                \caption{\textbf{\textit{Random Forest}}: Tuned hyperparameters for all nine datasets (Part II). The performance metric used for the provided results is macro F1-score.}
                \begin{tabular}{| l | l | l | l |} 
                \hline
                 Dataset & Implementation & Tuned Hyperparameters & F1 Score \\ [0.5ex] 
                \hline
                 
                  \multirow{7}{*}{Statlog (Heart)} & \multirow{2}{*}{Stand-Alone Random Forest} & $n\_estimators = 1000$ & \multirow{2}{*}{ 0.839}\\
                 \cline{3-3}
                  &  & $max\_depth = 2$ & \\
                 
                 \cline{2-4}
                  & \multirow{5}{*}{ChaosFEX + Random Forest} & $n\_estimators = 10000$ & \multirow{5}{*}{0.830}\\
                 \cline{3-3}
                  &  & $max\_depth = 4$ & \\
                 \cline{3-3}
                  &  & $q = 0.08$ & \\
                 \cline{3-3}
                  & & $b = 0.06$ & \\
                 \cline{3-3}
                  & & $\epsilon = 0.17$ & \\
                 \hline

                 \multirow{7}{*}{Seeds} & \multirow{2}{*}{Stand-Alone Random Forest} & $n\_estimators = 100$ & \multirow{2}{*}{0.918}\\
                 \cline{3-3}
                  &  & $max\_depth = 5$ & \\
                 
                 \cline{2-4}
                  & \multirow{5}{*}{ChaosFEX + Random Forest} & $n\_estimators = 1000$ & \multirow{5}{*}{0.941}\\
                 \cline{3-3}
                  &  & $max\_depth = 5$ & \\
                 \cline{3-3}
                  & & $q = 0.020$ & \\
                 \cline{3-3}
                  & & $b = 0.070$ & \\
                 \cline{3-3}
                  & & $\epsilon = 0.238$ & \\
                  
                 \hline

                 \multirow{7}{*}{FSDD} & \multirow{2}{*}{Stand-Alone Random Forest} & $n\_estimators = 1000$ & \multirow{2}{*}{0.9479}\\
                 \cline{3-3}
                  &  & $max\_depth = 9$ & \\
                 
                 \cline{2-4}
                  & \multirow{5}{*}{ChaosFEX + Random Forest} & $n\_estimators = 1000$ & \multirow{5}{*}{0.921}\\
                 \cline{3-3}
                  &  & $max\_depth = 8$ & \\
                 \cline{3-3}
                  & & $q = 0.340$ & \\
                 \cline{3-3}
                  & & $b = 0.499$ & \\
                 \cline{3-3}
                  & & $\epsilon = 0.178$ & \\
                  
                 \hline
                \end{tabular}
                
                \label{table:Hyperparameter Tuning 2 - Random Forest}
                \end{table}
\newpage                
\subsubsection{AdaBoost}
Following are the hyperparameters tuned for Adaptive Boosting (AdaBoost):   
\begin{enumerate}
\item \textbf{n\_estimators}: An upper limit for the number of stumps being grown. The values are tuned from the array \\$[1, 10, 50, 100, 500, 1000, 5000, 10000]$.

\end{enumerate}
All  remaining  hyperparameters  offered  by  scikit-learn  are  retained  in  their default forms. The results of hyperparameter tuning for AdaBoost are available in Table ~\ref{table:Hyperparameter Tuning - AdaBoost}.\\

                \begin{table}[!ht]
                \centering
                \caption{\textbf{\textit{AdaBoost}}: Tuned hyperparameters for all nine datasets. The performance metric used for the provided results is macro F1-score.}
                \begin{tabular}{| l | l | l | l |} 
                \hline
                 Dataset & Implementation & Tuned Hyperparameters & F1 Score \\ [0.5ex] 
                \hline
                 \multirow{5}{*}{Iris} & \multirow{1}{*}{Stand-Alone AdaBoost} & $n\_estimators = 50$ & \multirow{1}{*}{0.929}\\
                 
                 \cline{2-4}
                  & \multirow{4}{*}{ChaosFEX + AdaBoost} & $n\_estimators = 10$ & \multirow{4}{*}{0.915}\\
                 \cline{3-3}
                 
                  &  & $q = 0.21$ & \\
                 \cline{3-3}
                  & & $b = 0.969$ & \\
                 \cline{3-3}
                  & & $\epsilon = 0.03$ & \\
                \hline

                  \multirow{5}{*}{Ionosphere} & \multirow{1}{*}{Stand-Alone AdaBoost} & $n\_estimators = 50$ & \multirow{1}{*}{0.929}\\
                 
                 \cline{2-4}
                  & \multirow{4}{*}{ChaosFEX + AdaBoost} & $n\_estimators = 500$ & \multirow{4}{*}{0.921}\\
                 \cline{3-3}
                 
                  &  & $q = 0.21$ & \\
                 \cline{3-3}
                  & & $b = 0.969$ & \\
                 \cline{3-3}
                  & & $\epsilon = 0.02$ & \\
                \hline

                  \multirow{5}{*}{Wine} & \multirow{1}{*}{Stand-Alone AdaBoost} & $n\_estimators = 10$ & \multirow{1}{*}{0.896}\\
                 
                 \cline{2-4}
                  & \multirow{4}{*}{ChaosFEX + AdaBoost} & $n\_estimators = 10$ & \multirow{4}{*}{0.893}\\
                 \cline{3-3}
                 
                  &  & $q = 0.21$ & \\
                 \cline{3-3}
                  & & $b = 0.969$ & \\
                 \cline{3-3}
                  & & $\epsilon = 0.05$ & \\
                \hline

                \multirow{5}{*}{Bank Note Authentication} & \multirow{1}{*}{Stand-Alone AdaBoost} & $n\_estimators = 500$ & \multirow{1}{*}{0.999}\\
                 
                 \cline{2-4}
                  & \multirow{4}{*}{ChaosFEX + AdaBoost} & $n\_estimators = 500$ & \multirow{4}{*}{0.934}\\
                 \cline{3-3}
                 
                  &  & $q = 0.080$ & \\
                 \cline{3-3}
                  & & $b = 0.250$ & \\
                 \cline{3-3}
                  & & $\epsilon = 0.233$ & \\
                \hline

                  \multirow{5}{*}{Haberman's Survival} & \multirow{1}{*}{Stand-Alone AdaBoost} & $n\_estimators = 10$ & \multirow{1}{*}{0.620}\\
                 
                 \cline{2-4}
                  & \multirow{4}{*}{ChaosFEX + AdaBoost} & $n\_estimators = 1$ & \multirow{4}{*}{0.639}\\
                 \cline{3-3}
                 
                  &  & $q = 0.81$ & \\
                 \cline{3-3}
                  & & $b = 0.14$ & \\
                 \cline{3-3}
                  & & $\epsilon = 0.003$ & \\
                \hline

                 \multirow{5}{*}{Breast Cancer Wisconsin (Heart)} & \multirow{1}{*}{Stand-Alone AdaBoost} & $n\_estimators = 1000$ & \multirow{1}{*}{0.962}\\
                 
                 \cline{2-4}
                  & \multirow{4}{*}{ChaosFEX + AdaBoost} & $n\_estimators = 100$ & \multirow{4}{*}{0.952}\\
                 \cline{3-3}
                 
                  &  & $q = 0.930$ & \\
                 \cline{3-3}
                  & & $b = 0.490$ & \\
                 \cline{3-3}
                  & & $\epsilon = 0.159$ & \\
                \hline

                \multirow{5}{*}{Statlog (Heart)} & \multirow{1}{*}{Stand-Alone AdaBoost} & $n\_estimators = 10$ & \multirow{1}{*}{0.816}\\
                 
                 \cline{2-4}
                  & \multirow{4}{*}{ChaosFEX + AdaBoost} & $n\_estimators = 50$ & \multirow{4}{*}{0.815}\\
                 \cline{3-3}
                 
                  &  & $q = 0.08$ & \\
                 \cline{3-3}
                  & & $b = 0.06$ & \\
                 \cline{3-3}
                  & & $\epsilon = 0.17$ & \\
                \hline

                  \multirow{5}{*}{Seeds} & \multirow{1}{*}{Stand-Alone AdaBoost} & $n\_estimators = 500$ & \multirow{1}{*}{0.563}\\
                 
                 \cline{2-4}
                  & \multirow{4}{*}{ChaosFEX + AdaBoost} & $n\_estimators = 10$ & \multirow{4}{*}{0.918}\\
                 \cline{3-3}
                 
                  &  & $q = 0.020$ & \\
                 \cline{3-3}
                  & & $b = 0.070$ & \\
                 \cline{3-3}
                  & & $\epsilon = 0.238$ & \\
                  
                \hline

                  \multirow{5}{*}{FSDD} & \multirow{1}{*}{Stand-Alone AdaBoost} & $n\_estimators = 50$ & \multirow{1}{*}{0.086}\\
                 
                 \cline{2-4}
                  & \multirow{4}{*}{ChaosFEX + AdaBoost} & $n\_estimators = 50$ & \multirow{4}{*}{0.170}\\
                 \cline{3-3}
                 
                  &  & $q = 0.340$ & \\
                 \cline{3-3}
                  & & $b = 0.499$ & \\
                 \cline{3-3}
                  & & $\epsilon = 0.178$ & \\
                
                \hline
                 \end{tabular}
                \label{table:Hyperparameter Tuning - AdaBoost}
                \end{table}
                \newpage

\subsubsection{Support Vector Machine}
Following are the hyperparameters tuned for Support Vector Machine (SVM):       

\begin{enumerate}
\item \textbf{C}: Controls the bias-variance trade-off of the algorithm, known as the \textit{``Regularization Parameter"}. It is tuned from $0.1$ to $100.0$ with a step-size of $0.1$
\end{enumerate}
All  remaining  hyperparameters  offered  by  scikit-learn  are  retained  in  their default forms. The results of hyperparameter tuning for Support Vector Machine are available in Table ~\ref{table:Hyperparameter Tuning - SVM}.\\
               \begin{table}[!ht]
                \centering
                \caption{\textbf{\textit{Support Vector Machine (SVM)}}: Tuned hyperparameters for all nine datasets. Only FSDD uses the `linear' kernel. All remaining datasets, use the `rbf' kernel. The performance metric used for the provided results is macro F1-score.}
                \begin{tabular}{| l | l | l | l |} 
                \hline
                 Dataset & Implementation & Tuned Hyperparameters & F1 Score \\ [0.5ex] 
                \hline
                
                 \multirow{5}{*}{Iris} & \multirow{1}{*}{Stand-Alone SVM} & $C = 4.4$ & \multirow{1}{*}{0.954}\\
                 
                 \cline{2-4}
                  & \multirow{4}{*}{ChaosFEX + SVM} & $C = 38.7$ & \multirow{4}{*}{0.958}\\
                 \cline{3-3}
                 
                  &  & $q = 0.21$ & \\
                 \cline{3-3}
                  & & $b = 0.969$ & \\
                 \cline{3-3}
                  & & $\epsilon = 0.13$ & \\
                \hline

                  \multirow{5}{*}{Ionosphere} & \multirow{1}{*}{Stand-Alone SVM} & $C = 2.6$ & \multirow{1}{*}{0.934}\\
                 
                 \cline{2-4}
                  & \multirow{4}{*}{ChaosFEX + SVM} & $C = 3.4$ & \multirow{4}{*}{0.926}\\
                 \cline{3-3}
                 
                  &  & $q = 0.21$ & \\
                 \cline{3-3}
                  & & $b = 0.969$ & \\
                 \cline{3-3}
                  & & $\epsilon = 0.37$ & \\
                \hline

                  \multirow{5}{*}{Wine} & \multirow{1}{*}{Stand-Alone SVM} & $C = 0.6$ & \multirow{1}{*}{0.975}\\
                 
                 \cline{2-4}
                  & \multirow{4}{*}{ChaosFEX + SVM} & $C = 16.0$ & \multirow{4}{*}{0.958}\\
                 \cline{3-3}
                 
                  &  & $q = 0.21$ & \\
                 \cline{3-3}
                  & & $b = 0.969$ & \\
                 \cline{3-3}
                  & & $\epsilon = 0.01$ & \\
                \hline
                 
                \multirow{5}{*}{Bank Note Authentication} & \multirow{1}{*}{Stand-Alone SVM} & $C = 1.3$ & \multirow{1}{*}{1.0}\\
                 
                 \cline{2-4}
                  & \multirow{4}{*}{ChaosFEX + SVM} & $C = 16.8$ & \multirow{4}{*}{0.965}\\
                 \cline{3-3}
                 
                  &  & $q = 0.080$ & \\
                 \cline{3-3}
                  & & $b = 0.250$ & \\
                 \cline{3-3}
                  & & $\epsilon = 0.233$ & \\
                \hline

                  \multirow{5}{*}{Haberman's Survival} & \multirow{1}{*}{Stand-Alone SVM} & $C = 19.8$ & \multirow{1}{*}{0.597}\\
                 
                 \cline{2-4}
                  & \multirow{4}{*}{ChaosFEX + SVM} & $C = 15.7$ & \multirow{4}{*}{0.609}\\
                 \cline{3-3}
                 
                  &  & $q = 0.81$ & \\
                 \cline{3-3}
                  & & $b = 0.14$ & \\
                 \cline{3-3}
                  & & $\epsilon = 0.003$ & \\
                \hline
                 
                  \multirow{5}{*}{Breast Cancer Wisconsin (Diagnostic)} & \multirow{1}{*}{Stand-Alone SVM} & $C = 16.0$ & \multirow{1}{*}{0.981}\\
                 
                 \cline{2-4}
                & \multirow{4}{*}{ChaosFEX + SVM} & $C = 20.3$ & \multirow{4}{*}{0.948}\\
                 \cline{3-3}
                 
                  &  & $q = 0.930$ & \\
                 \cline{3-3}
                  & & $b = 0.490$ & \\
                 \cline{3-3}
                  & & $\epsilon = 0.159$ & \\
                \hline

                \multirow{5}{*}{Statlog (Heart)} & \multirow{1}{*}{Stand-Alone SVM} & $C = 0.2$ & \multirow{1}{*}{0.852}\\
                 
                 \cline{2-4}
                & \multirow{4}{*}{ChaosFEX + SVM} & $C = 2.5$ & \multirow{4}{*}{0.825}\\
                 \cline{3-3}
                 
                  &  & $q = 0.08$ & \\
                 \cline{3-3}
                  & & $b = 0.06$ & \\
                 \cline{3-3}
                  & & $\epsilon = 0.17$ & \\
                \hline

                \multirow{5}{*}{Seeds} & \multirow{1}{*}{Stand-Alone SVM} & $C = 0.9$ & \multirow{1}{*}{0.948}\\
                 
                 \cline{2-4}
                & \multirow{4}{*}{ChaosFEX + SVM} & $C = 2.4$ & \multirow{4}{*}{0.909}\\
                 \cline{3-3}
                 
                  &  & $q = 0.020$ & \\
                 \cline{3-3}
                  & & $b = 0.070$ & \\
                 \cline{3-3}
                  & & $\epsilon = 0.238$ & \\
                  
                \hline

                \multirow{5}{*}{FSDD} & \multirow{1}{*}{Stand-Alone SVM (linear kernel)} & $C = 0.7$ & \multirow{1}{*}{0.952}\\
                 
                 \cline{2-4}
                & \multirow{4}{*}{ChaosFEX + SVM (linear kernel)} & $C = 6.1$ & \multirow{4}{*}{0.978}\\
                 \cline{3-3}
                 
                  &  & $q = 0.340$ & \\
                 \cline{3-3}
                  & & $b = 0.499$ & \\
                 \cline{3-3}
                  & & $\epsilon = 0.178$ & \\
                 
                \hline

                 \end{tabular}
                \label{table:Hyperparameter Tuning - SVM}
                \end{table}
                \newpage

\subsubsection{k-Nearest Neighbors}
Following are the hyperparameters tuned for $k$-Nearest Neighbors:
\begin{enumerate}
\item {$k$}: Defines the number of nearest training data samples to the testing data sample to be chosen. It is tuned from $1$ to $6$ with a step-size of $2$.

\end{enumerate}
All  remaining  hyperparameters  offered  by  scikit-learn  are  retained  in  their default forms.The results of hyperparameter tuning for $k$-Nearest Neighbors are available in Table ~\ref{table:Hyperparameter Tuning - $k$-NN}.\\

                \begin{table}[!ht]
                \centering
                \caption{\textbf{\textit{$k$-Nearest Neighbors ($k$-NN)}}: Tuned hyperparameters for all nine datasets. The performance metric used for the provided results is macro F1-score.}
                \begin{tabular}{| l | l | l | l |} 
                \hline
                 Dataset & Implementation & Tuned Hyperparameters & F1 Score \\ [0.5ex] 
                \hline
                 \multirow{5}{*}{Iris} & \multirow{1}{*}{Stand-Alone $k$-NN} & $k = 5$ & \multirow{1}{*}{0.944}\\
                 
                 \cline{2-4}
                  & \multirow{4}{*}{ChaosFEX + $k$-NN} & $k = 1$ & \multirow{4}{*}{0.936}\\
                 \cline{3-3}
                 
                  &  & $q = 0.21$ & \\
                 \cline{3-3}
                  & & $b = 0.969$ & \\
                 \cline{3-3}
                  & & $\epsilon = 0.12$ & \\
                \hline

                  \multirow{5}{*}{Ionosphere} & \multirow{1}{*}{Stand-Alone $k$-NN} & $k = 1$ & \multirow{1}{*}{0.814}\\
                 
                 \cline{2-4}
                  & \multirow{4}{*}{ChaosFEX + $k$-NN} & $k = 1$ & \multirow{4}{*}{0.886}\\
                 \cline{3-3}
                 
                  &  & $q = 0.21$ & \\
                 \cline{3-3}
                  & & $b = 0.969$ & \\
                 \cline{3-3}
                  & & $\epsilon = 0.11$ & \\
                \hline

                  \multirow{5}{*}{Wine} & \multirow{1}{*}{Stand-Alone $k$-NN} & $k = 5$ & \multirow{1}{*}{0.957}\\
                 
                 \cline{2-4}
                  & \multirow{4}{*}{ChaosFEX + $k$-NN} & $k = 1$ & \multirow{4}{*}{0.966}\\
                 \cline{3-3}
                 
                  &  & $q = 0.21$ & \\
                 \cline{3-3}
                  & & $b = 0.969$ & \\
                 \cline{3-3}
                  & & $\epsilon = 0.10$ & \\
                \hline

                \multirow{5}{*}{Bank Note Authentication} & \multirow{1}{*}{Stand-Alone $k$-NN} & $k = 5$ & \multirow{1}{*}{0.998}\\
                 
                 \cline{2-4}
                  & \multirow{4}{*}{ChaosFEX + $k$-NN} & $k = 3$ & \multirow{4}{*}{0.967}\\
                 \cline{3-3}
                 
                  &  & $q = 0.080$ & \\
                 \cline{3-3}
                  & & $b = 0.250$ & \\
                 \cline{3-3}
                  & & $\epsilon = 0.233$ & \\
                 \hline

                 \multirow{5}{*}{Haberman's Survival} & \multirow{1}{*}{Stand-Alone $k$-NN} & $k = 1$ & \multirow{1}{*}{0.562}\\
                 
                 \cline{2-4}
                  & \multirow{4}{*}{ChaosFEX + $k$-NN} & $k = 5$ & \multirow{4}{*}{0.659}\\
                 \cline{3-3}
                 
                  &  & $q = 0.81$ & \\
                 \cline{3-3}
                  & & $b = 0.14$ & \\
                 \cline{3-3}
                  & & $\epsilon = 0.003$ & \\
                 \hline

                 \multirow{5}{*}{Breast Cancer Wisconsin (Diagnostic)} & \multirow{1}{*}{Stand-Alone $k$-NN} & $k = 5$ & \multirow{1}{*}{0.964}\\
                 
                 \cline{2-4}
                  & \multirow{4}{*}{ChaosFEX + $k$-NN} & $k = 1$ & \multirow{4}{*}{0.932}\\
                 \cline{3-3}
                 
                  &  & $q = 0.930$ & \\
                 \cline{3-3}
                  & & $b = 0.490$ & \\
                 \cline{3-3}
                  & & $\epsilon = 0.159$ & \\
                \hline

                \multirow{5}{*}{Statlog (Heart)} & \multirow{1}{*}{Stand-Alone $k$-NN} & $k = 5$ & \multirow{1}{*}{0.803}\\
                 
                 \cline{2-4}
                  & \multirow{4}{*}{ChaosFEX + $k$-NN} & $k = 5$ & \multirow{4}{*}{0.791}\\
                 \cline{3-3}
                 
                  &  & $q = 0.08$ & \\
                 \cline{3-3}
                  & & $b = 0.06$ & \\
                 \cline{3-3}
                  & & $\epsilon = 0.17$ & \\
                  
                \hline
                 \multirow{5}{*}{Seeds} & \multirow{1}{*}{Stand-Alone $k$-NN} & $k = 5$ & \multirow{1}{*}{0.930}\\
                 
                 \cline{2-4}
                  & \multirow{4}{*}{ChaosFEX + $k$-NN} & $k = 1$ & \multirow{4}{*}{0.876}\\
                 \cline{3-3}
                 
                  &  & $q = 0.020$ & \\
                 \cline{3-3}
                  & & $b = 0.070$ & \\
                 \cline{3-3}
                  & & $\epsilon = 0.238$ & \\
                  
                \hline
                 \multirow{5}{*}{FSDD} & \multirow{1}{*}{Stand-Alone $k$-NN} & $k = 1$ & \multirow{1}{*}{0.834}\\
                 
                 \cline{2-4}
                  & \multirow{4}{*}{ChaosFEX + $k$-NN} & $k = 1$ & \multirow{4}{*}{0.909}\\
                 \cline{3-3}
                 
                  &  & $q = 0.340$ & \\
                 \cline{3-3}
                  & & $b = 0.499$ & \\
                 \cline{3-3}
                  & & $\epsilon = 0.178$ & \\

                \hline
                 \end{tabular}
                \label{table:Hyperparameter Tuning - $k$-NN}
                \end{table}
                \newpage

\subsubsection{ChaosNet}

\begin{enumerate}
\item $q$: Initial Neural Activity, it is varied from $0.01$ to $0.49$ with a step-size of $0.01$.
\item $b$: Discrimination Threshold, it is varied from $0.01$ to $0.49$ with a step-size of $0.01$.
\item $\epsilon$: Noise Intensity, it is varied from $0.001$ to $0.499$ with a step-size of $0.001$.
\end{enumerate}

\subsection{Results}
\subsubsection{Decision Tree}
The results of all experiments using Decision Tree in the high training sample regime are shown in Table~\ref{table:Experiment Results - Decision Tree}.
                \begin{table}[!ht]
                \centering
                \caption{\textbf{\textit{Decision Tree}}: Experiment results (test data macro F1-scores) for all nine datasets in the high training sample regime.}
                \begin{tabular}{|l |l| c|} 
                \hline
                 Dataset & Implementation & F1 Score \\ [0.5ex] 
                  & & (Test Data) \\ [0.5ex] 
                \hline
                 \multirow{2}{*}{Iris} & Stand-Alone Decision Tree & 0.967\\
                 \cline{2-3}
                  & ChaosFEX + Decision Tree & 1.0 \\
                 
                \hline
                
                 \multirow{2}{*}{Ionosphere} & Stand-Alone Decision Tree & 0.881\\
                 \cline{2-3}
                  & ChaosFEX + Decision Tree & 0.891\\

                \hline
                 \multirow{2}{*}{Wine} & Stand-Alone Decision Tree & 0.904\\
                 \cline{2-3}
                  & ChaosFEX + Decision Tree &  0.822\\

                \hline
                 \multirow{2}{*}{Bank Note Authentication} & Stand-Alone Decision Tree & 0.933\\
                 \cline{2-3}
                  & ChaosFEX + Decision Tree &  0.978\\

                \hline
                 \multirow{2}{*}{Haberman's Survival} & Stand-Alone Decision Tree & 0.516\\
                 \cline{2-3}
                  & ChaosFEX + Decision Tree &  0.482\\
                  
                \hline
                 \multirow{2}{*}{Breast Cancer Wisconsin (Diagnostic)} & Stand-Alone Decision Tree & 0.696\\
                 \cline{2-3}
                  & ChaosFEX + Decision Tree &  0.882\\

                \hline
                 \multirow{2}{*}{Statlog (Heart)} & Stand-Alone Decision Tree & 0.697\\
                 \cline{2-3}
                  & ChaosFEX + Decision Tree &  0.878\\  
                  
                \hline
                 \multirow{2}{*}{Seeds} & Stand-Alone Decision Tree & 0.880\\
                 \cline{2-3}
                  & ChaosFEX + Decision Tree &  0.880\\

                \hline
                 \multirow{2}{*}{FSDD} & Stand-Alone Decision Tree & 0.603\\
                 \cline{2-3}
                  & ChaosFEX + Decision Tree &  0.579\\
                  
                \hline
                 \end{tabular}
                \label{table:Experiment Results - Decision Tree}
                \end{table}
\newpage

\subsubsection{Random Forest}\label{Random Forest}

The results of all experiments using Random Forest in the high training sample regime are shown in Table ~\ref{table:Experiment Results - Random Forest}.
                \begin{table}[!ht]
                \centering
                \caption{\textbf{\textit{Random Forest}}: Experiment results (test data macro F1-scores) for all nine datasets in the high training sample regime.}
                \begin{tabular}{|l| l| c|} 
                \hline
                 Dataset& Implementation & F1 Score \\ [0.5ex] 
                  & & (Test Data) \\ [0.5ex] 
                \hline
                 \multirow{2}{*}{Iris} & Stand-Alone Random Forest & 1.0\\
                 \cline{2-3} 
                   & ChaosFEX + Random Forest & 1.0 \\
                \hline
                
                 \multirow{2}{*}{Ionosphere} & Stand-Alone Random Forest & 0.909\\
                 \cline{2-3} 
                  & ChaosFEX + Random Forest & 0.924\\
                \hline

                 \multirow{2}{*}{Wine} & Stand-Alone Random Forest & 0.966\\
                 \cline{2-3} 
                  & ChaosFEX + Random Forest &  0.943\\
                \hline

                 \multirow{2}{*}{Bank Note Authentication} & Stand-Alone Random Forest & 0.974\\
                 \cline{2-3} 
                  & ChaosFEX + Random Forest &  0.978\\
                \hline

                 \multirow{2}{*}{Haberman's Survival} & Stand-Alone Random Forest & 0.560\\
                 \cline{2-3} 
                  & ChaosFEX + Random Forest &  0.398\\
                \hline

                 \multirow{2}{*}{Breast Cancer Wisconsin (Diagnostic)} & Stand-Alone Random Forest & 0.919\\
                 \cline{2-3} 
                  & ChaosFEX + Random Forest &  0.918\\
                \hline

                 \multirow{2}{*}{Statlog (Heart)} & Stand-Alone Random Forest & 0.838 \\
                 \cline{2-3} 
                  & ChaosFEX + Random Forest &  0.838\\
                \hline

                 \multirow{2}{*}{Seeds} & Stand-Alone Random Forest & 0.877 \\
                 \cline{2-3} 
                  & ChaosFEX + Random Forest &  0.828\\
                  
                \hline
                  
                \multirow{2}{*}{FSDD} & Stand-Alone Random Forest & 0.970 \\
                 \cline{2-3} 
                  & ChaosFEX + Random Forest &  0.937\\

                \hline
                 
                 \end{tabular}
                \label{table:Experiment Results - Random Forest}
                \end{table}
                \newpage

\subsubsection{Adaptive Boosting (AdaBoost)}\label{AdaBoost}

The results of all experiments using AdaBoost in the high training sample regime are shown in Table ~\ref{table:Experiment Results - AdaBoost}.

                \begin{table}[!ht]
                \centering
                \caption{\textbf{\textit{AdaBoost}}: Experiment results (test data macro F1-scores) for all nine datasets in the high training sample regime.}
                \begin{tabular}{|l| l| c| } 
                \hline
                 Dataset & Implementation & F1 Score \\ [0.5ex] 
                  & & (Test Data) \\ [0.5ex] 
                \hline
                 \multirow{2}{*}{Iris} & Stand-Alone AdaBoost & 0.967\\
                 \cline{2-3}
                  &  ChaosFEX + AdaBoost & 0.865 \\
                 
                \hline
                 \multirow{2}{*}{Ionosphere} & Stand-Alone AdaBoost & 0.926\\
                 \cline{2-3}
                  &  ChaosFEX + AdaBoost & 0.925\\
                
                \hline
                
                 \multirow{2}{*}{Wine} & Stand-Alone AdaBoost & 0.833\\
                 \cline{2-3}
                  & ChaosFEX + AdaBoost &  0.846\\
                 
                \hline

                 \multirow{2}{*}{Bank Note Authentication} & Stand-Alone AdaBoost & 0.985\\
                 \cline{2-3}
                  &  ChaosFEX + AdaBoost &  0.910\\
                  
                \hline
                 
                 \multirow{2}{*}{Haberman's Survival} & Stand-Alone AdaBoost & 0.505 \\
                 \cline{2-3}
                  &  ChaosFEX + AdaBoost &  0.609\\

                \hline
                 
                 \multirow{2}{*}{Breast Cancer Wisconsin (Diagnostic)} & Stand-Alone AdaBoost & 0.858\\
                 \cline{2-3}
                  &  ChaosFEX + AdaBoost &  0.881\\
                  
                \hline
                 
                 \multirow{2}{*}{Statlog (Heart)} & Stand-Alone AdaBoost & 0.777\\
                 \cline{2-3}
                  &  ChaosFEX + AdaBoost &  0.717\\

                \hline
                
                 \multirow{2}{*}{Seeds} & Stand-Alone AdaBoost & 0.746\\
                 \cline{2-3}
                  &  ChaosFEX + AdaBoost &  0.873\\
                  
                \hline
                
                 \multirow{2}{*}{FSDD} & Stand-Alone AdaBoost & 0.080\\
                 \cline{2-3}
                  &  ChaosFEX + AdaBoost &  0.397\\

                \hline
                 
                 \end{tabular}
                \label{table:Experiment Results - AdaBoost}
                \end{table}
                \newpage

\subsubsection{Support Vector Machine (SVM)}

The results of all experiments using Support Vector Machine (SVM) in the high training sample regime are shown in Table ~\ref{table:Experiment Results - SVM}.

                \begin{table}[!ht]
                \centering
                \caption{\textbf{\textit{Support Vector Machine (SVM)}}: Experiment results (test data macro F1-scores) for all nine datasets in the high training sample regime.}
                \begin{tabular}{|l| l| c| } 
                \hline
                 Dataset & Implementation & F1 Score \\ [0.5ex] 
                  & & (Test Data) \\ [0.5ex] 
                \hline
                 \multirow{2}{*}{Iris} & Stand-Alone SVM & 0.966\\
                 \cline{2-3}
                  &  ChaosFEX + SVM & 0.933 \\
                 
                \hline
                 \multirow{2}{*}{Ionosphere} & Stand-Alone SVM & 0.924\\
                 \cline{2-3}
                  &  ChaosFEX + SVM & 0.909\\
               
                \hline
                
                 \multirow{2}{*}{Wine} & Stand-Alone SVM & 0.928\\
                 \cline{2-3}
                  & ChaosFEX + SVM &  0.896\\
                 
                \hline

                 \multirow{2}{*}{Bank Note Authentication} & Stand-Alone SVM & 0.993\\
                 \cline{2-3}
                  &  ChaosFEX + SVM &  0.978\\
                  
                \hline
                 
                 \multirow{2}{*}{Haberman's Survival} & Stand-Alone SVM & 0.437\\
                 \cline{2-3}
                  &  ChaosFEX + SVM &  0.447\\
                
                \hline
                 
                 \multirow{2}{*}{Breast Cancer Wisconsin (Diagnostic)} & Stand-Alone SVM & 0.824\\
                 \cline{2-3}
                  &  ChaosFEX + SVM &  0.918\\

                \hline
                 
                 \multirow{2}{*}{Statlog (Heart)} & Stand-Alone SVM & 0.844\\
                 \cline{2-3}
                  &  ChaosFEX + SVM &  0.801\\
                  
                \hline
                 
                 \multirow{2}{*}{Seeds} & Stand-Alone SVM & 0.924\\
                 \cline{2-3}
                  &  ChaosFEX + SVM &  0.827\\
                  
                \hline
                 
                 \multirow{2}{*}{FSDD} & Stand-Alone SVM & 0.952\\
                 \cline{2-3}
                  &  ChaosFEX + SVM &  0.978\\

                \hline
                 
                 \end{tabular}
                \label{table:Experiment Results - SVM}
                \end{table}
                \newpage

\subsubsection{k-Nearest Neighbors}

The results of all experiments using $k$ - Nearest Neighbors in the high training sample regime are shown in Table ~\ref{table:Experiment Results - $k$-NN}.

               \begin{table}[!ht]
                \centering
                \caption{\textbf{\textit{$k$-Nearest Neighbors ($k$-NN)}}: Experiment results (test data macro F1-scores) for all nine datasets in the high training sample regime.}
                \begin{tabular}{|l| l| c| } 
                \hline
                 Dataset & Implementation & F1 Score \\ [0.5ex] 
                  & & (Test Data) \\ [0.5ex] 
                \hline
                 \multirow{2}{*}{Iris} & Stand-Alone $k$-NN & 1.0\\
                 \cline{2-3}
                  &  ChaosFEX + $k$-NN & 0.902 \\
                 
                \hline
                 \multirow{2}{*}{Ionosphere} & Stand-Alone $k$-NN & 0.821\\
                 \cline{2-3}
                  &  ChaosFEX + $k$-NN & 0.939\\
                
                \hline
                
                 \multirow{2}{*}{Wine} & Stand-Alone $k$-NN & 0.943\\
                 \cline{2-3}
                  & ChaosFEX + $k$-NN &  0.873\\

                \hline
                 
                 \multirow{2}{*}{Bank Note Authentication} & Stand-Alone $k$-NN & 0.993\\
                 \cline{2-3}
                  &  ChaosFEX + $k$-NN &  0.971\\
                  
                \hline
                 
                 \multirow{2}{*}{Haberman's Survival} & Stand-Alone $k$-NN & 0.480\\
                 \cline{2-3}
                  &  ChaosFEX + $k$-NN &  0.455\\
                  
                \hline
                 
                 \multirow{2}{*}{Breast Cancer Wisconsin (Diagnostic)} & Stand-Alone $k$-NN & 0.954\\
                 \cline{2-3}
                  &  ChaosFEX + $k$-NN &  0.935\\
                  
                \hline
                 
                 \multirow{2}{*}{Statlog (Heart)} & Stand-Alone $k$-NN & 0.847\\
                 \cline{2-3}
                  &  ChaosFEX + $k$-NN &  0.739\\
                  
                \hline
                 
                 \multirow{2}{*}{Seeds} & Stand-Alone $k$-NN & 0.924\\
                 \cline{2-3}
                  &  ChaosFEX + $k$-NN &  0.854\\
                  
                \hline
                 
                 \multirow{2}{*}{FSDD} & Stand-Alone $k$-NN & 0.885\\
                 \cline{2-3}
                  &  ChaosFEX + $k$-NN &  0.938\\

                \hline
                 
                 \end{tabular}
                \label{table:Experiment Results - $k$-NN}
                \end{table}
                \newpage

\subsubsection{Gaussian Naive Bayes}
 
The results of all experiments using Gaussian Naive Bayes (GNB) in the high training sample regime are shown in Table ~\ref{table:Experiment Results - Gaussian Naive Bayes}.
                \begin{table}[!ht]
                \centering
                \caption{\textbf{\textit{Gaussian Naive Bayes (GNB)}}: Experiment results (test data macro F1-scores) for all nine datasets in the high training sample regime.}
                \begin{tabular}{|l |c |c |} 
                \hline
                 Dataset & Implementation & F1 Score \\ [0.5ex] 
                  & & (Test Data) \\ [0.5ex] 
                \hline
                 \multirow{2}{*}{Iris} & Stand-Alone GNB & 0.966\\
                 \cline{2-3}
                  &  ChaosFEX + GNB & 0.933 \\
                 
                \hline
                 \multirow{2}{*}{Ionosphere} & Stand-Alone GNB & 0.862\\
                 \cline{2-3}
                  &  ChaosFEX + GNB & 0.771\\
                
                \hline
                
                 \multirow{2}{*}{Wine} & Stand-Alone GNB & 1.0\\
                 \cline{2-3}
                  & ChaosFEX + GNB &  0.976\\
                 
                \hline

                 \multirow{2}{*}{Bank Note Authentication} & Stand-Alone GNB & 0.785\\
                 \cline{2-3}
                  &  ChaosFEX + GNB &  0.781\\

                \hline
                 
                 \multirow{2}{*}{Haberman's Survival} & Stand-Alone GNB & 0.572\\
                 \cline{2-3}
                  &  ChaosFEX + GNB &  0.535\\

                \hline
                 
                 \multirow{2}{*}{Breast Cancer Wisconsin (Diagnostic)} & Stand-Alone GNB & 0.858\\
                 \cline{2-3}
                  &  ChaosFEX + GNB &  0.812\\

                \hline
                 
                 \multirow{2}{*}{Statlog (Heart)} & Stand-Alone GNB & 0.826\\
                 \cline{2-3}
                  &  ChaosFEX + GNB &  0.788\\ 
                  
                \hline
                 
                 \multirow{2}{*}{Seeds} & Stand-Alone GNB & 0.846\\
                 \cline{2-3}
                  &  ChaosFEX + GNB &  0.849\\
                  
                \hline
                 
                 \multirow{2}{*}{FSDD} & Stand-Alone GNB & 0.919\\
                 \cline{2-3}
                  &  ChaosFEX + GNB &  0.687\\
                
                \hline
                 \end{tabular}
                \label{table:Experiment Results - Gaussian Naive Bayes}
                \end{table}
                \newpage

\end{document}